\documentclass[runningheads]{llncs}
\usepackage{graphicx}
\usepackage{tikz}
\usepackage{comment}
\usepackage{amsmath,amssymb} % define this before the line numbering.
\usepackage{color}
\usepackage{enumitem}
\usepackage{url}

\usepackage[accsupp]{axessibility}  % Improves PDF readability for those with disabilities.

\DeclareMathOperator{\simmatch}{sim}

\begin{document}
% \linenumbers
\pagestyle{headings}
\mainmatter
\def\ECCVSubNumber{4513}  % Insert your submission number here

\title{Visual Cross-View Metric Localization with Dense Uncertainty Estimates} % Replace with your title

\begin{comment}
\titlerunning{ECCV-22 submission ID \ECCVSubNumber} 
\authorrunning{ECCV-22 submission ID \ECCVSubNumber} 
\author{Anonymous ECCV submission}
\institute{Paper ID \ECCVSubNumber}
\end{comment}
%******************

% \begin{comment}
\titlerunning{Visual Cross-View Metric Localization with Dense Uncertainty Estimates}
\author{Zimin Xia\inst{1}\orcidID{0000-0002-4981-9514} \and
Olaf Booij\inst{2} \and
% \orcidID{1111-2222-3333-4444} \and
Marco Manfredi\inst{2}\orcidID{0000-0002-2618-2493} \and
Julian F. P. Kooij\inst{1}\orcidID{0000-0001-9919-0710}
}
\authorrunning{Z. Xia et al.}
\institute{Intelligent Vehicles Group, Technical University Delft, The Netherlands \email{\{z.xia,j.f.p.kooij\}@tudelft.nl} \and
TomTom, Amsterdam, The Netherlands 
\email{\{olaf.booij,marco.manfredi\}@tomtom.com}
% \\
}
% \end{comment}
%******************
\maketitle
\setcounter{footnote}{0}
%%%%%%%%% ABSTRACT
% CVPR guidelines: max 4000 characters to fit in registration form
\begin{abstract}

%Ground-to-satellite image retrieval has received increasing attention in recent years since the retrieved satellite patch for a query ground-level camera image serves as a rough global localization estimate similar to GPS/GNSS or temporal filtering.
%After such rough self-localization, results could be improved through more fine-grained metric localization within the local satellite patch, a mostly unexplored task in the vision domain.
%Past work on cross-view metric localization instead considered ground-level measurements with range sensors (Lidar, Radar), as their 3D data can readily be projected onto the top-down viewpoint for comparison to the local satellite map.
%Our work addresses cross-view metric localization from a ground-level color image, which lacks accurate range information, within a given satellite patch.

This work addresses visual cross-view metric localization for outdoor robotics.
Given a ground-level color image and a satellite patch that contains the local surroundings, the task is to identify the location of the ground camera within the satellite patch.
Related work addressed this task for range-sensors (LiDAR, Radar), but for vision, only as a secondary regression step after an initial cross-view image retrieval step.
Since the local satellite patch could also be retrieved through any rough localization prior (e.g. from GPS/GNSS, temporal filtering), we drop the image retrieval objective and focus on the metric localization only.
We devise a novel network architecture with denser satellite descriptors, similarity matching at the bottleneck (rather than at the output as in image retrieval), and a dense spatial distribution as output to capture multi-modal localization ambiguities.
We compare against a state-of-the-art regression baseline that uses global image descriptors. 
Quantitative and qualitative experimental results on the recently proposed VIGOR and the Oxford RobotCar datasets validate our design.
The produced probabilities are correlated with localization accuracy,
and can even be used to roughly estimate the ground camera's heading when its orientation is unknown.
Overall, our method reduces the median metric localization error by 51\%, 37\%, and 28\% compared to the state-of-the-art when generalizing respectively in the same area, across areas, and across time. 

\end{abstract}

%%%%%%%%% BODY TEXT
\section{Introduction}
\label{sec:intro}

Ground-to-aerial/satellite image matching, also known as cross-view image matching, has shown notable performance in large-scale geolocalization \cite{LocOriStreet,WideAreaImageGeolocalization,ground-to-aerialgeolocalization,CVM-Net,CVACT,SAFA,zhu2021vigor,Toker_2021_CVPR}. 
% Such approaches formulate the localization task as image retrieval.
% The term geolocalization indicates that the focus is, given a ground-level query image, finding its location in a large-scale environment, e.g. city- or country-scale~\cite{CVACT,CVUSA}.
Usually, this global localization task is formulated as image retrieval.
For each ground-level query image
% e.g. as taken with a dashboard camera in a car, 
the system retrieves the most similar geo-tagged aerial/satellite patch in the database and uses the location of the center pixel in that patch as the location of the query.
In practice, global localization can also be obtained by other means in outdoor robotics, such as temporal filtering or coarse GPS/GNSS \cite{tang2020rsl,Geolocal_feature,9449965}, but can still have errors of tens of meters \cite{Geolocal_feature,9449965,benmoshe2011urbangnss}.
%Therefore, an equally important follow-up task is localization within the local uncertainty region.
In this work, we therefore follow~\cite{tang2020rsl,Geolocal_feature,9449965} by exploiting a coarse location estimate, and zoom into fine-grained \textit{metric localization} within a known satellite image,
i.e.~to identify which image coordinates in the satellite patch correspond to the location of ground measurement.
We adopt the common assumption \cite{CVACT,SAFA,zhu2021vigor,shi2020optimal,Toker_2021_CVPR} of known orientation,
e.g.~the center of a ground panorama points north,
though we will seek to loosen this restriction in our experiments and roughly estimate 
% the orientation (i.e. the camera's ``heading'') too.
the camera's heading too.
%While orientation could be estimated during image retrieval \cite{zhu2021revisiting,shi2020looking}, or though magnetometer sensors fusion \cite{won2015performance},
%it would ideally be estimated jointly with metric localisation.

%However, we additionally show in our experiments that 
%Additionally, we experiment the localization with perturbed orientation or to directly infer the rough orientation.

% \begin{figure}[t!]
%     \centering
%     \begin{subfigure}{0.4\textwidth}
%     \includegraphics[width=1\textwidth]{figures/introduction/multi_ground_14323.pdf}\\\vspace*{-1.6em}
%     \includegraphics[width=1\textwidth]{figures/introduction/multi_matching_14323.pdf}
%     \end{subfigure}\hspace*{-0.6em}
%     \begin{subfigure}{0.6\textwidth}
%     \includegraphics[width=1\textwidth]{figures/introduction/multi_heatmap_14323.pdf}
%     \end{subfigure}
%     \caption{Exemplar cross-view metric localization. Top left: input ground image. Bottom left: coarse localization heat map produced by our proposed method. Right: input satellite images overlayed with our dense estimation of the (log) probability heat map (red, peak at the yellow star). 
%     When the scene layout is ambiguous, our model explores this information, while the baseline regresses to the midpoint among two modes. Best viewed in color.
%     }
%     \label{fig:VIGOR_qualitative}
% \end{figure}

\begin{figure}[t!]
    \centering
    \includegraphics[width=0.8\textwidth]{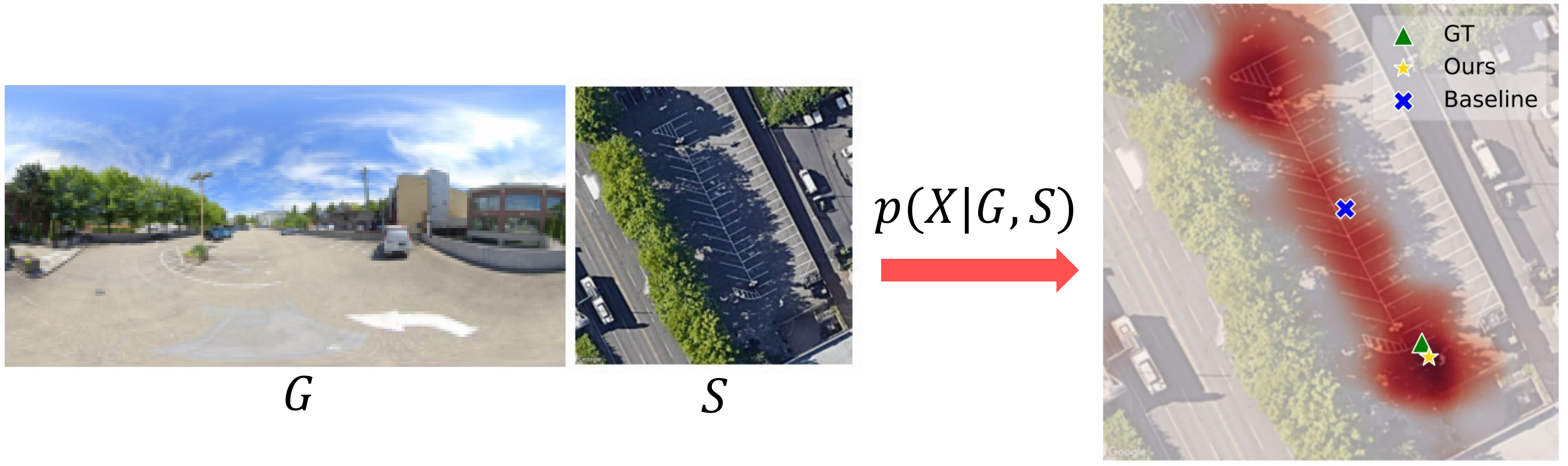} 
    \caption{Example of visual cross-view metric localization. Given a ground-level image $G$ (left), and a satellite patch $S$ (middle) with its local area, we aim to identify the location $X$ within $S$ where $G$ was taken.
    Our method estimates a dense probability distribution over the satellite image. The resulting (log) probability heat map is overlayed in red on top of the satellite patch (right). 
    Compared to the regression-based baseline that tends to roughly regress to the midpoint among multiple modes, our method captures the underlying multi-modal distribution. Our final predicted location, $\textrm{argmax}(p(X|G,S))$, is closer to the ground truth.
    }
    \label{fig:intro}
\end{figure}

In vision, even though ground-to-ground metric localization is a well-studied task \cite{agarwal2015metric,6731512,8963763}, so far in the cross-view setting, the only end-to-end approach that considers metric localization is the regression-based approach proposed in \cite{zhu2021vigor}, which we will refer here to as \textit{Cross-View Regression} (CVR) for simplicity.
CVR tries to solve both the global coarse localization and local metric localization. 
As a result, its metric localization regressor is built on top of global image descriptors and might miss fine-grained scene information from the satellite image.

Rather than formulating visual cross-view metric localization as a regression task, we propose to produce a dense multi-modal distribution to capture localization ambiguities, and avoid regressing to the midpoint between multiple visually similar places,
see Figure~\ref{fig:intro}.
To capture more spatial information, we compute multiple local satellite image descriptors rather than a single global one, and train these in a locally discriminative manner.
We note that dense uncertainty output for localization was shown to be successful with range-sensing modalities, like LiDAR and Radar, for localization within
% e.g. to localize in the satellite image~\cite{tang2020rsl,tang2021self,tangget} or other 
top-down maps \cite{yin2021rall,barsan2020learning,wei2019learning}.
However, these methods are not directly applicable to monocular vision, as they rely on highly accurate depth information which images lack.
Unlike existing literature \cite{LocOriStreet,WideAreaImageGeolocalization,CVM-Net,CVACT,SAFA,zhu2021vigor,shi2020optimal,Toker_2021_CVPR}, we address local metric localization as a standalone task in visual cross-view matching,
and make the following contributions:
% We are the first to address end-to-end visual cross-view metric localization as a standalone task. As part of an outdoor localization stack, the relevant satellite patch can be retrieved via a prior localization estimate, hence we avoid the need for global image descriptors required for joint image retrieval.
% Unlike the regression approach used in the state-of-the-art, we propose to predict a dense multi-modal distribution, a desirable property for integration into a localization stack. 
%found in the retrieval-focused state-of-the-art.
% (i)~
% We propose a new Siamese-like network
% that exploits multiple local satellite descriptors
% and uses similarity matching in the fusion bottleneck.
% It combines the metric learning paradigm from image retrieval
% with dense probabilistic output via a UNet-style decoder,
% found previously only in range-based cross-view localization.
% The proposed method enables us to predict a dense multi-modal distribution, a more desirable property than uni-modal regression.
(i)~
We propose to predict a dense multi-modal distribution for localization, which can represent localization ambiguity.
For this, we propose a new Siamese-like network
that exploits multiple local satellite descriptors
and uses similarity matching in the fusion bottleneck.
It combines the metric learning paradigm from image retrieval
with dense probabilistic output via a UNet-style decoder,
found previously only in range-based cross-view localization.
% (ii)~
% We show that the produced distribution correlates with localization quality and that it can express localization ambiguity. Importantly, this also results in significantly lower median localization error than the state-of-the-art.
(ii)~
We show that the produced distribution correlates with localization quality, a desirable property for outlier detection, temporal filtering, and multi-sensor fusion.
Besides, we also achieved significantly lower median localization error than the state-of-the-art.
(iii)~
We show our proposed method is robust against small perturbations on the assumed orientation,
and that the model's probabilistic output can even be used to classify a ground image orientation when it is unknown.

Our experiments use the recent large-scale VIGOR dataset for standalone cross-view metric localization to test generalization to new locations in both known and unknown areas.
We also collect and stitch additional satellite data for data augmentation and metric localization on the Oxford RobotCar dataset, testing generalization to new measurements along the same route across time.\footnote{Models and code, plus extended data  are available at \\ \url{https://github.com/tudelft-iv/CrossViewMetricLocalization}}

% We open our code, and evaluate 

% (i) We propose the first architecture for visual ground-to-satellite metric localization with an uncertainty estimate. Compared to the regression-based method which limits the output to uni-modal, our model naturally handles multi-modal distributions.
% More importantly, unlike the regression-based baseline that only outputs a single location, our estimated uncertainty helps us to interpret our prediction, which is essential in improving the trust in safety-crucial applications, such as autonomous driving. 
% (ii) Comparing to the SOTA baseline \cite{zhu2021vigor}, we reduce the median localization error by 55\% when generalizing to new measurements in the same area, and 44\% when generalizing to images in different areas on the VIGOR dataset. On the Oxford RobotCar dataset, we reduce median localization error by 19\%.
% (iii) We include the probability at ground truth as an additional metric. It is important in practice when applying temporal filtering to the measurements. Our model consistently predicts higher probabilities at ground truth locations than the baseline with a post-processing probability estimation. 
% (iv) We extensively study visual ground-to-satellite metric localization, including the problem formulation, loss function, network architecture, and the way to present the training data. 
\section{Related Work}
\label{sec:relatedwork}
We here review the works most related to visual cross-view metric localization.

\textbf{Cross-view image retrieval} is a special case of image retrieval.
For place recognition \cite{lowry2015visual}, a majority of works~\cite{torii201524,chen2011city,7054472} construct a reference database using ground-level images, but it is infeasible to guarantee the coverage of the images everywhere.
Alternatively, satellite images provide continuous coverage over the world and are publicly available.
Given this advantage, a series of approaches \cite{ground-to-aerialgeolocalization,WideAreaImageGeolocalization,LocOriStreet} have been proposed to solve large-scale geolocalization using ground-to-satellite cross-view image retrieval.
% Despite the large domain gap between satellite and ground views, over the past, the community has witnessed a steady increase in the performance of deep cross-view image retrieval networks.
CVM-Net \cite{CVM-Net} adopts the powerful image descriptor NetVLAD \cite{NetVLAD} to summarize the view-point invariant information for the cross-view image retrieval.
In \cite{CVACT}, the authors encode the azimuth and altitude of the pixels in the ground-level query to guide the ground-to-satellite matching.
To explicitly minimize the visual difference between satellite and ground domains, various improvements have been proposed. 
SAFA \cite{SAFA} proposes to use a polar transformation to warp the satellite patch towards the ground-level panorama and uses attention modules to extract the specific features that are visible from both views.
In \cite{regmi2019bridging,Toker_2021_CVPR}, a conditional GAN \cite{isola2017image} is used to generate synthetic satellite images from the ground-level panorama or to synthesize the panoramic street view from the satellite image to direct the cross-view matching.
Instead of constructing a visually similar input, CVFT \cite{shi2020optimal} tries to transport the features from the ground domain towards the satellite domain inside an end-to-end network.
Some works \cite{zhu2021revisiting,shi2020looking,LocOriStreet} jointly estimate the orientation of the ground query during retrieval without any metric localization.
Recently, transformers \cite{yang2021cross,zhu2022transgeo} are also used in cross-view image retrieval.

% Triggered by the impressive performance of cross-view image retrieval, many works try to understand how the model performs the matching.
% Many works \cite{CVM-Net,ground-to-aerialgeolocalization,LocOriStreet} show that the top retrieved candidates are indeed visually similar. 
% Later, a few works \cite{SAFA,zhu2021visual,shi2020optimal} back-propagate \cite{selvaraju2017grad,zeiler2014visualizing} the feature to the input image to see which part of the input contributes to the matching.
% Targeting at local localization \cite{Geolocal_feature} shows that the same model can extract features on different objects comparing global localization.
% All those qualitative results prove that the models have enough capability to match images from two drastically different domains.

% \textbf{Other forms of visual geolocalization} have also been studied in past years.
% PlaNet \cite{weyand2016planet} divides the earth into geographical cells and classifies ground images w.r.t. those.
% However, it is infeasible to use this for fine-grain localization with high map resolution.
% Plus, akin to PoseNet family \cite{kendall2015posenet,kendall2017geometric}, PlaNet encodes the map into the model, the generalization to unseen area is difficult \cite{sattler2019understanding}.

\textbf{Limitations in cross-view image retrieval} are also evident despite its increasing popularity for geolocalization.
Recently, \cite{zhu2021vigor} points out that cross-view image retrieval methods assume that query ground images correspond to the center of satellite patches in the database, and this assumption is not valid during test time.
To break this assumption, \cite{zhu2021vigor} introduces a new cross-view matching benchmark VIGOR in which the ground images are not aligned with the center of satellite patches. 
Another limitation of retrieval is the trade-off between localization accuracy and computation or dataset density. To acquire meter-level localization accuracy, reference satellite patches often have a large overlap with each other, such as sampling the patch every 5m as done in \cite{CVMJournal,9449965}.
% The accuracy is gained by trading the computation and database size off.
% And they propose a method to first retrieve a satellite patch and then regress the offset between the ground-level query and the center of the retrieved satellite patch. 

\textbf{Range sensing sensors-to-satellite metric localization} received more attention than its visual counterpart.
RSL-Net \cite{tang2020rsl} localizes Radar scans on a known satellite image. 
This task is formulated as generating a top-down Radar scan conditioned on the satellite image using \cite{isola2017image}, and then comparing the online scan to synthetic scan for pose estimation.
Later, this idea is extended to self-supervised learning \cite{tang2021self}.
% As these works target the fine-grain metric localization, a realistic assumption is made that the coarse location prior is known from other means. 
% This means that we know which satellite patch covers the ground measurement and the goal is to refine metric localization.
In \cite{tangget}, the top-down representation of a LiDAR scan is compared to UNet~\cite{ronneberger2015u} encoded satellite features for metric localization.
The range information is crucial in representing the measurement in a top-down view.

\textbf{LiDAR-to-BEV map metric localization} is another frontier that benefited from the range sensing.
Dense pixel-to-pixel matchable LiDAR and bird's eye view (BEV) map embeddings can be learned by a deep network \cite{barsan2020learning}. 
Localization becomes finding the position that has the maximum cross-correlation between two embeddings.
Later work \cite{wei2019learning} shows that it is possible to localize the online LiDAR sweep on HD maps in a similar manner.
Those works deliver a dense probabilistic output by formulating the localization task as a classification problem.
This property is ideal in probabilistic robot localization \cite{Thrun:2005:PR:1121596}, as it enables multi-sensor fusion and temporal filtering.

\textbf{Visual ground-to-satellite metric localization} cannot directly reuse
the same architecture used to localize LiDAR scans in a BEV map,
since an RGB ground image does not provide reliable depth information. % not perceive depth accurately
%, it is hard to build a metric-aware top-down representation from it.
Hence pixel-level dense comparison, such as cross-correlation, cannot be leveraged.
% The main disadvantage in vision is the lacking of the ability to sense depth accurately.
% Therefore, it is difficult to build a dense pixel-to-pixel comparable representation from the perspective ground image and top-down satellite image.
\cite{zhai2017predicting} predicts ground-view semantics from aerial imagery for orientation estimation, and shows only qualitatively that metric localization is possible by comparing the predicted semantics across viewpoints.
To the best of our knowledge, CVR \cite{zhu2021vigor} is the only end-to-end approach in the vision domain that attempts metric localization on a satellite patch.
Given a ground-level query, it first retrieves the matched satellite patch and then regresses the offset between the ground image and satellite patch center.
However, its offset regression is based on global feature descriptors, which might cause the regression head to miss detailed scene layout information,
and it limits the output to uni-modal estimates.
% Thus, there is still large room to improve the localization accuracy.
% Plus, CVR lacks uncertainty estimation, dense output to identify ambiguous locations, or a way to filter out unreliable results. 
Plus, CVR lacks dense uncertainty estimation to identify ambiguous locations, or a way to filter out unreliable results.
% This reduces its usefulness for multi-sensor fusion and temporal filtering.
A concurrent work \cite{shi2020beyond} perform unimodal localization and orientation estimation by warping features across views and solving an iterative optimization.
\begin{figure}[t] 
    \centering
    \includegraphics[width=12cm]{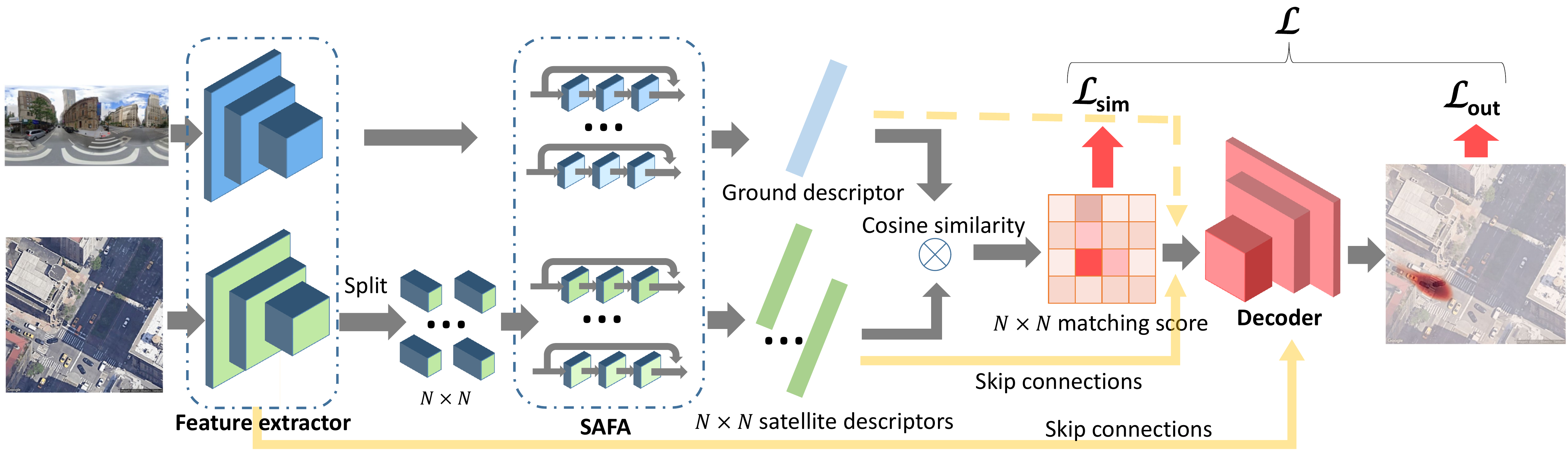}
    \caption{An overview of the proposed cross-view metric localization architecture (trainable parts in bold). 
    Dashed skip connection is optional, see ablation study.
    % It is an encoder-decoder architecture with the ground-to-satellite descriptor matching at the bottleneck. The combined loss function guides the matching at the bottleneck and enforces the ground truth location in the final dense output has the highest probability.
    We overlay an exemplar output heat map on top of the input satellite image for intuition.}
    \label{fig:network_architecture}
\end{figure}

\section{Methodology}
\label{sec:methodology}
In our work, we assume that a rough prior localization estimate is available, e.g.~through GPS/GNSS, odometry, or some other robot-localization techniques~\cite{Geolocal_feature,9449965,tang2020rsl}.
Given a ground-level image $G$ and a top-down $L \times L$ satellite image $S$ that represents the local area where $G$ was taken, our metric localization objective is to estimate the 2D image coordinates $X \in [0,1]^2$ within $S$ that correspond to the ground location of the camera of $G$.
Moreover, we aim for a dense probabilistic output to benefit a downstream sensor fusion task, similar to~\cite{barsan2020learning}.
Note that in practice, $G$ and $S$ are often provided with their heading pre-aligned \cite{zhu2021vigor,CVACT}, such that the center vertical line of $G$ points in the up direction of $S$.

Both the baseline CVR~\cite{zhu2021vigor} and our proposed method adapt a common cross-view image retrieval architecture~\cite{SAFA}.
This basic backbone is a Siamese-like architecture without weight-sharing.
Both the ground and satellite input branches consist of a VGG \cite{VGG} feature extractor.
E.g. for the satellite branch, these features form a $L' \times L' \times 512$ volume.
On the feature volume 8 Spatial-Aware Feature Aggregation (SAFA) modules \cite{SAFA} are applied, each generating a 512-dimensional vector, which is all concatenated.  Each branch thus yields a single global $1 \times 1 \times 4096$-dimensional descriptor.
In an image retrieval task, this network would be trained through metric learning such that descriptors of matching $(S, G)$ pairs are close together in the 4906-dimensional space.

Importantly, our proposed architecture and CVR make distinct choices on (1) the used descriptor representation for $S$, (2) how the descriptors are fused, (3) how the output head represents the localization result, and (4) consequently, the losses.
We explain these choices for both methods in turn.

% \subsection{Cross-view metric localization}
% Cross-view image geo-localization has attracted a lot of attention in the past years \cite{WideAreaImageGeolocalization,LocOriStreet,CVM-Net,SAFA}.
% The term geo-localization indicates that the focus is, given a query, recognizing its location in a large-scale environment, e.g. city- or country-scale \cite{CVACT,CVUSA}.
% Aiming at this, a continuous aerial/satellite image is divided into multiple patches, and a reference database is created by storing those patches or their image descriptors together with their geographical location information.

% In Lidar and Radar domain, cross-view localization zooms into the fine-grain metric localization \cite{tang2020rsl,tang2021self,tangget} with the objective of identifying which pixel in a known satellite patch corresponds to the location of the ground measurement.
% This can be seen as a follow-up task of cross-view geolocalization, but in practice, there are also other means to acquire the global coarse localization.

% So far, in the vision domain, the cross-view metric localization is less studied. 
% Compared to the range sensing sensors, a single camera view cannot accurately capture the depth. 
% This makes it difficult to build one-to-one densely comparable representations for the measurement and top-down satellite image.
% So far, the only work in the vision domain that tackles the cross-view metric localization is VIGOR \cite{zhu2021vigor}, and we will use this as our baseline.

\subsection{Baseline Cross-View Regression}
The CVR method in \cite{zhu2021vigor} uses a single architecture for a two-step approach. First global localization is done through image retrieval by comparing descriptor $G$ to descriptors of all known satellite patches.
After retrieving satellite patch $S$, metric localization is performed using the already computed descriptors of both $G$ and $S$.
We employ CVR here for the metric localization task only, and therefore keep its proposed architecture, but will not train it for image retrieval.
Focusing on metric localization only, our CVR baseline makes the following design choices:
%Our objective is metric localization in a known satellite patch.
%To make a fair comparison to this baseline, we will train it with regression task only for cross-view metric localization in our experiments.

\textbf{Feature descriptors:}
CVR follows the image-retrieval concept of encoding the satellite and ground image each into a single image-global 4096-dimensional descriptor.
Both descriptors are fed as-is to the fusion step.
\textbf{Fusion:}
CVR simply concatenates the two feature descriptors into a single 8192-dimensional vector.
\textbf{Output head:}
A multi-layer perceptron is used on the fused descriptors which outputs the relative 2D offset $\Delta X$ between $G$'s true location within $S$ and the center $X_S = (0.5, 0.5)$ of the satellite patch, s.t.~$X = X_S + \Delta X$.
\textbf{Loss:}
The standard L2 regression loss is used on the predicted offset and true offset.
%In this case, the offset regression is meaningful only when the input satellite patch covers the ground query.
%In other words, when the global coarse localization is correct.

We note that most of these choices follow from the need to use a single global descriptor for a whole satellite patch, as such descriptors are necessary for image retrieval.
Our argument is however that if a localization prior is already available and global image retrieval is not necessary, this state-of-the-art architecture is sub-optimal for metric localization only compared to our proposed approach.

\subsection{Proposed method}
% By checking the qualitative results from VIGOR \cite{zhu2021vigor}, we observe that the predicted location is sometimes on the building while the query locates on the road.
% We reckon that this could be because the global image descriptors for offset regression lacks scene layout information from the satellite view.
% Besides, there is still large room to improve the overall metric localization accuracy.

% In this work, we adopt the same assumption used in \cite{tang2020rsl,tang2020self,tang2021self} that a coarse global location estimate is available and therefore we know which satellite patch, e.g. covers $\mathtt{\sim}70 meters \times70 meters$ ground area, contains the current query location.
% This is usually valid for the task of outdoor localization, such as in autonomous driving.
% Hence, we aim at a more important task, metric localization, namely finding out which pixel in the satellite patch corresponds to the ground query.

Our proposed architecture starts with a mostly similar Siamese-like backbone.
%While other feature extractors could be used, we also stick to VGG and the SAFA feature aggregation modules commonly used in the cross-view image retrieval literature.
The method overview is shown in Figure~\ref{fig:network_architecture}.
It differs from CVR as follows:

\textbf{Feature descriptors:}
%Instead of a single image-global feature,
%we increase spatial resolution by building multiple image-local features for the satellite patch to better represent the local neighborhoods.
%The feature from the ground image $G$ and satellite patch $S$ are extracted by two VGG feature extractors without weight-sharing.
Instead of building one image-global descriptor to represent $S$,
we increase the top-down spatial resolution by splitting the satellite $L'\times L' \times 512$ feature volume along spatial directions into $N \times N$ sub-volumes, where $N$ is a hyper-parameter.
Now the 8 SAFA \cite{SAFA} modules are applied to each $L'/N \times L'/N \times 512$ sub-volume in parallel, resulting in an $N \times N \times 4096$ descriptor $g(S)$ for the satellite branch, shown as the green vectors in Figure~\ref{fig:network_architecture}.
Let $g(S)^{ij}$ denote the i-th row j-th column of the satellite descriptor, $1 \leq i,j < N$.
The ground image is still encoded as a single global $4096$-dimensional descriptor $f(G)$, shown as the blue vector in Figure~\ref{fig:network_architecture}.

\textbf{Fusion:}
To help distinguish different satellite image sub-regions, we compute the cosine similarity between $f(G)$ and each $g(S)^{ij}$, and use this similarity as a feature itself at this fusion bottleneck.
% before the output head.
This similarity computation results in a $N \times N \times 1$ matching score map $M$,
thus $M^{ij} = \simmatch(f(G), g(S)^{ij})$.
To complete our fusion step, the $M$ is concatenated to the satellite descriptors $g(S)$ through a skip connection, shown as the upper yellow solid arrow in Figure~\ref{fig:network_architecture}.
Optionally, one could also concatenate $f(G)$ again into the fused descriptor (yellow dashed arrow), similar to CVR; we explore this in our experiments.

\textbf{Output head:}
Rather than treating metric localization as a regression task,
we seek to generate a dense distribution over the image coordinates.
Such output enables us to represent localization ambiguities
and estimate the (un)certainty of our prediction.
Towards this, we feed the fusion volume to a decoder which can progressively up-sample the $N \times N$ matching map to higher resolutions.
Akin to the UNet architecture \cite{ronneberger2015u}, skip connections between satellite encoder and decoder are used to pass the fine-grained scene layout information to guide the decoding.
Finally, 
a softmax activation function
%\footnote{We also considered a per-pixel sigmoid activation that produces a heat map with a likelihood between 0 and 1 at each pixel. We did not find benefits from this formulation, see supplementary materials for more detail.}
is applied on
the last layer, and outputs a $L \times L \times 1$ heat map $H$, where each pixel $H^{u,v} = p(X \in c(u,v)|G, S)$ represents the probability of $G$ being located within pixel area $c(u,v)$.
This heat map is useful by itself, e.g.~in a sensor fusion framework.
For a single frame estimate, we simply output the center image coordinates $\overline{c}[\cdot]$ of the most probable pixel, i.e.~ $X = \overline{c}[\textrm{argmax}_{(u,v)} H^{u,v}]$. %where $1 \leq u, v \leq L$.

\textbf{Losses:}
A benefit of our framework is that we can add losses on both the final output and the fusion bottleneck.
The full loss $\mathcal{L} = \mathcal{L}_\text{out} + \beta \times \mathcal{L}_\text{sim}$ is thus a weighted sum of the output loss, $\mathcal{L}_\text{out}$, and the bottleneck loss, $\mathcal{L}_\text{sim}$,
where $\beta$ is a hyper-parameter.
We discuss each term next.

% For the dense prediction output head, we consider two options for the final activation, namely a softmax activation, 
Since the output $H$
is a discrete probability distribution that sums to one,
we treat our task as a multi-class classification problem. $\mathcal{L}_\text{out}$ is simply a cross-entropy loss over the $L \times L$ output cells.
% Note that, we also considered a binary classification problem with a sigmoid activation, 
%to minimize the difference between the output and ground truth probabilistic distributions,
%\begin{align}
%    \mathcal{L}_\text{out} = -\sum^{L^2}_c y_c\times\log(p_c).
%    \label{eq:CE}
%\end{align}
%In Equation~\eqref{eq:CE}, the total number of classes is ${L^2}$, and $p_c$ and $y_c$ are the predicted and ground truth probability for c-th class respectively.
The ground truth is one-hot encoded as a heat map with the same  $L \times L$ resolution and label 1 at the true location and 0 elsewhere, 
In practice, we will apply Gaussian label smoothing to the one-hot encoding of the output head, and tune the smoothing $\sigma$ as part of the hyperparameter optimization.

% TODO: are we underselling the novelty, or the importance that our architecture allows this?
To guide the model to already learn locally discriminative satellite descriptors at the fusion bottleneck,
we apply the infoNCE loss \cite{oord2018representation} from contrastive representation learning \cite{khosla2020supervised}, 
which can be seen as a generalized version of triplet loss \cite{schroff2015facenet} used in image retrieval in the case of multiple negative samples are presented at the same time,
\begin{align}
    \mathcal{L'}(ij^{+}) = -\log\frac{\exp(\simmatch(f(G), g(S)^{ij^{+}})/\tau)}{\sum_{i,j} \exp(\simmatch(f(G), g(S)^{ij})/\tau)}.
    \label{eq:infoNCE_loss}
\end{align}
Here $\tau$ is a hyper-parameter introduced by \cite{oord2018representation}, and its role is similar to the margin between positive and negative samples in triplet loss, and $(ij^{+})$ is the cell index of the positive satellite descriptor w.r.t. the ground descriptor.

We reuse the smoothed one-hot encoding from the output loss to allow multiple soft positives if the true location is near a cell border. We max-pool the $L \times L$ target map to the $N \times N$ resolution and renormalize it to generate `positiveness' weights $w^{+}_{ij}$ for each cell $1 \leq i,j \leq N$.
Our bottleneck loss is simply a weighted version of Equation~\eqref{eq:infoNCE_loss},
%\begin{align}
$\mathcal{L}_\text{sim} =
    \sum_{i,j} 
     w^{+}_{ij} \;
     \mathcal{L'} (ij)$.
%\end{align}
% The loss function, Equation~\ref{eq:loss}, takes two aspects into account.
% First, the $\mathcal{L}_\text{classification}$ formulates a classification problem between the output heat map and ground truth heat map.

% The second aspect considered by the loss $\mathcal{L}$ is the ground-to-satellite descriptor matching at the model bottleneck.

\section{Experiments}
\label{sec:experiments}
In this section, we first introduce the two datasets and evaluation metrics for our experiments.
Then we motivate each of our design choices and provide a detailed ablation study.
Finally, our model is compared to the baseline CVR approach \cite{zhu2021vigor} to show our advantage in metric localization in generalizing to new measurements in the same area, across areas, and across time.

\subsection{Datasets}
\label{sec:dataset}

The first used dataset, \textbf{VIGOR} \cite{zhu2021vigor}, contains geo-tagged ground-level panoramic images and satellite images collected in four cities in the US. 
Unlike previous cross-view image retrieval datasets \cite{CVACT,lin2013cross,tian2017cross,zhai2017predicting}, the satellite patches in VIGOR  seamlessly cover the target area.
Importantly, the ground-level panoramas are not located at the center of satellite patches.
Each satellite patch corresponds to $72.96m \times 72.96m$ ground area with a ground resolution of $0.114m$.
The orientation of the satellite patch and ground panorama are aligned in a way that the vertical line at the center of the panorama corresponds to the north direction in the satellite patch.
Typically, each patch has $\mathtt{\sim}50\%$ overlap with its neighboring patch in the North, South, East, and West direction.
This means every ground image is covered by 4 satellite patches. If the ground image is at the center 1/4 area of a satellite patch, the patch is denoted as ``positive'', otherwise ``semi-positive''.
In practice, ``positive'' samples simulate the case that the global localization prior is more accurate, e.g. error $< \sqrt{2} \times 18.24m$ in the case of VIGOR.
%If we crop a patch at the prior location from a continuous satellite image, the ground image would locate in the central area of the cropped patch.
Similarly, ``positive + semi-positive'' samples would be a result of a coarser localization prior, e.g. error $< \sqrt{2} \times 36.48m$.
During training, we include both ``positives'' and ``semi-positives'' samples.
Our main evaluation will be based on positive samples, since it is representative for most real-world situations, e.g. localization prior from GNSS positioning in an open area or temporal filtering.
For completeness, we also evaluate on ``positive + semi-positive'', to showcase how the methods behave with a less certain localization prior, e.g. GNSS positioning in an urban canyon.
% We adopt two train-test splits from \cite{zhu2021vigor}.
% To test generalization to images at different locations in the same cities, 
% the ``same-area'' split uses half of the images for training and the other half for testing in all four cities. The ``cross-area'' split has New York and Seattle in the training set, while the test images are from San Francisco and Chicago.
% To find one set of hyper-parameters for both ``same-area'' and ``cross-area'',
% we create a subset of the shared training data from New York as a smaller ``tuning'' split with 11108/2777 training/validation samples.
We adopt the ``same-area'' and ``cross-area'' splits from \cite{zhu2021vigor} to test the model's generalization in the same cities and across different cities.
To find one set of hyper-parameters for both ``same-area'' and ``cross-area'',
we create a subset of the shared training data from New York as a smaller ``tuning'' split with 11108/2777 training/validation samples.

The second dataset, \textbf{Oxford RobotCar} \cite{OxfordRobotCar1,OxfordRobotCar2} contains multi-sensor measurements from multiple traversals over a consistent route through Oxford collected over a year.
The original dataset does not contain satellite images.
To enable cross-view metric localization, we stitch the satellite patches provided by \cite{Geolocal_feature,9449965} and our additionally collected ones to create a continuous satellite map that covers the target area. 
We follow the same data split as in \cite{9449965} to test how our method generalizes to new ground images collected at different time.
In total, there are 17067, 1698, and 5089 ground-level front-viewing images in the training, validation, and test set respectively.
The test set contains 3 traversals collected at later times of day than the training recordings.
Benefiting from a full continuous satellite map, we randomly and uniformly sample satellite patches around the ground image locations during training for data augmentation.
Each patch is rotationally aligned with the view direction of the ground image and has a resolution of $800 pixels \times 800 pixels$, which corresponds to $73.92m \times 73.92m$ on the ground.
% A fixed set of satellite patches with 50\% overlapping is used for validation and testing, and we select the patch whose central 1/4 area covers the ground image to avoid lacking mutual information between the satellite and ground front-facing views.
A fixed set of satellite patches is used for validation and testing.
Each patch has a 50\% area overlap with the closest neighboring patches.
We pair each ground image with the patch at the closest center location to allow for mutual information between the satellite and ground front-facing views.

Similarly, during training, we also control the sampled locations to make sure the ground image locates inside the central area of the sampled satellite patch.

\subsection{Evaluation metrics}
\label{sec:experiments_metrics}
To measure the localization error, we report the mean and median distance between the predicted location and ground truth location in meters over all samples.
Note that the mean error can be biased by a few samples with large error, and including the median error provides a measurement more robust w.r.t. outliers.
% To make the comparison fair, we train the baseline VIGOR \cite{zhu2021vigor} with offset regression only.
% During the evaluation, we always provide the ground image with the satellite patch whose center area covers the ground image.
In practice, a localization method that operates on a single image frame can be extended to process a sequence of data using a Bayesian filter \cite{CVMJournal,9449965}.
In such a setting, the estimated probability at the ground truth location plays an important role in accurately localizing over the whole path.
Motivated by this, we include the probability at the ground truth pixel area as an additional metric.
% For our method, we first normalize the output likelihood over the heat map to acquire probability distribution.
The baseline CVR method does not have any probability estimation on its output.
Hence, we post-process the baseline output by assuming the regressed location is the mean of an isotropic Gaussian distribution, and we estimate the standard deviation of this Gauss on the validation set.

\subsection{Hyper-parameters and ablation study} 
\label{sec:hypertune}
We first discuss our hyper-parameter choices, then investigate the main components in our proposed architecture.
The weight $\beta$ in our loss function is set to $10^4$, and the $\tau$ in infoNCE loss is set to $0.1$, as done in \cite{oord2018representation}.
The loss is optimized by Adam optimizer \cite{Adam} with a learning rate of $1 \times 10^{-5}$, and the VGG feature extractors are pre-trained on ImageNet \cite{deng2009imagenet}.
Our main model variations and hyper-parameters are now compared on the VIGOR ``tuning'' split.% (see Section~\ref{sec:dataset}).

We initially set $N=8$ for the matching at the bottleneck.
%Instead of using one-hot encoded ground truth heat map, we perform label smoothing on the heat map by putting a Gaussian distribution with std.dev. $\sigma$ at the ground truth location.
The infoNCE loss at the bottleneck is key to improve the final model output.
The model trained with it achieves a much better mean error, $14.30m$, than the model trained without it, $19.25m$.
Label smoothing with $\sigma = 4$ pixels further reduces the mean error to $13.39m$ (we tested $\sigma = 1, 2, 4, 8$).
We use both in all future experiments.
% This validates the effectiveness of our metric learning component.

Next, we study the influence of different resolutions $N \times N$ at the model bottleneck.
When $N=1,2,4,\textbf{8},16$, the mean error is $19.62, 15.98, 15.23, \textbf{13.39}, 15.04$ meters respectively (best in bold).
With $N=1$ no infoNCE loss is applied, and the decoder receives a single matching score concatenated with features from the satellite branch.
Increasing $N$ improves the spatial resolution at the model bottleneck.
However, with larger $N$ the decoder also operates on larger inputs but with fewer upsampling layers.
We observe a balance at $N=8$.
% , which delivers the best results.
% \begin{table} 
% \centering
% \caption{Error on tuning split with $N \times N$ similarity matching, and satellite descriptors, at the fusion bottleneck. Best in bold}
% \label{table:matching_resolution}
% \begin{tabular}{ |p{2.2cm}|p{1cm}|p{1cm}|p{1cm}|p{1cm}|p{1cm}|} 
% \hline
% $N$ & 1 & 2 & 4 & 8 & 16\\
% \hline
% % Binary & 19.65 & 16.08 & 14.64 & \textbf{14.06} & 15.55 \\
% % \hline
% Mean error (m)& 19.62 & 15.98 & 15.23 & \textbf{13.39} & 15.04\\
% \hline
% \end{tabular}
% \end{table}

To further explore the role of metric learning at the bottleneck and the feature concatenation, we create four extra model variations in Table~\ref{table:matching_vs_concat}.
Directly concatenating the ground with a single global satellite descriptor (see ``1,S+G'') is akin to CVR's fusion with a decoder head instead of regression,
but this change alone does not perform well.
The model does not work when the decoder
operates on only a single channel map (``8,M'') without any context from the satellite patch.
% Besides, when $N=8$, we observe that the model overfits quickly given the relatively small amount of the training data in our ``tuning'' set, 
Increasing the satellite resolution is also still insufficient with only ground descriptors (``8,S+G''), the descriptors must also be trained to be locally discriminative.
Interestingly, we do not observe any benefit from also concatenating the ground descriptor (``8,S+M+G'') 
to our default of satellite descriptors with matching scores (``8,S+M'').
% , presumably due to overfitting.
%Our proposed matching scores in the bottleneck already capture most relevant information.
In all next experiments, we fuse only  the satellite descriptors and the matching score.
% This shows the merit of explicitly supervising the model to compare features from two branches and then progressively up-sampling the resulting matching score into a full-resolution localization heat map.

\begin{table}[t]
\centering
\caption{ Fusion bottleneck, error on tuning split: ``S'' stands for satellite descriptors $g(S)$, ``G'' for ground descriptors $f(G)$, ``M'' for  cosine similarity feature. Best in bold}
\label{table:matching_vs_concat}
\begin{tabular}{ |p{2.5cm}|p{1.2cm}|p{1.2cm}|p{1.2cm}|p{1.8cm}|p{1.2cm}|} 
\hline
$N$, descriptors & 1,S+G & 8,M & 8,S+G & 8,S+M+G & 8,S+M \\
\hline
Mean error (m) & 18.62 & 24.37 & 18.35 & 13.83 & \textbf{13.39}\\
\hline
\end{tabular}
\end{table}

%We measured the runtime of our best performing model and CVR. 
We note that to forward-pass an input pair from VIGOR on a Tesla V100 GPU, CVR uses 0.020s, and our best-performing model 0.034s (i.e. $\sim30$ FPS).

\subsection{Generalization in the same area / across areas}

We now compare our method to CVR on the VIGOR splits for generalizing to unseen ground images inside the same area, and across areas.
When tested on the ``same-area'' correctly retrieved samples, CVR trained for only regression has a better mean (-1.5m) and median (-1.1m) localization error than CVR trained for both retrieval and regression, as expected.
% From now on, we will always use the CVR model trained for regression only as the baseline.
From now on, we will always train CVR model for regression only as the baseline.

% \begin{table} 
% \centering
% \begin{tabular}{ |p{1.8cm}|p{1cm}|p{1cm}|p{1cm}|p{1cm}|} 
% \hline
% Same-area & \multicolumn{2}{|c|}{Positives} & \multicolumn{2}{|c|}{Semi-positives}\\
% \hline
% Error (m) & Mean & Median & Mean & Median\\
% \hline
% Oracle R. & 14.15 & 14.82 & 32.33 & 32.39 \\
% \hline
% CVR \cite{zhu2021vigor} & 10.55 & 9.31 & 18.66 & 16.31\\
% \hline
% Ours & \textbf{9.86} & \textbf{4.58} & \textbf{14.65} & \textbf{5.70} \\
% \hline
% \hline
% Cross-area & \multicolumn{2}{|c|}{Positives} & \multicolumn{2}{|c|}{Semi-positives}\\
% \hline
% Error (m) & Mean & Median & Mean & Median\\
% \hline
% Oracle R. & 14.07 & 14.07 & 32.38 & 32.44 \\
% \hline
% CVR & \textbf{11.26} & 10.02 & 21.13 & 20.03\\
% \hline
% Ours & 13.06 & \textbf{6.31} & \textbf{18.49} & \textbf{8.19}\\
% \hline
% \end{tabular}
% \caption{Mean and median localization error (best results in bold). ``Retrieval'' means an oracle cross-view image retrieval model. The term ``Positives'' means we input a positive satellite patch together with the ground image. ``Pos.+Semi.'' means we take the mean over the results from ground image with the positive satellite image and three semi-positive satellite patches.}
% \label{table:VIGOR_test}
% \end{table}

\begin{table}[t]
\centering
\caption{Localization error on VIGOR. Best in bold. ``Center-only'' denotes using satellite patch center as the  prediction. The term ``Positives'' stands for evaluation on positive satellite patches. ``Pos.+semi-pos.'' takes the mean over the results from the positive satellite patches and all semi-positive satellite patches}
\label{table:VIGOR_test}
\begin{tabular}{|p{1.8cm}|p{1.1cm}|p{1.1cm}|p{1.1cm}|p{1.1cm}||p{1.1cm}|p{1.1cm}|p{1.1cm}|p{1.1cm}|} 
\hline
& \multicolumn{4}{|c||}{\textit{Same-area}} & \multicolumn{4}{|c|}{\textit{Cross-area}} \\
\cline{2-9}
& \multicolumn{2}{|c|}{Positives} & \multicolumn{2}{|c||}{Pos.+semi-pos.} &  \multicolumn{2}{|c|}{Positives} & \multicolumn{2}{|c|}{Pos.+semi-pos.}\\
\hline
\textit{Error (m)} & Mean & Median & Mean & Median & Mean & Median & Mean & Median\\
\hline
Center-only & 14.15 & 14.82 & 27.78 & 28.85 & 14.07 & 14.07 & 27.80 & 28.89 \\
\hline
CVR \cite{zhu2021vigor} & 10.55 & 9.31 & 16.64 & 13.82 & \textbf{11.26} & 10.02 & 18.66 & 16.73\\
\hline
Ours & \textbf{9.86} & \textbf{4.58} & \textbf{13.45} & \textbf{5.39} & 13.06 & \textbf{6.31} & \textbf{17.13} & \textbf{7.78}\\
\hline
\end{tabular}
\end{table}

% \begin{table} 
% \centering
% \begin{tabular}{ |p{1.8cm}|p{1cm}|p{1cm}|p{1cm}|p{1cm}|} 
% \hline
% Pos. + semi. & \multicolumn{2}{|c|}{Same-area} & \multicolumn{2}{|c|}{Cross-area}\\
% \hline
% Error (m) & Mean & Median & Mean & Median\\
% \hline
% Oracle R. & 27.78 & 28.85 & 27.80 & 28.89 \\
% \hline
% CVR \cite{zhu2021vigor} & 16.64 & 13.82 & 18.66 & 16.73\\
% \hline
% Ours & \textbf{13.45} & \textbf{5.39} & \textbf{17.13} & \textbf{7.78} \\
% \hline
% \end{tabular}
% \caption{Mean and median localization error (best results in bold). ``Retrieval'' means an oracle cross-view image retrieval model. Each ground image is compared to all its positive and semi-positives.}
% \label{table:VIGOR_test_all_pos}
% \end{table}

% \begin{figure}[t]
%     \centering
%     % \hspace*{-0.5em}
%     \includegraphics[height=0.14\textwidth]{figures/experiments/errorplots_pos.pdf}\hspace*{-0.5em}
%     \includegraphics[height=0.14\textwidth]{figures/experiments/errorplots_pos_semipos.pdf}\hspace*{-0.5em}
%     \includegraphics[height=0.14\textwidth]{figures/experiments/errorplots_pos_cumulative.pdf}\hspace*{-0.5em}
%     \includegraphics[height=0.14\textwidth]{figures/experiments/errorplots_pos_semipos_cumulative.pdf}
%     \caption{Error distributions (plots 1,2: regular, 3,4: cumulative, 1,3: positives, 2,4: positives and semi-positives) on VIGOR, for same-area and cross-area experiments.
%     }
%     \label{fig:errorplots}
% \end{figure}

\begin{figure}[t]
    \centering
    % \hspace*{-0.5em}
    \includegraphics[width=1.0\textwidth]{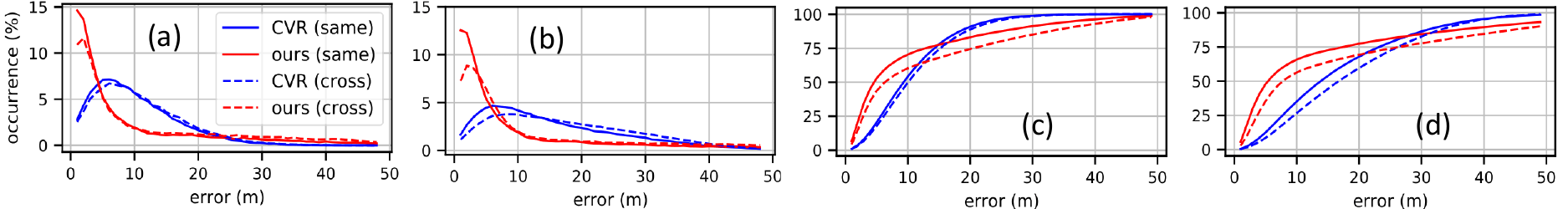}
    \caption{Error distributions (plots a,b: regular, c,d: cumulative, a,c: positives, b,d: positives and semi-positives) on VIGOR, for same-area and cross-area experiments.
    }
    \label{fig:errorplots}
\end{figure}

\textbf{Metric error:~}
The quantitative comparison against CVR on both VIGOR splits  is summarized in Table~\ref{table:VIGOR_test}.
To highlight the value of conducting cross-view metric localization, we also include a ``center-only'' prediction
which always outputs $X = (0.5, 0.5)$.
%which sets a lower bound on the localization error for all retrieval-only methods on this dataset.
Note that retrieval-only methods typically assume that the center of a satellite patch is representative of the true location.
%and the best retrieval-only methods can achieve is for every query retrieving the center of the correct satellite patch.
%By comparing the  to CVR and our model, we show that conducting  cross-view metric localization improves the localization accuracy on top of cross-view image retrieval both within the same area and across areas.

%Moreover, all performance gain does not require extra training data.
When the ground image is compared to the positive-only satellite patches, our model reduces the median error by 51\% over CVR when generalizing within the same area ($4.58m$ vs $9.31m$), and by 37\% when generalizing across areas ($6.31m$ vs $10.02m$).
Generally, our model improves over the baselines, but across areas, our mean error is higher than that of CVR.
The error (cumulative) distribution in Figure~\ref{fig:errorplots} confirms that this is 
due to a  few large-error outliers in our prediction.
These outliers are a result of selecting a wrong mode, or of large uncertainty in our multi-modal output, whereas regression might pick an averaged location in the middle resulting in neither small nor very large errors.
We will show below that our location's probability can be used to detect such potential large error cases.
This would aid an external sensor fusion module, which can also directly integrate the distribution to reduce the uncertainty through other measurements.

When ground images could be located further from the center, as in the ``positive+semi-positive'' test cases, there is less matchable visual information between the two views.
In this case, the performance of both CVR and our model somewhat degenerates, though our model suffers less than CVR.
Our mean and median errors are lower than CVR's, both within the same area and across areas.
Moreover, our method's advantage in the median error further increases.
% Besides, since the ground truth locations are no longer in the center, CVR might benefit less from regressing to a central location.
In the Sup. Mat. we furthermore investigate the effect of using a CVR-like regression layer and loss on top of our dense output but find it hurts median performance.

\begin{figure*}[t]
    \centering
    \includegraphics[width=0.155\textwidth]{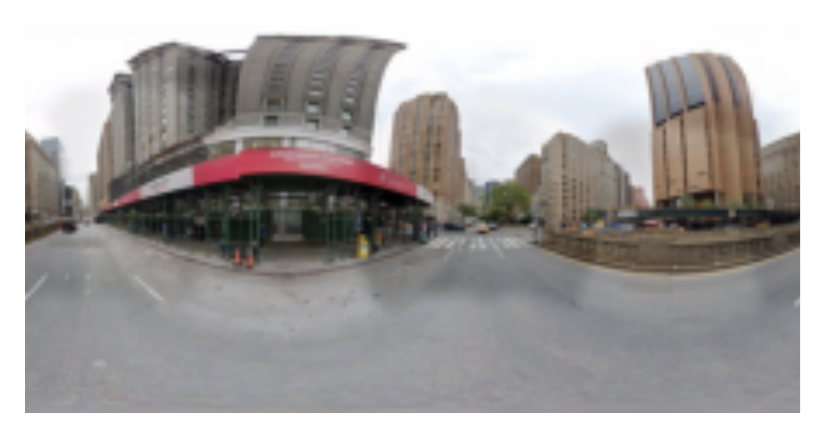}
    \includegraphics[width=0.075\textwidth]{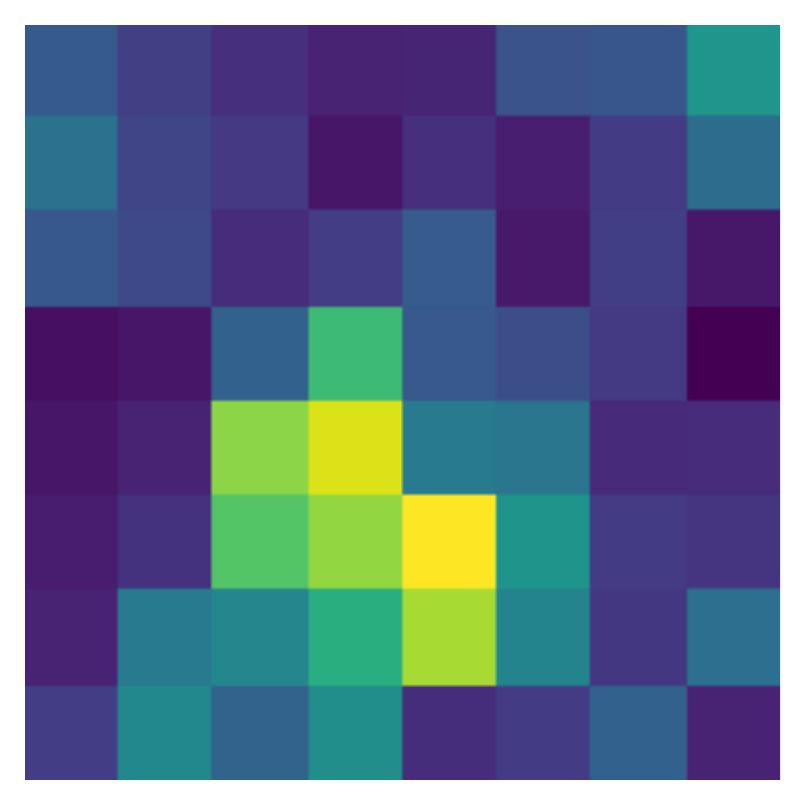}
    \includegraphics[width=0.155\textwidth]{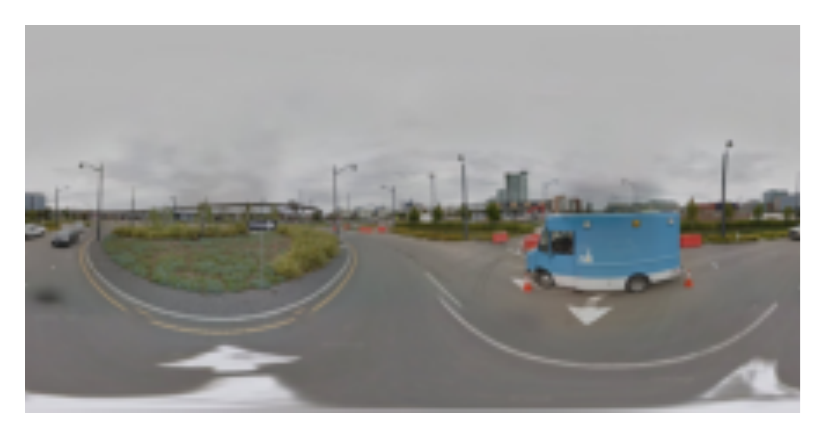}
     \includegraphics[width=0.075\textwidth]{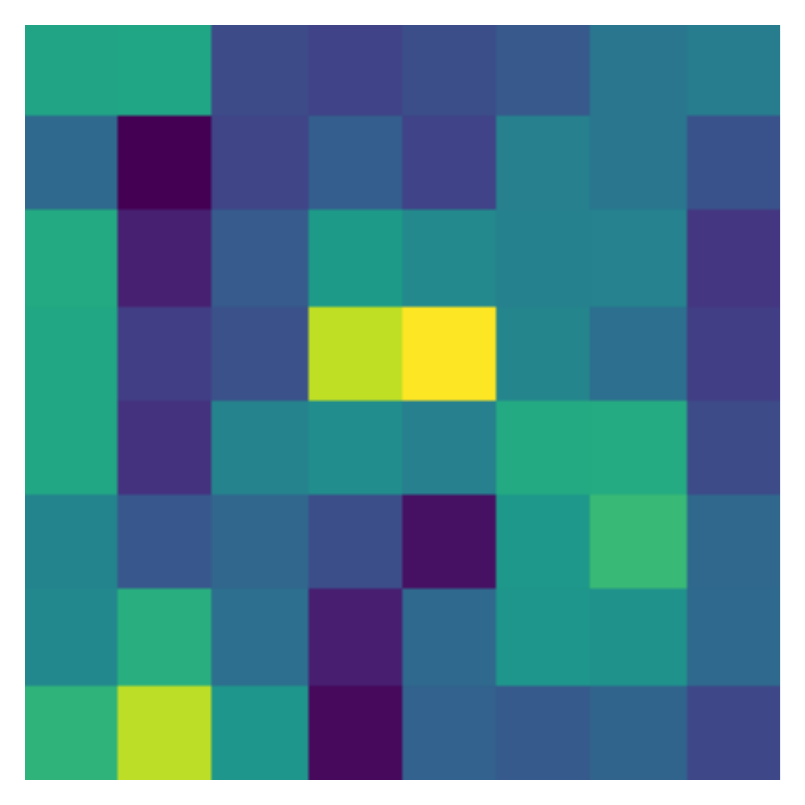}
    \includegraphics[width=0.155\textwidth]{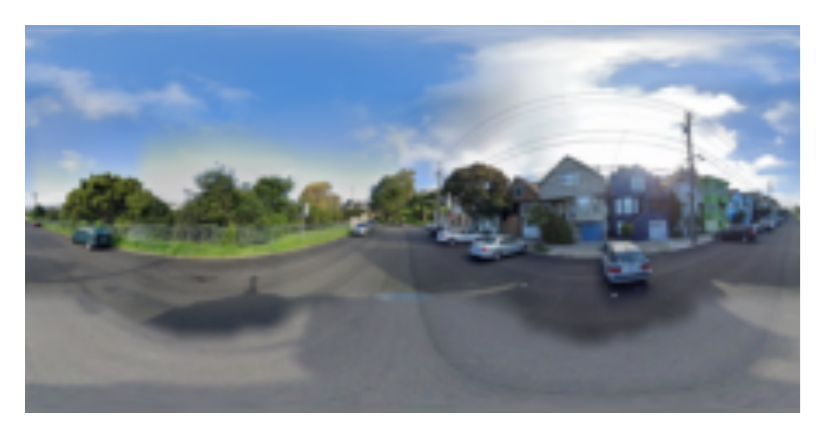}
     \includegraphics[width=0.075\textwidth]{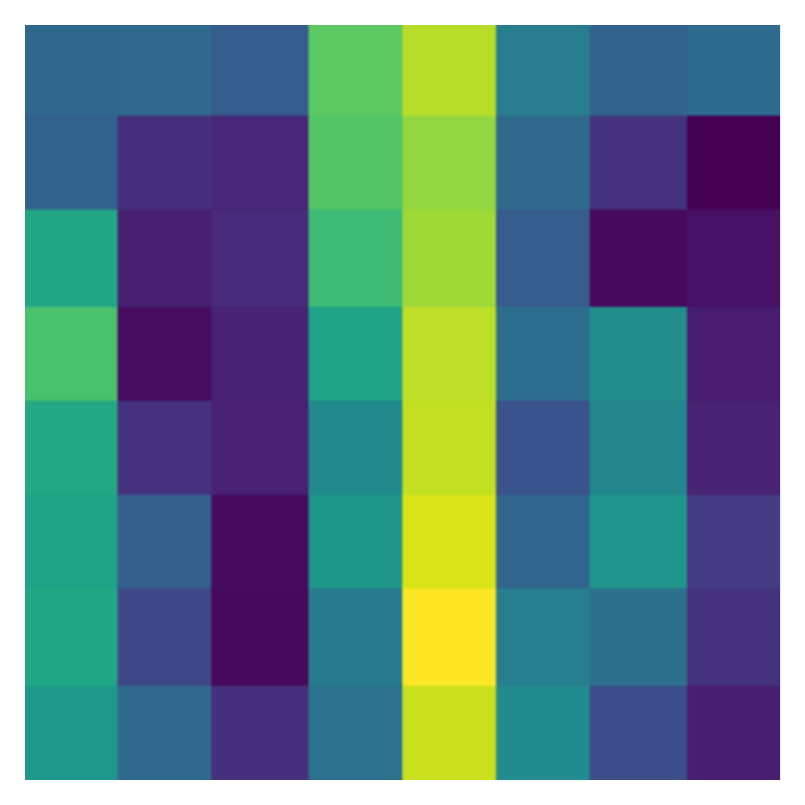}
    \includegraphics[width=0.155\textwidth]{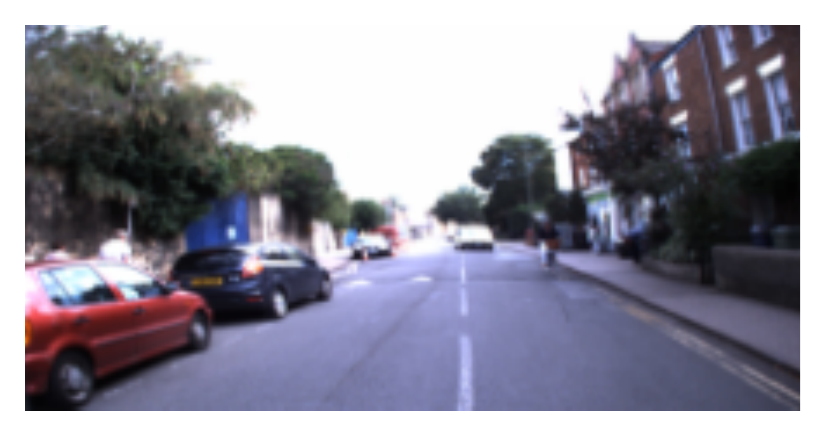}
     \includegraphics[width=0.075\textwidth]{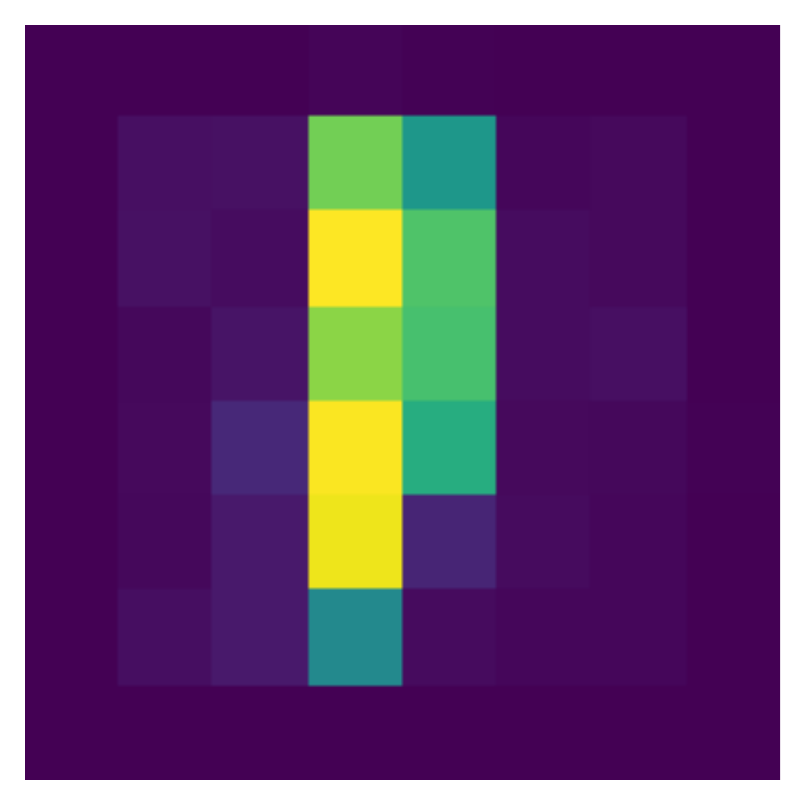}
    \\
    \includegraphics[width=0.24\textwidth]{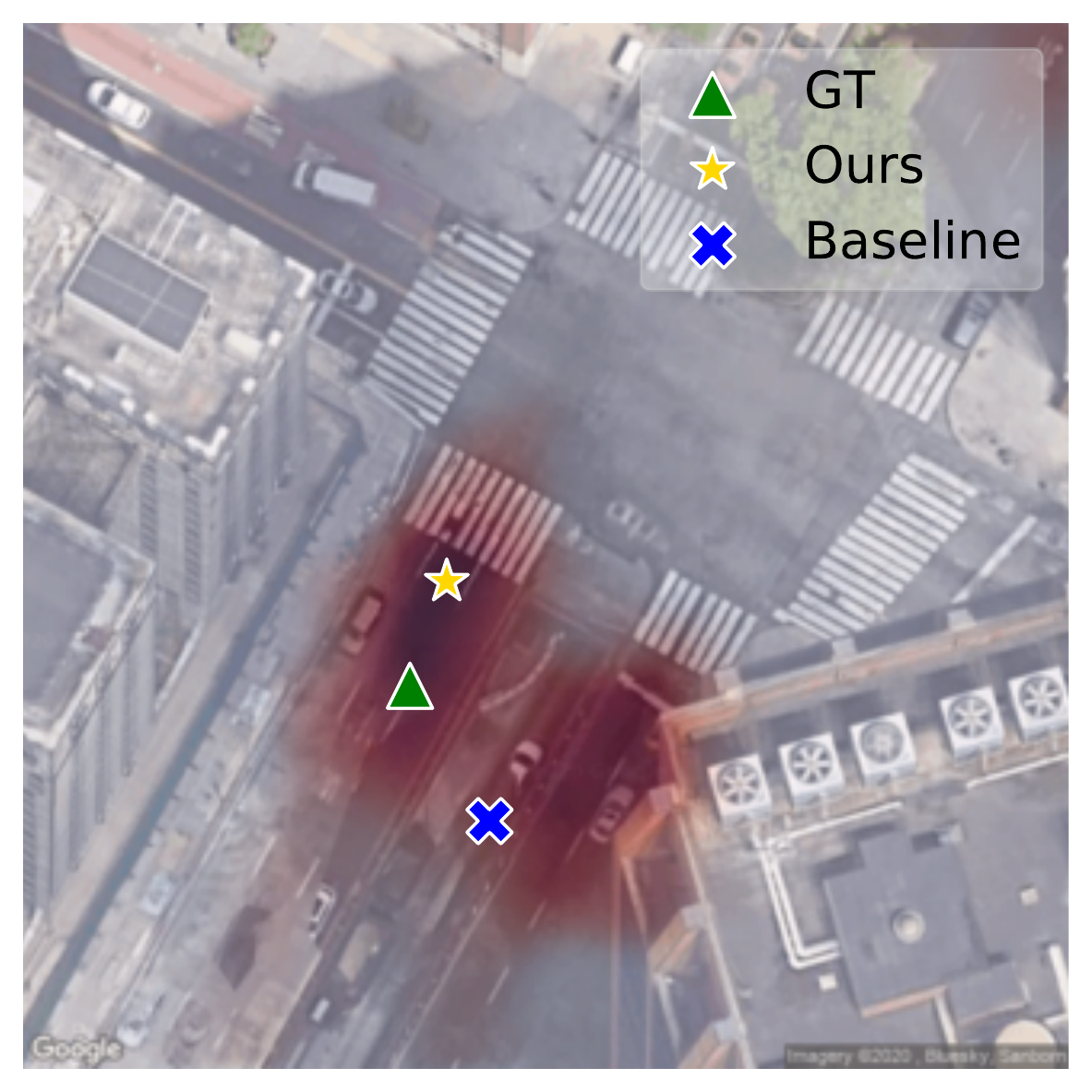}
    \includegraphics[width=0.24\textwidth]{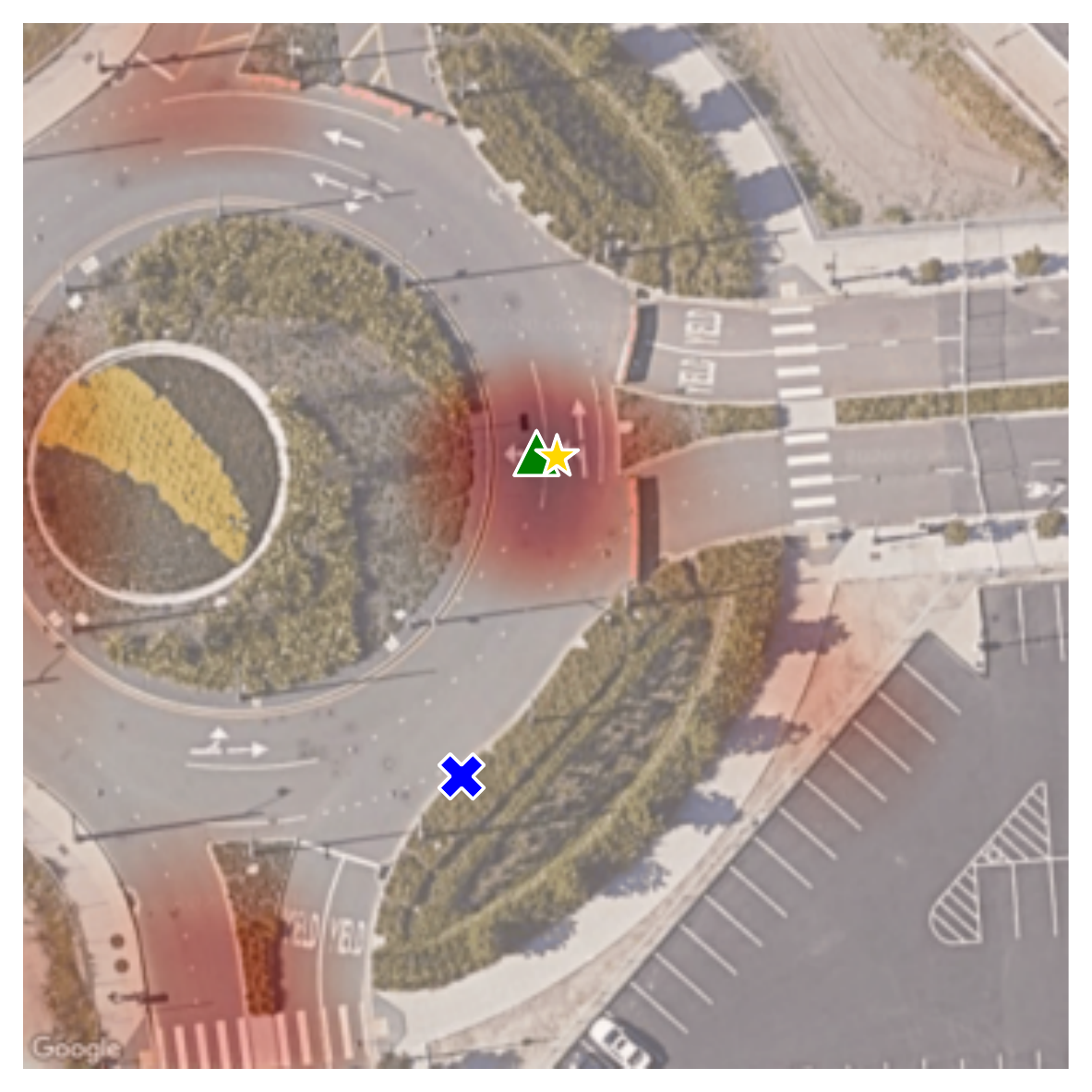}
    \includegraphics[width=0.24\textwidth]{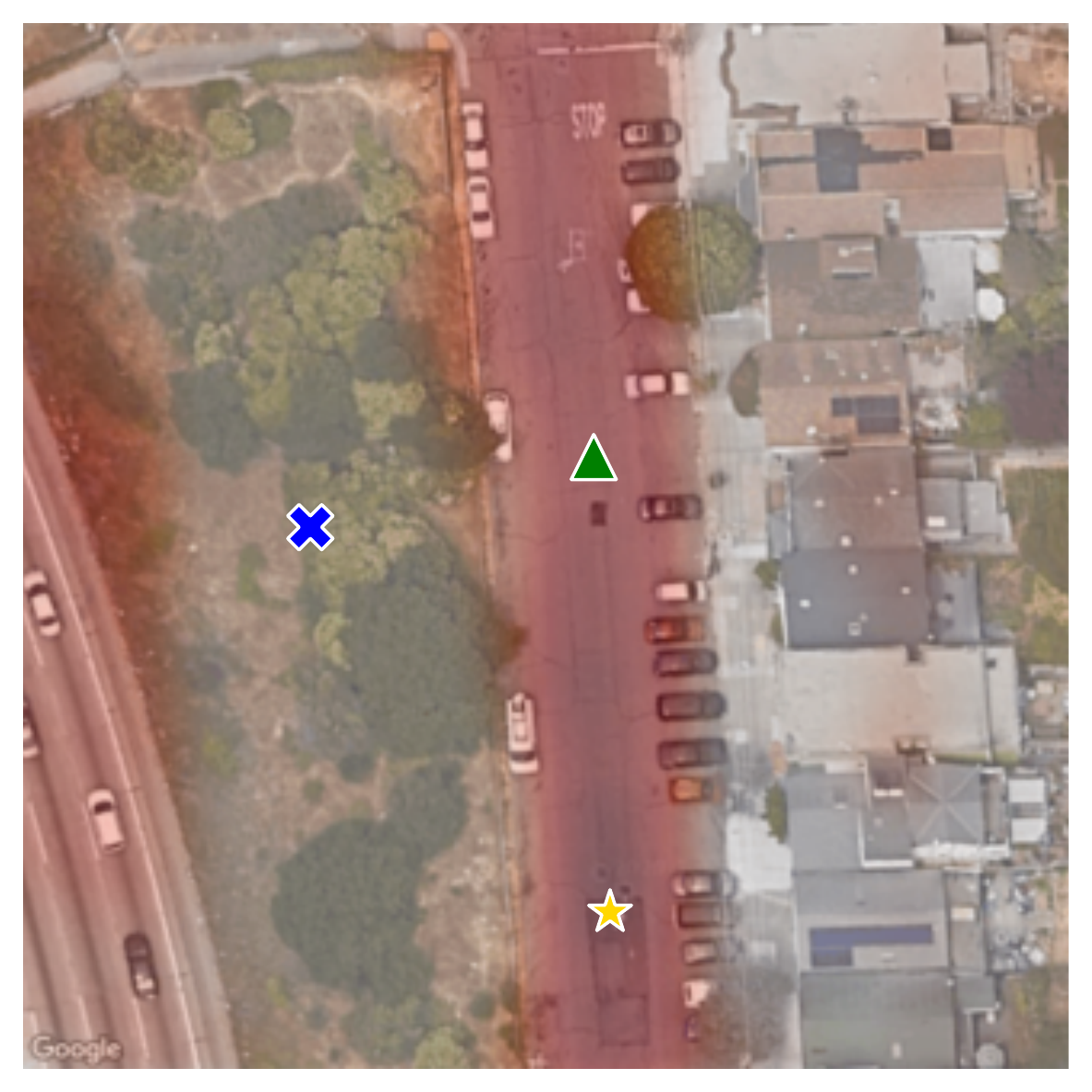}
    \includegraphics[width=0.24\textwidth]{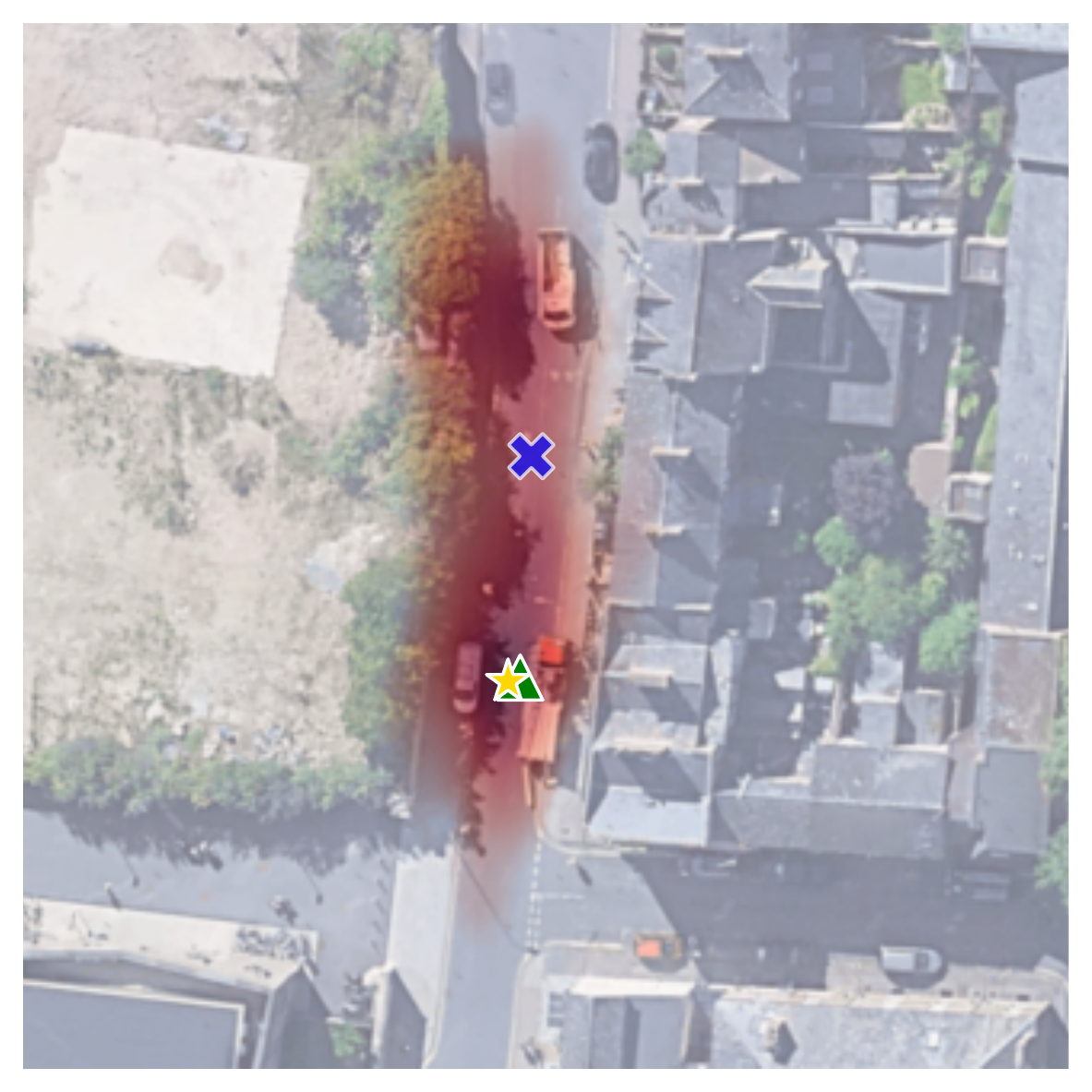}
    \caption{Top: input ground images and matching score maps at the model bottleneck, bottom: input satellite image overlayed with outputs from CVR and our method. From left to right: 1: VIGOR, same-area, 2,3: VIGOR: cross-area, 4: Oxford RobotCar.
    % Best viewed in color.
    }
    \label{fig:qualitative_results}
\end{figure*}

\textbf{Qualitative results:~}
To intuitively understand where our advantage comes from, we provide qualitative examples of success and failure cases in Figure~\ref{fig:qualitative_results}.
In the context of image-based cross-view metric localization, there can exist multiple visually similar locations on the satellite image given a ground image. 
In such cases, it is important for the model to have the capability to express the underlying localization uncertainty.
In our model, the uncertainty is already present at the model bottleneck, see Figure~\ref{fig:qualitative_results} top row. This distribution is upsampled by the decoder and aligned with the observed environmental features, such as roads and crossings, resulting in the dense multi-modal uncertainty map.
We emphasize that no explicit semantic map information, e.g. on road layout, was used during training.
%Still, our method learns to identify road surfaces as high-probability locations for the camera to be located.
Since the regression-based baseline method forces the output to be a single location, it risks `averaging' multiple similar locations and provides a wrong final estimate without any uncertainty information.

We argue that in practice our outliers are still more acceptable than CVR's errors.
When our model is uncertain about the exact location, our output heat map can be rather homogeneous.
As shown in Figure~\ref{fig:qualitative_results} example 3, given a ground image taken on the road, our model assigns high probability to roads in the center and on the left.
In this situation, the distance between our predicted location and the ground truth can be large.
Instead, CVR tends to output the average between the visually similar areas, which can result in a location near the center but that is intuitively unreasonable, e.g. within some vegetation, even though it may have a smaller distance to the ground truth location.

% \begin{table} 
% \centering
% \begin{tabular}{ |p{1.1cm}|p{1.25cm}|p{1.25cm}|p{1.25cm}|p{1.25cm}|} 
% \hline
% \multicolumn{5}{|c|}{Same-area, on positive satellite patches} \\
% \hline
% Range & Top 40\% & Top 30\% & Top 20\% & Top 10\%\\
% \hline
% Mean & 4.74 & 4.14 & 3.66 & 3.29\\
% \hline
% Median & 2.62 & 2.43 & 2.25 & 2.14\\
% \hline
% \multicolumn{5}{|c|}{Cross-area, on positive satellite patches} \\
% \hline
% Range & Top 40\% & Top 30\% & Top 20\% & Top 10\%\\
% \hline
% Mean & 6.87 & 5.93 & 5.10 & 4.07\\
% \hline
% Median & 3.25 & 2.97 & 2.71 & 2.44\\
% \hline
% \end{tabular}
% \caption{Given a range (top X\%) of probability, we summarize the mean and median error (m) over the samples within that range.}
% \label{table:VIGOR_likelihood}
% \end{table}

\begin{figure}[t]
    \centering
    % \hspace*{-0.5em}
    \includegraphics[width=0.7\textwidth]{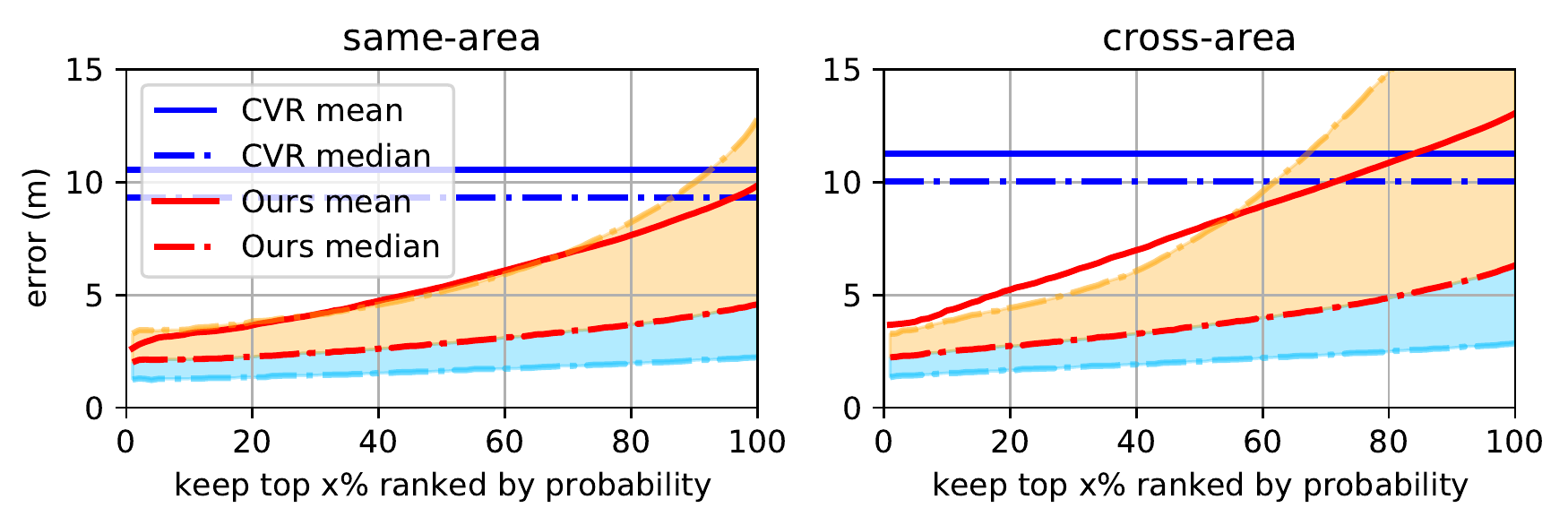}\\
    \caption{Ranking the predictions using their probabilities on VIGOR ``positives''. Red lines show our error statistics over the top $x\%$.  Cyan/orange: error between median and 25\% / 75\% quantile line. As $x$ decreases, only the more probable predictions are kept.
    Blue lines: CVR cannot rank predictions this way.
    }
    \label{fig:prop_error}
\end{figure}

\textbf{Probability evaluation:~}
Apart from the metric localization error, we will compare how well each model can predict the probability at the ground truth location.
For CVR we estimate this assuming a fixed Gaussian error distribution, see Section~\ref{sec:experiments_metrics}.
Table~\ref{table:probability} reports both mean and median probabilities at the ground truth pixel.
Our multi-modal approach outperforms the fixed error distribution for CVR.
Importantly, the probability at our predicted location (the maximum in $H$) is correlated to its localization error.
If we apply a rejection threshold to only keep the top $x$\%, predictions, we can reduce the expected error. See Figure~\ref{fig:prop_error}
with the statistics over the top-ranked estimates.
%This property is desired as in practice we can potentially filter out bad predictions by thresholding the predicted probability.
These properties are beneficial when the single frame localization results are temporally filtered or fused with other sensors.

\begin{table} 
\centering
\caption{Probabilities at the ground truth pixel on VIGOR. Best in bold. The magnitude of the probabilities is low due to the normalization over the $512 \times 512$ grid. ``Uniform'' shows for reference the prob. at GT for a homogeneous map, $1/(512 \times 512)$}
\label{table:probability}
\begin{tabular}{ |p{2cm}|p{2cm}|p{2cm}|p{2cm}|p{2cm}|} 
\hline
\textit{Prob.~at GT,}
& \multicolumn{2}{|c|}{Same-area} & \multicolumn{2}{|c|}{Cross-area}\\
\cline{2-5} % unlike \hline, this skips the first cell
\textit{Positives} & Mean & Median & Mean & Median\\
\hline
Uniform & $3.81\times10^{-6}$ & $3.81\times10^{-6}$ & $3.81\times10^{-6}$ & $3.81\times10^{-6}$\\
\hline
CVR \cite{zhu2021vigor} & $1.55\times10^{-5}$ & $1.70\times10^{-5}$ & $1.57\times10^{-5}$ & $1.72\times10^{-5}$\\
\hline
% Ours-binary & 3.50e-5 & 3.14e-5 & 3.18e-5 & 2.85e-5\\
% \hline
Ours & $\mathbf{2.93\times10^{-4}}$ & $\mathbf{1.17\times10^{-4}}$ & $\mathbf{1.54\times10^{-4}}$ & $\mathbf{7.06\times10^{-5}}$ \\
\hline
\end{tabular}
\end{table}

\begin{figure}[t]
    \centering
    \includegraphics[height=0.225\textwidth]{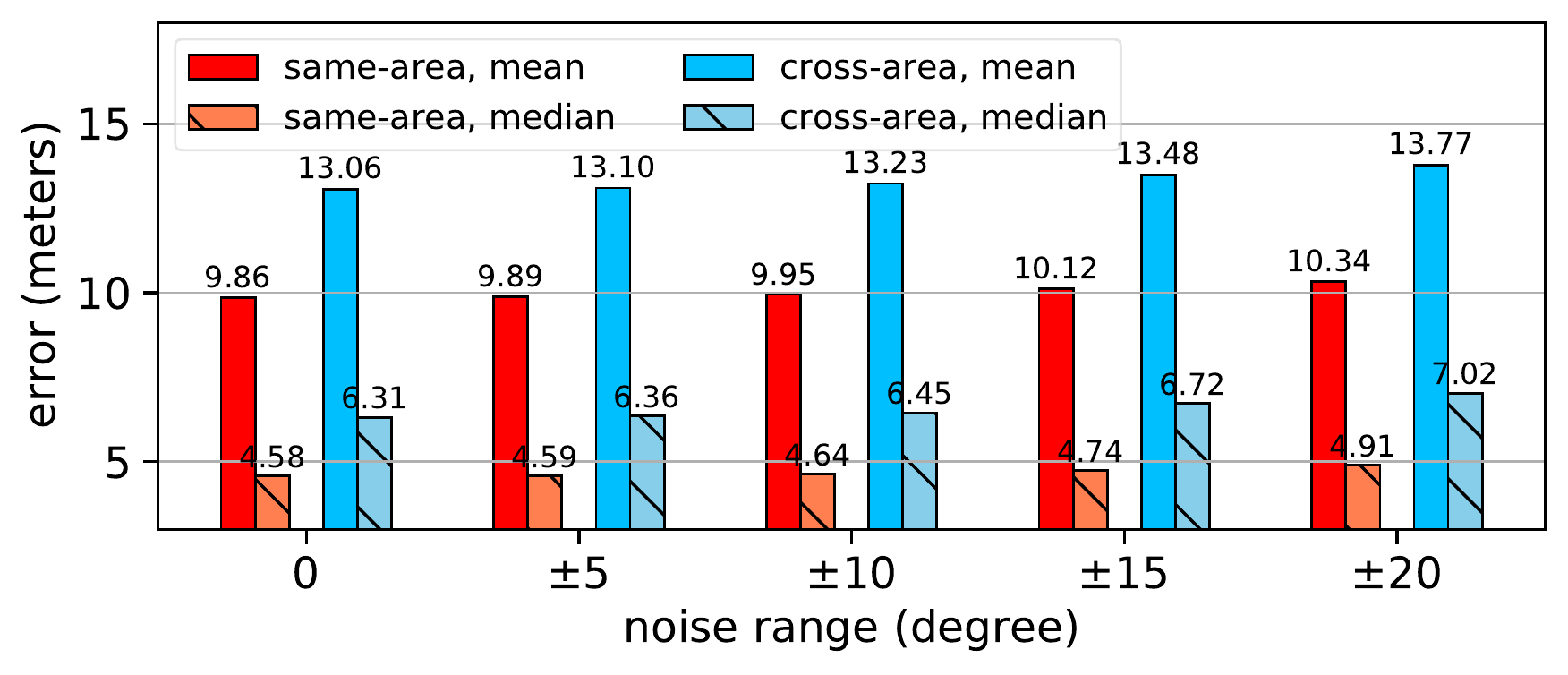}
    \includegraphics[height=0.225\textwidth]{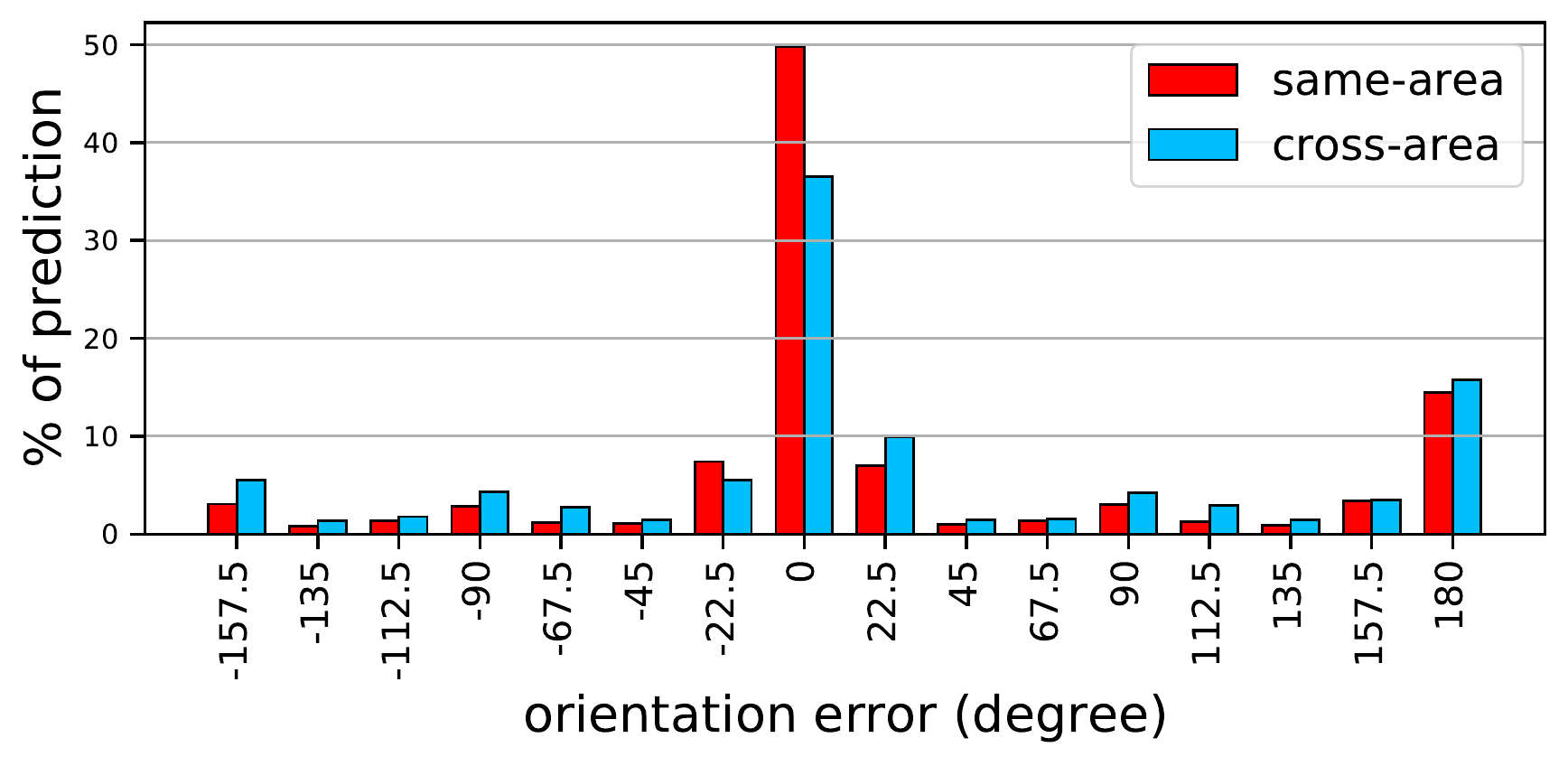}
    \caption{Left: robustness of our model against small perturbations in orientation. Right: directly using our model to infer the unknown orientation.
    }
    \label{fig:orientation}
\end{figure}

\textbf{Orientation:~}
Till now, we have relied on a known orientation during test time,
e.g.~estimated in the preceding retrieval step \cite{zhu2021revisiting,shi2020looking,LocOriStreet} or by the sensor stack~\cite{won2015performance}.
We here study our model's robustness against orientation perturbations and
ability to infer orientation when it is unknown \textit{without retraining}.
To test robustness, we uniformly sample angular noise from a range up to $\pm 20^{\circ}$ \cite{shi2020optimal} to horizontally shift the ground panoramas (i.e. ``rotate'' the heading) at test time.
% In Table~\ref{table:perturbed_orientation}, we report our model performance on those samples.
% Additionally, we include a model trained with such perturbed orientation to show the effect of using noisy orientation as data augmentation.
As shown in Figure~\ref{fig:orientation} left, the predicted location of our model remains stable under such noise. 
% Besides, augmenting the training data with $<\pm20^{\circ}$ makes the model more robust against this, but not for
% \begin{table} 
% \centering
% \caption{Model performance with perturbed orientation}
% \label{table:perturbed_orientation}
% \begin{tabular}{ |p{2cm}|p{2cm}|p{2cm}|p{2cm}|p{2cm}|} %
% \hline
% \textit{Error (m)}
% & \multicolumn{2}{|c|}{Same-area, positives} & \multicolumn{2}{|c|}{Cross-area, positives}\\
% \cline{2-5} % unlike \hline, this skips the first cell
% \textit{Noise range $^{\circ}$} & Mean & Median & Mean & Median\\
% \hline
% [-5, 5] & 9.89 & 4.59  & 13.10 & 6.36 \\
% \hline
% [-10, 10] & 9.95  & 4.64  & 13.23  & 6.45 \\
% \hline
% [-15, 15] & 10.12 & 4.74  & 13.48 & 6.72 \\
% \hline
% [-20, 20] & 10.34 & 4.91  & 13.77 & 7.02 \\
% \hline
% \end{tabular}
% \end{table}
%

Still, the model's confidence is not invariant to orientation shift,
as our prediction confidence
can help classify a ground panorama's  unknown orientation.
We rotate the ground panorama by multiples of $22.5^{\circ}$ up to $360^\circ$, apply our model to each rotated panorama with the satellite patch, and collect all 16 activation maps before the final softmax operation.
The classification output is the orientation of the map with the highest activation.
As shown in Figure~\ref{fig:orientation} right, for same and across areas, our model correctly classifies 50\% and 37\%  samples into the true orientation class out of 16 classes.
Most erroneous predictions have an error of $180^{\circ}$,
corresponding to the opposite driving direction.
\subsection{Generalization across time}
Finally, we test how our method generalizes to new measurements collected at different times and days on the Oxford RobotCar dataset.
%Unlike the VIGOR dataset where the ground images are located sparsely, here the training ground images are densely distributed over the route, and the main variation between training and test ground images is the lighting condition and weather.
% Again, we train the CVR model with the regression part only as our metric localization baseline. 
%Additional to the ``center-only'' estimate that operates on a set of satellite patches with 50\% overlap, 
For a comparison to cross-view image retrieval, we include ``GeolocalRetrieval'', which was previously also proposed on the Oxford RobotCar dataset~\cite{9449965}.
While regular image retrieval is trained to be globally discriminative, this method learns descriptors that are only discriminative for nearby satellite patches within a $50m$ radius,
and thus assumes a localization prior during both training and testing, similar to our task.
To increase its localization accuracy, we feed it a larger idealized dataset of satellite patches at more densely sampled locations ($200+$ patches in a $50m$ radius) including even patches centered on the actual test locations.
Therefore GeolocalRetrieval could obtain zero meter error if it correctly retrieves the exact satellite patch at each test image location.

%We keep its idealized
%database with satellite patches centered around both training and \textit{test} ground locations.
%The reported metric error is the distance between the query location and the center of the top-1 retrieved satellite patch.
%This setting favors GeolocalRetrieval since it can retrieve the perfect satellite patch for each test image, which would incur zero error.

Table~\ref{table:Oxford_test} shows the localization error among all included methods.
% The ``center-only'' demonstrates that, without densifying the reference satellite patches, the localization error is large.
As expected, with the idealized satellite patches, GeolocalRetrieval delivers lower error than ``center-only''.
However, metric localization methods (CVR and ours) show a clear advantage over GeolocalRetrieval.
This highlights the benefit of conducting metric localization over simply densifying the dataset for retrieval.
Moreover, using our model for metric localization reduces the mean error by 23\% and the median error by 28\% compared to using CVR.
Qualitatively, we again observe the benefit of expressing multi-modal distribution over the CVR's regression
even without the use of panoramic ground images, see Figure~\ref{fig:qualitative_results} example 4.
% In Table~\ref{table:Oxford_probability} we also observe a similar trend as on the VIGOR dataset regarding the probability evaluation.
% Our predicted probability at the ground truth pixel is higher on all test traversals than that under CVR with its estimated error distribution.
The probability evaluation also
aligns with the findings on VIGOR.
Our probability at the ground truth pixel are consistently higher than that under CVR with its estimated error distribution over three test traversals.
Averaged over three test traversals, the mean/median probability at the ground truth pixel for CVR are $1.67\times10^{-4}$/$1.89\times10^{-4}$, and for ours are $1.54\times10^{-3}$/$1.38\times10^{-3}$. 

We also test classification of the orientation 
on this non-panoramic dataset. 
Instead of shifting the ground image, we now rotate the satellite patch 16 times with $22.5^{\circ}$,
starting at $0^{\circ}$ where north points in the vertical up direction.
%Note that there will not be a perfectly aligned satellite patch in the set. % [JK] this now follows from the sentence above
The orientation of a ground image is inferred by selecting the peak probability as we did for VIGOR.
On three test traversals, $72.3\%$, $70.7\%$, and $70.6\%$ of the test samples are predicted with the correct orientation out of the 16 possible directions. More details on orientation classification and the localization results with unknown orientation can be found in the Supplementary Material.

To summarize, also for test images at new days our method shows all-round  superiority, similar to generalization within the same and across areas.

% \begin{table}[ht]
% \centering
% \begin{tabular}{ |p{4cm}|p{1cm}|p{1.1cm}|}
% \hline
% Error (meters) & Mean & Median \\
% \hline
% T.1 Oracle retrieval & 12.06 & 12.66  \\
% T.1 GeolocalRetrieval$^\star$ \cite{9449965} & 5.44 & 4.35  \\
% T.1 CVR \cite{zhu2021vigor}& 1.88 & 1.47  \\
% % T.1 Ours-binary & 1.67 & 1.26  \\
% T.1 Ours & \textbf{1.42} & \textbf{1.10}  \\
% \hline
% T.2 Oracle retrieval & 12.12 & 12.66  \\
% T.2 GeolocalRetrieval$^\star$ & 6.97 & 5.31  \\
% T.2 CVR & 2.64 & 1.99  \\
% % T.2 Ours-binary & 2.30 & 1.49  \\
% T.2 Ours & \textbf{1.95} & \textbf{1.33}  \\
% \hline
% T.3 Oracle retrieval & 12.10 & 12.63  \\
% T.3 GeolocalRetrieval$^\star$  & 5.62 & 4.21  \\
% T.3 CVR & 2.35 & 1.71  \\
% % T.3 Ours-binary & 2.13 & 1.41  \\
% T.3 Ours & \textbf{1.94} & \textbf{1.29}  \\
% \hline
% Mean Oracle retrieval & 12.09 & 12.65  \\
% Mean GeolocalRetrieval$^\star$ & 6.01 & 4.62  \\
% Mean CVR & 2.29 & 1.72  \\
% % Mean Ours-binary & 2.03 & 1.39  \\
% Mean Ours & \textbf{1.77} &  \textbf{1.24} \\
% \hline
% \end{tabular}
% \caption{Evaluation on Oxford RobotCar Test Sets (best results in bold). T.N stands for the Nth testing traversal, and the mean error over 3 traversals is in the bottom row. The method with $^\star$ indicates that the training and test ground images are the same but satellite patches are different.}
% \label{table:Oxford_test}
% \end{table}

\begin{table}[t]
\centering
\caption{Localization error on Oxford RobotCar. Shown are the average $\pm$ standard deviation of `mean' and `median' errors over 3 test traversals.
Best results in bold.
$\star$: uses the same  training and test ground images, but more overlapping satellite patches to obtain finer localization through image retrieval only}
\label{table:Oxford_test}
\begin{tabular}{|p{3.4cm}|p{2cm}|p{2cm}|}
\hline
\textit{Error (meters)} & Mean & Median \\
\hline
Center-only & $12.09 \pm 0.02$ & $12.65 \pm 0.01$ \\
GeolocalRetrieval$^\star$ \cite{9449965}& $6.01 \pm 0.68$ & $4.62 \pm 0.49$  \\
CVR \cite{zhu2021vigor} & $2.29 \pm 0.31$& $1.72 \pm 0.21$  \\
Ours & $\textbf{1.77} \pm0.25$ &  $\textbf{1.24}\pm0.10$ \\
\hline
\end{tabular}
\end{table}

\section{Conclusion}
\label{sec:conclusion}
In this work, we focused on visual cross-view metric localization on a known satellite image, a relatively unexplored task.
%We addressed the limitations of the state-of-the-art vision baseline that was originally proposed for joint global geolocalization through image retrieval.
%, by combining the insights from metric learning and dense probabilistic prediction.
In contrast to the state-of-the-art regression-based baseline, our method provides a dense multi-modal spatial distribution.
We studied the architectural design differences, and showed generalization to new measurements in the same area, across areas, and generalizing across time on two state-of-the-art datasets.
Our method surpasses the state-of-the-art by 51\%, 37\%, and 28\% respectively in the median localization error.
In a few cases the multi-modal output yields higher distance errors, e.g.~when an incorrect mode is deemed more probable.
Still, our probabilities can be used to filter such large errors and have less risk of excluding the true location.
%In the above circumstances, we not only demonstrate a clear advantage of metric localization over image retrieval in the context of outdoor vehicle localization, but also
We show that our method is robust against small orientation noise, and is capable to roughly classify the orientation from its prediction confidence.
Future work will address temporal filtering and fine-grained orientation estimation.

\subsubsection*{Acknowledgements.}
% https://www.nwo.nl/financiering/hoe-werkt-dat/acknowledgement
% Placeholder acknowledgements
This work is part of the research programme Efficient Deep Learning (EDL) with project number P16-25, which is (partly) financed by the Dutch Research Council (NWO).

%%%%%%%%% REFERENCES
\clearpage
\bibliographystyle{splncs04}
\bibliography{egbib}

%%%%%%%%% APPENDIX
\clearpage
\appendix
\section*{Appendix}
The content in this appendix is related to the main paper as follows:
\begin{enumerate}[label=\Alph*]
    \item More details on our used datasets in \textbf{Section 4.1 Datasets}.
    \item Supplemental details on \textbf{Section 4.2 Evaluation metrics} about our post-processing on baseline CVR for generating probability estimation.
    \item Additional results for \textbf{Section 4.4 Generalization in the same area / across areas}.
    \item Additional results for \textbf{Section 4.5 Generalization across time}.
    \item A detailed network diagram to supplement \textbf{Section 3.2 Proposed method}.
\end{enumerate}

\section{Supplemental details on datasets}
The original \textbf{VIGOR dataset} \cite{zhu2021vigor} does not provide a validation set for either the ``same-area'' nor the ``cross-area'' split.
Therefore, we randomly split 20\% of data from the training set as our validation set, used to determine the stopping epoch during training. The test set is kept as is and is used to report the evaluation results.
Thereby, in the ``same-area'' split, there are 42087, 10522, and 52605 ground images in training, validation, and test set respectively. All three sets share the 90618 satellite images.
In the ``cross-area'' setting, there are 41216, 10304, and 53694 ground images in training, validation, and test set respectively, and there are 44055 satellite images for training and validation in the first two cities, and 46563 satellite images for across-city testing.

\begin{figure}[h]
    \centering
    \includegraphics[width=0.5\textwidth]{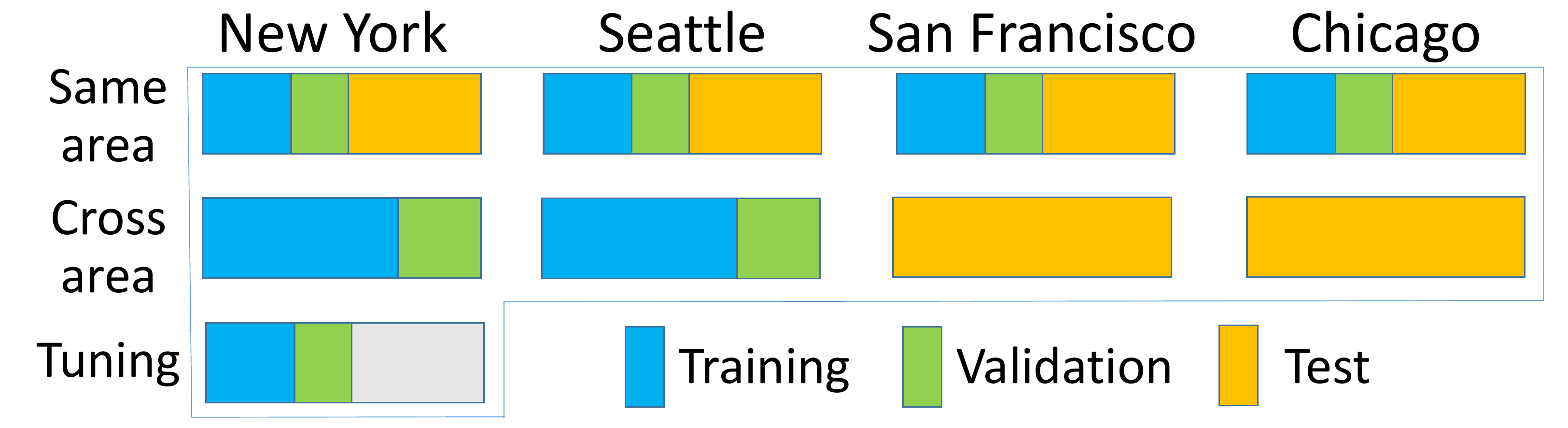}
    \caption{Overview of our data split on VIGOR dataset
    }
    \label{fig:data_split}
\end{figure}

Note that we do not use our validation set to find one set of hyper-parameters for both the ``same-area'' and ``cross-area'' experiments because the ``same-area'' validation set partially overlaps with the ``cross-area'' test set and vice versa.
Hence, we make use of the ``same-area'' training image from New York to create a separate ``tuning'' set (11108 training and 2777 validation) to search for one set of hyper-parameters for all experiments and speed up the hyper-parameter search, see Figure~\ref{fig:data_split}. We will release our data splits.

The original \textbf{Oxford RobotCar dataset} \cite{OxfordRobotCar1,OxfordRobotCar2} does not contain satellite images.
We make use of the satellite patches provided by \cite{Geolocal_feature,9449965}.
We intend to conduct data augmentation such that, given a ground-level image, we can crop any satellite patch from a continuous satellite image that contains the ground image.
However, the satellite patches from \cite{Geolocal_feature,9449965} do not provide enough coverage of the target area.

\begin{figure}[t]
    \centering
    \includegraphics[width=0.6\textwidth]{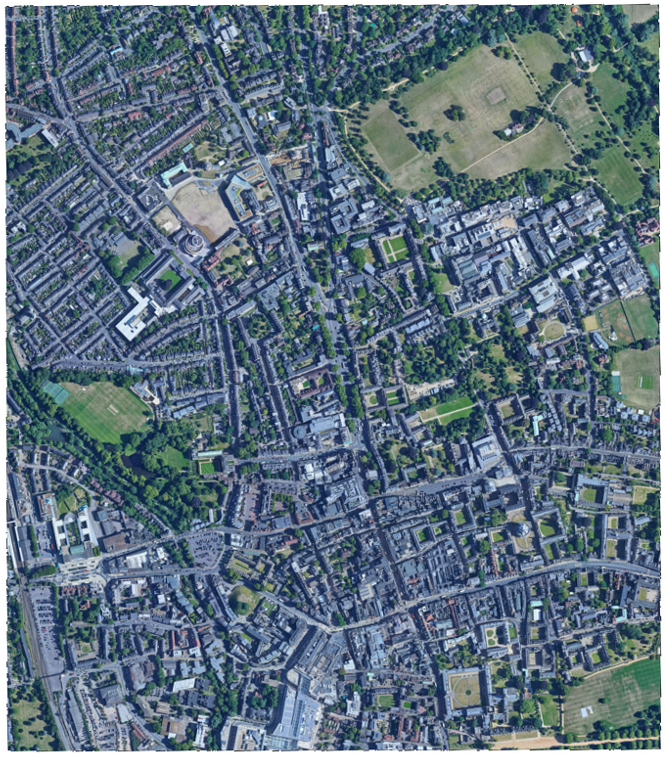}
    \caption{Overview of our stitched continuous satellite image
    }
    \label{fig:Oxford_satellite_image}
\end{figure}

To enable the intended data augmentation, we collect additional satellite patches from Google Maps Static API\footnote{\url{https://developers.google.com/maps/documentation/maps-static/overview}},
and stitch all satellite patches to create a continuous satellite map that covers the target area ($\sim1.5km \times 1.8km$), see Figure~\ref{fig:Oxford_satellite_image}. 
The geographical coordinates of each pixel in our continuous satellite image are known.
Therefore, given the geographical location of a ground image, we can find its image pixel coordinates on the satellite image.
We \textit{will release} the raw data\footnote{We release the images in accordance with the ``fair use'' policy in Google Maps/Google Earth Terms of Service, version: Jan 12, 2022 (\url{https://www.google.com/help/terms_maps/}). The released data is for the reproducibility of scientific research and cannot be used for commercial purposes. Google remains the copyright of images.} we collected and the code for stitching satellite patches.

As mentioned in \textbf{main paper Section 4.1}, we align the rotation between the ground image and satellite patch.
Given a ground image and the heading of the camera we first crop a large satellite patch from the continuous map at the targeted cropping location and then rotate the satellite image such that the up direction in the cropped large patch corresponds to the viewing direction in the ground image.
Then we crop the satellite patch with the required resolution from the rotated large patch.
Note that, when testing classification of the orientation, \textbf{main paper Section 4.5 second to the last paragraph}, the satellite patches are rotated with a fixed set of angles. 
Thus the orientation-aligned patch will not appear in the set.
\section{Details on post-processing CVR localization}
\label{sec:post-processing_CVR}
As we illustrated in \textbf{main paper Section 4.2}, to acquire the probability estimation from the baseline CVR method, we assume the regressed location from CVR is the mean of an isotropic Gaussian distribution and we estimate the standard deviation of this Gaussian distribution on the validation set.

Specifically, we record the error distribution on the validation set, and calculate the standard deviation $SD$ using $SD = \sqrt{\frac{\sum^n{e^2}}{n}}$, in which $e$ is the distance error of each ground image in the validation set and n is the number of ground images. 
On the VIGOR dataset, the estimated standard deviation is 12.36m on the same-area split and is 11.64m on the cross-area split.
On the Oxford RobotCar dataset, the value is 3.36m.

\section{Additional results, generalization in the same area / across areas}
\label{sec:C}
Here, we provide more results for our experiments in \textbf{main paper Section 4.4}.

\subsection{Extra model variations}
We studied two extra variations of our proposed method.\\
Variation 1 has a multi-layer perceptron on top of our proposed model.
It uses the output heat map to regress the relative 2D location offset between the ground camera's location and the center of the satellite patch.
Variation 2 drops the whole decoder and selects the location of the highest similarity score out of the $8\times8$ matching score map. 
The ground descriptor is then concatenated with the satellite descriptor at the selected location.
The concatenated descriptor is used by a multi-layer perceptron to regress the relative 2D location offset between the ground camera's location and the center of the small satellite patch corresponding to the selected location. 
Note that, both variations only regress a single location without dense uncertainty estimates.
Similar to the post-processing on CVR's output, see Section~\ref{sec:post-processing_CVR}, we assumed the regressed location is the mean of an isotropic Gaussian distribution and estimate the standard deviation on the validation set to acquire the probabilistic output.
\begin{table}[t]
\centering
\caption{Localization error on VIGOR satellite patches. Best in bold.}
\label{table:model_variations_error}
\begin{tabular}{|p{1.8cm}|p{1.1cm}|p{1.1cm}|p{1.1cm}|p{1.1cm}||p{1.1cm}|p{1.1cm}|p{1.1cm}|p{1.1cm}|} 
\hline
& \multicolumn{4}{|c||}{\textit{Same-area}} & \multicolumn{4}{|c|}{\textit{Cross-area}} \\
\cline{2-9}
& \multicolumn{2}{|c|}{Positives} & \multicolumn{2}{|c||}{Pos.+semi-pos.} &  \multicolumn{2}{|c|}{Positives} & \multicolumn{2}{|c|}{Pos.+semi-pos.}\\
\hline
\textit{Error (m)} & Mean & Median & Mean & Median & Mean & Median & Mean & Median\\
\hline
Center-only & 14.15 & 14.82 & 27.78 & 28.85 & 14.07 & 14.07 & 27.80 & 28.89 \\
\hline
CVR \cite{zhu2021vigor} & 10.55 & 9.31 & 16.64 & 13.82 & 11.26 & 10.02 & 18.66 & 16.73\\
\hline
Ours & 9.86 & \textbf{4.58} & 13.45 & \textbf{5.39} & 13.06 & \textbf{6.31} & 17.13 & 7.78\\
\hline
Variation 1 & \textbf{8.01} & 6.29 & \textbf{12.26} & 9.20 & \textbf{9.62} & 7.57 & \textbf{14.25} & 11.42 \\
\hline
Variation 2 & 10.69 & 5.88 & 13.40 & 6.07 & 12.49 & 6.92 & 15.29 & \textbf{7.30} \\
\hline
\end{tabular}
\end{table}

\begin{table} 
\centering
\caption{Probability at the ground truth location on VIGOR positive satellite patches. Best in bold.}
\label{table:model_variations_probability}
\begin{tabular}{ |p{2cm}|p{2cm}|p{2cm}|p{2cm}|p{2cm}|} 
\hline
\textit{Prob.~at GT,}
& \multicolumn{2}{|c|}{Same-area} & \multicolumn{2}{|c|}{Cross-area}\\
\cline{2-5} % unlike \hline, this skips the first cell
\textit{Positives} & Mean & Median & Mean & Median\\
\hline
Uniform & $3.81\times10^{-6}$ & $3.81\times10^{-6}$ & $3.81\times10^{-6}$ & $3.81\times10^{-6}$\\
\hline
CVR \cite{zhu2021vigor} & $1.55\times10^{-5}$ & $1.70\times10^{-5}$ & $1.57\times10^{-5}$ & $1.72\times10^{-5}$\\
\hline
% Ours-binary & 3.50e-5 & 3.14e-5 & 3.18e-5 & 2.85e-5\\
% \hline
Ours & $\mathbf{2.93\times10^{-4}}$ & $\mathbf{1.17\times10^{-4}}$ & $\mathbf{1.54\times10^{-4}}$ & $\mathbf{7.06\times10^{-5}}$ \\
\hline
Variation 1 & $1.15\times10^{-5}$ & $8.40\times10^{-6}$ & $1.13\times10^{-5}$ & $9.36\times10^{-6}$ \\
\hline
Variation 2 & $7.25\times10^{-6}$ & $6.89\times10^{-6}$ & $7.09\times10^{-6}$ & $6.86\times10^{-6}$ \\
\hline
\end{tabular}
\end{table}

In Table~\ref{table:model_variations_error} and \ref{table:model_variations_probability}, we provide the comparison of all models on the VIGOR test splits.
Variation 1 is better than CVR~\cite{zhu2021vigor} all-round.
This benefit might come from the larger model size and the dense feature before the final regression head.
However, the regression-based variation 1 still has a significantly higher median error, which might be due to the regression head picking the midpoint between visually similar locations.
Regarding variation 2, on the same-area split, it performs worse than our model.
This shows the advantage of having a decoder to process the matching score and satellite feature.
However, on the cross-area split, our model performs slightly worse than variation 2.
Possibly, given the more uncertain matching score map and features in unseen areas, the decoder does not know how to reduce this uncertainty effectively.

Importantly, variation 1 and 2 are not optimized for higher probability at the ground truth location.
Thus, both variations have over one magnitude lower probability at ground truth than our model.
We highlight again that probability estimation is crucial in both sensor fusion and temporal filtering.
Last but not least, both variations miss the dense output to capture a multi-modal distribution.
Thus, we conclude that our proposed model and problem formulation is better.

\subsection{Probability evaluation}
\begin{figure}[ht]
    \centering
    \includegraphics[width=0.7\textwidth]{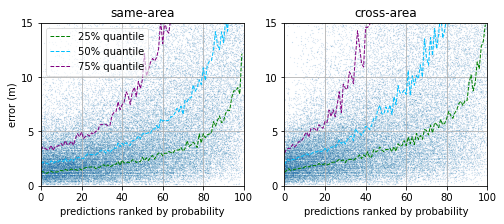}
    \caption{Ranking the predictions using their probabilities on VIGOR “positives”.}
    \label{fig:prop_error_nonaccumulative}
\end{figure}

Supplementary to the cumulative statistics shown in \textbf{main paper Figure 5}, we show the noncumulative statistics of the predictions ranked by their probabilities in Figure~\ref{fig:prop_error_nonaccumulative}.

\subsection{Orientation}
\label{sec:orientation_and_metric_error_VIGOR}
Additional to solely evaluating either localization or orientation, we also tested jointly evaluating both on the positive satellite patches.
As in \textbf{main paper Section 4.4. Orientation}, the ground panorama is rotated by multiples of $22.5^{\circ}$ up to $360^{\circ}$. We pair up each of the rotated panoramas with the satellite patch. Hence one mini-batch contains the same satellite image with 16 rotated ground panoramas.
We collect all 16 activation maps and pass them into a single softmax operation.
The location and corresponding orientation of the peak probability is then the localization and orientation estimation.
Under this setting, the mean/median metric localization error increases from 9.86/4.58 (main paper Table 2 Same-area Positives columns) to 13.99/8.28 on the same-area test set, and from 13.06/6.31 (main paper Table 2 Cross-area Positives columns) to 18.18/14.61 meters on cross-area test set.
This is a more challenging setting that leads to a noticeable decrease in performance.

\subsection{Localization qualitative results}
\begin{figure}[t]
    \centering
    \hspace*{-0.5em}
    \includegraphics[width=0.17\textwidth]{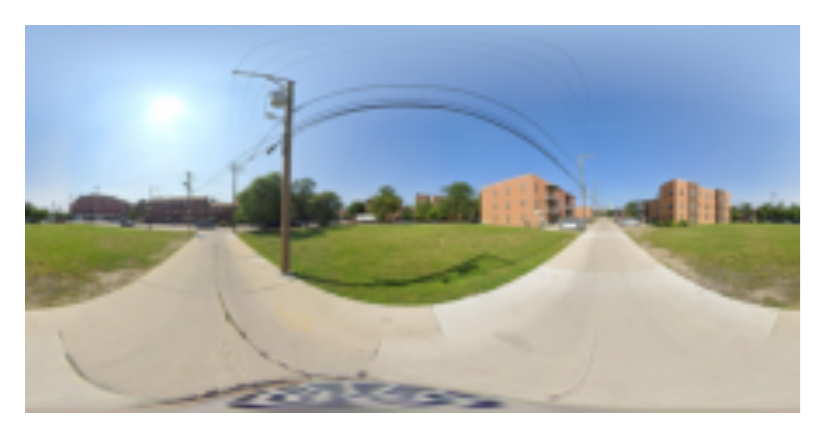}\hspace*{-0.4em}
    \includegraphics[width=0.085\textwidth]{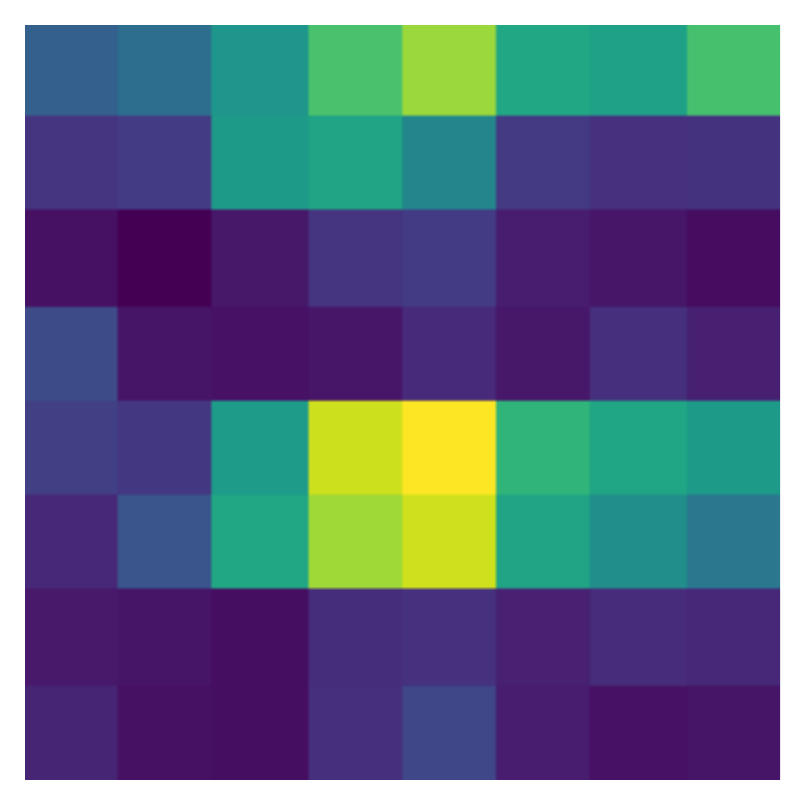}\hspace*{-0.4em}
    \includegraphics[width=0.17\textwidth]{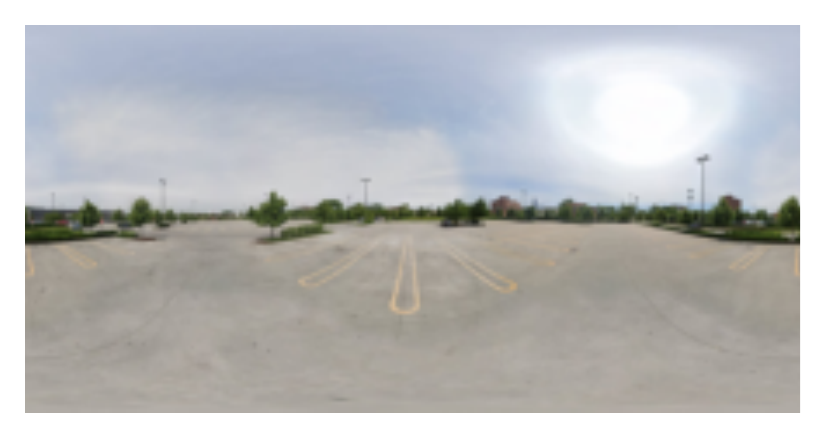}\hspace*{-0.4em}
     \includegraphics[width=0.085\textwidth]{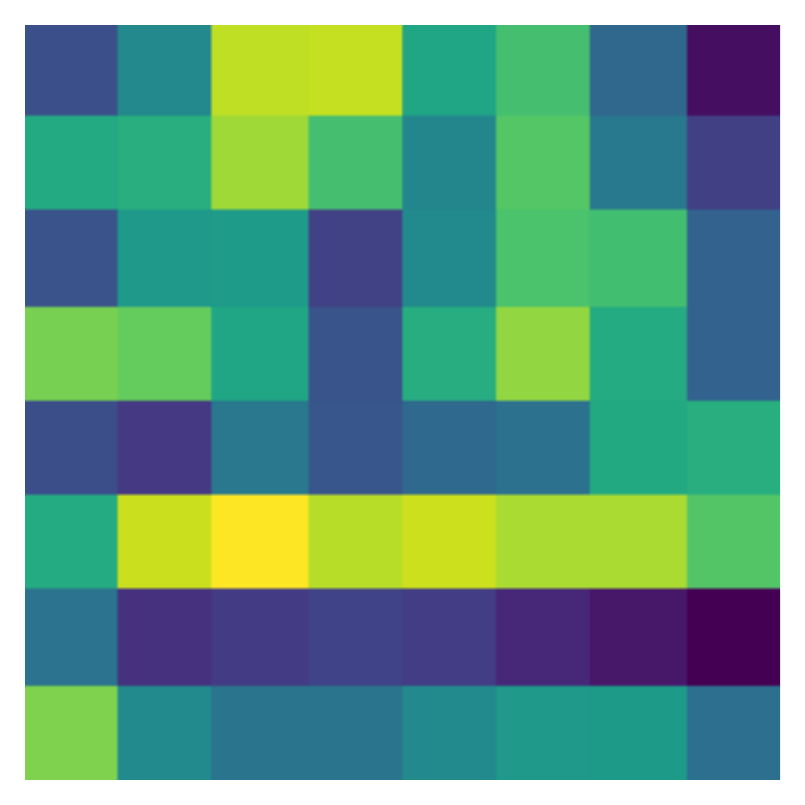}\hspace*{-0.4em}
    \includegraphics[width=0.17\textwidth]{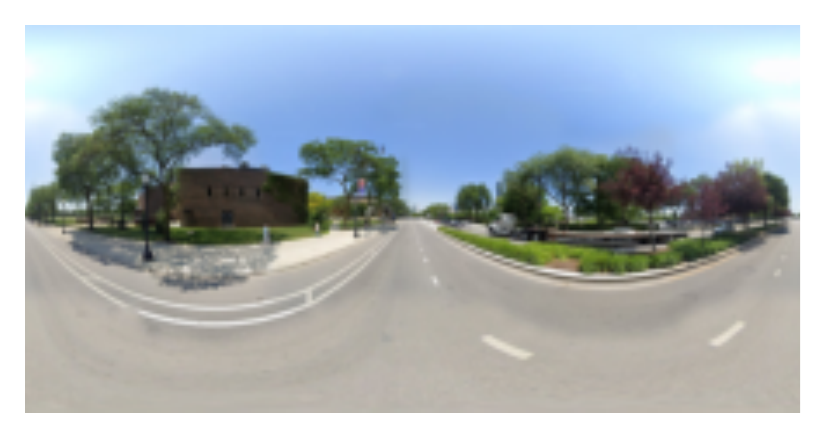}\hspace*{-0.4em}
     \includegraphics[width=0.085\textwidth]{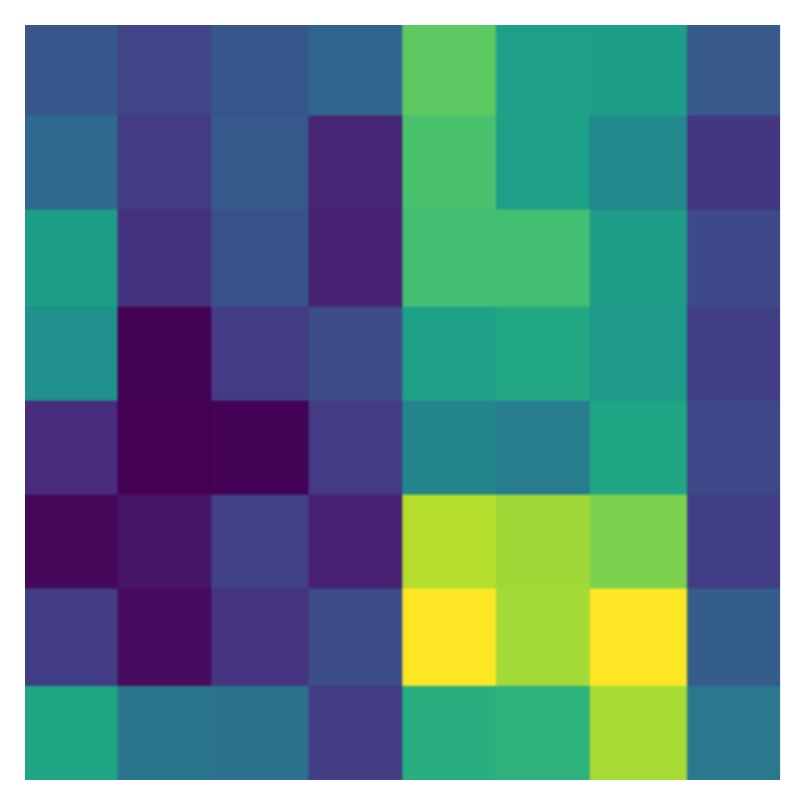}\hspace*{-0.4em}
    \includegraphics[width=0.17\textwidth]{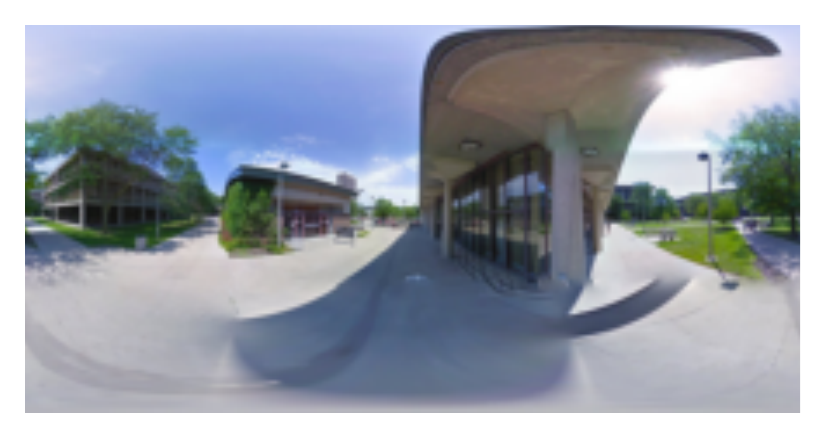}\hspace*{-0.4em}
     \includegraphics[width=0.085\textwidth]{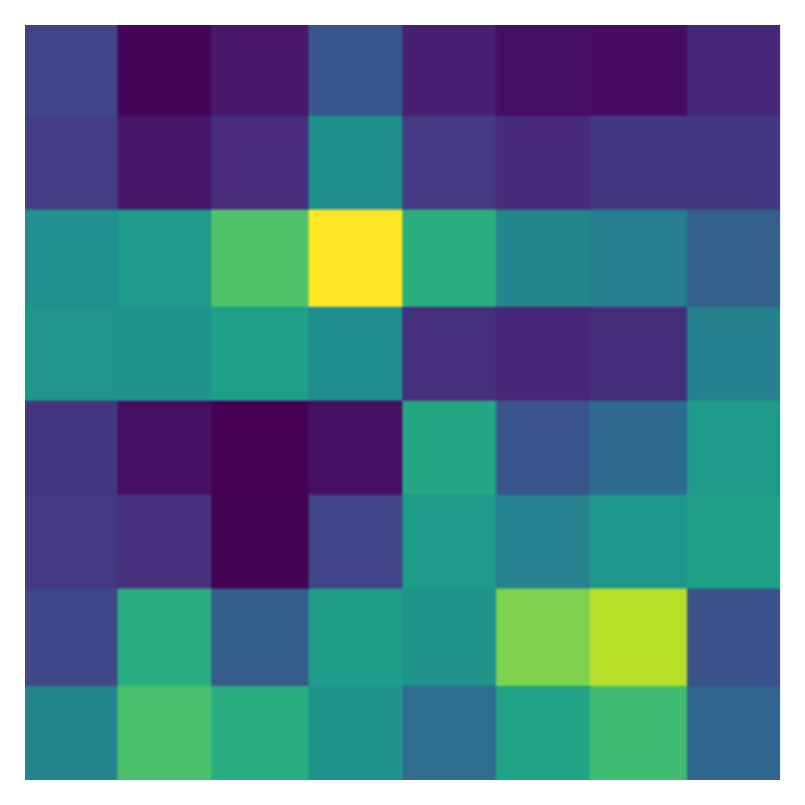}
    \\
    \includegraphics[width=0.25\textwidth]{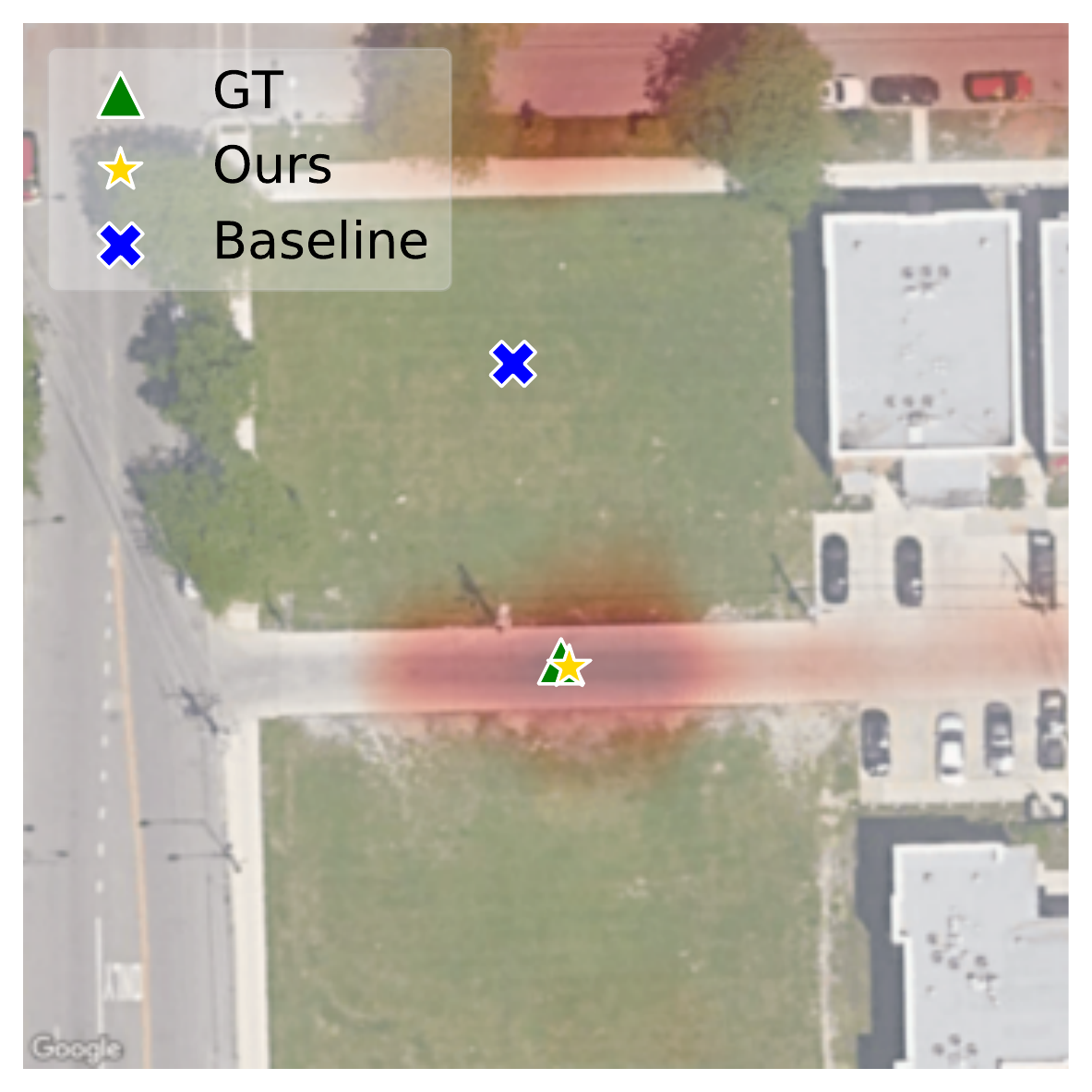}\hspace*{-0.3em}
    \includegraphics[width=0.25\textwidth]{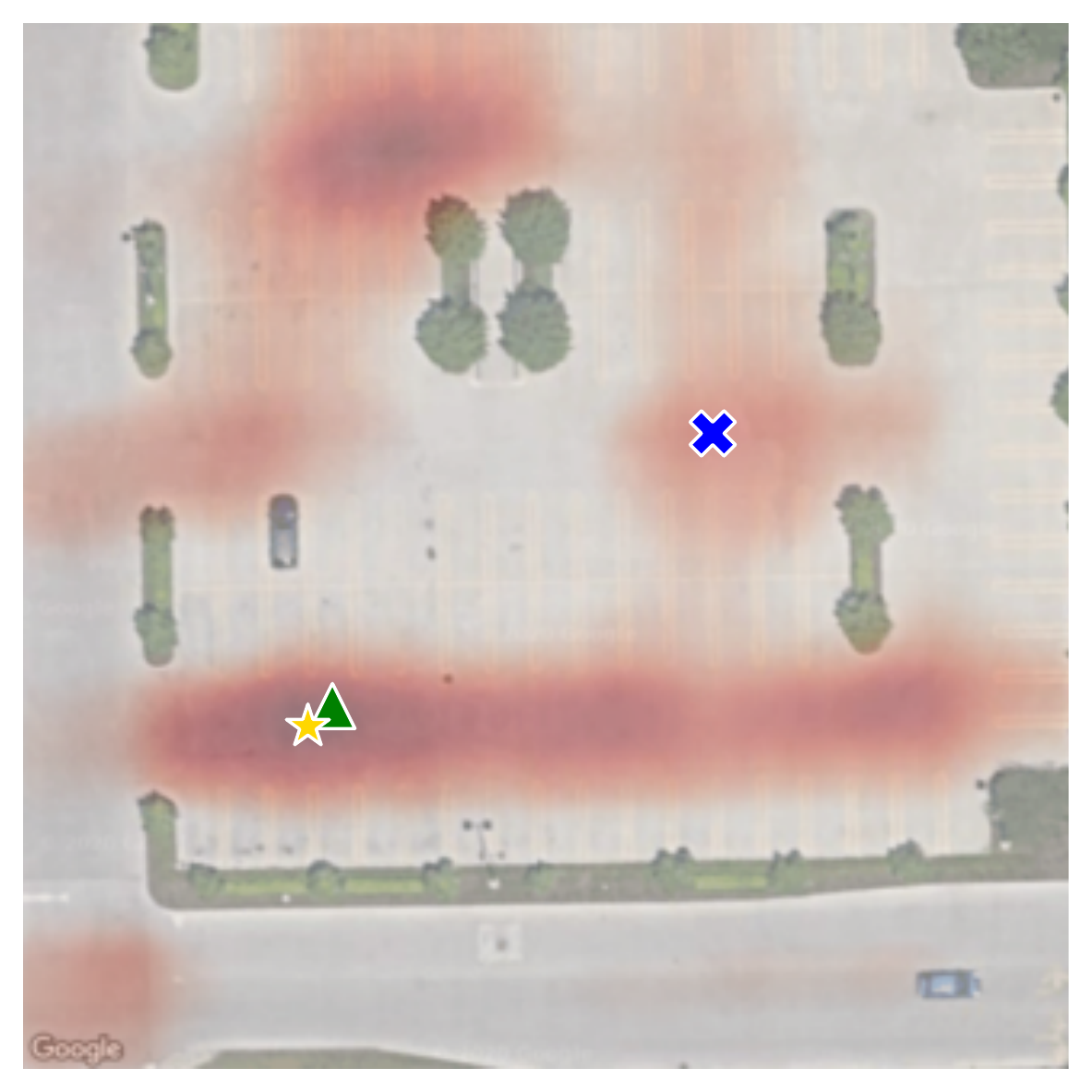}\hspace*{-0.3em}
    \includegraphics[width=0.25\textwidth]{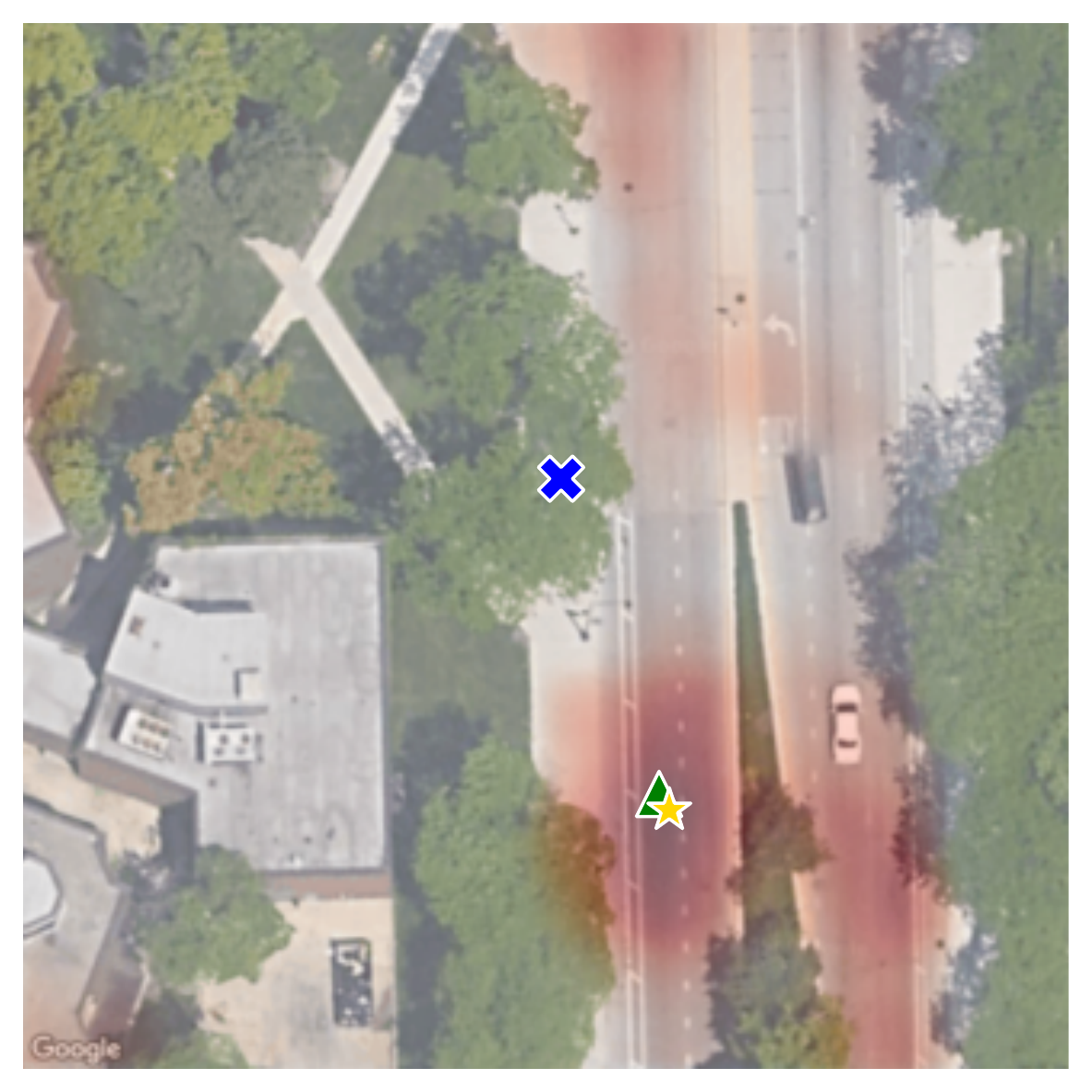}\hspace*{-0.3em}
    \includegraphics[width=0.25\textwidth]{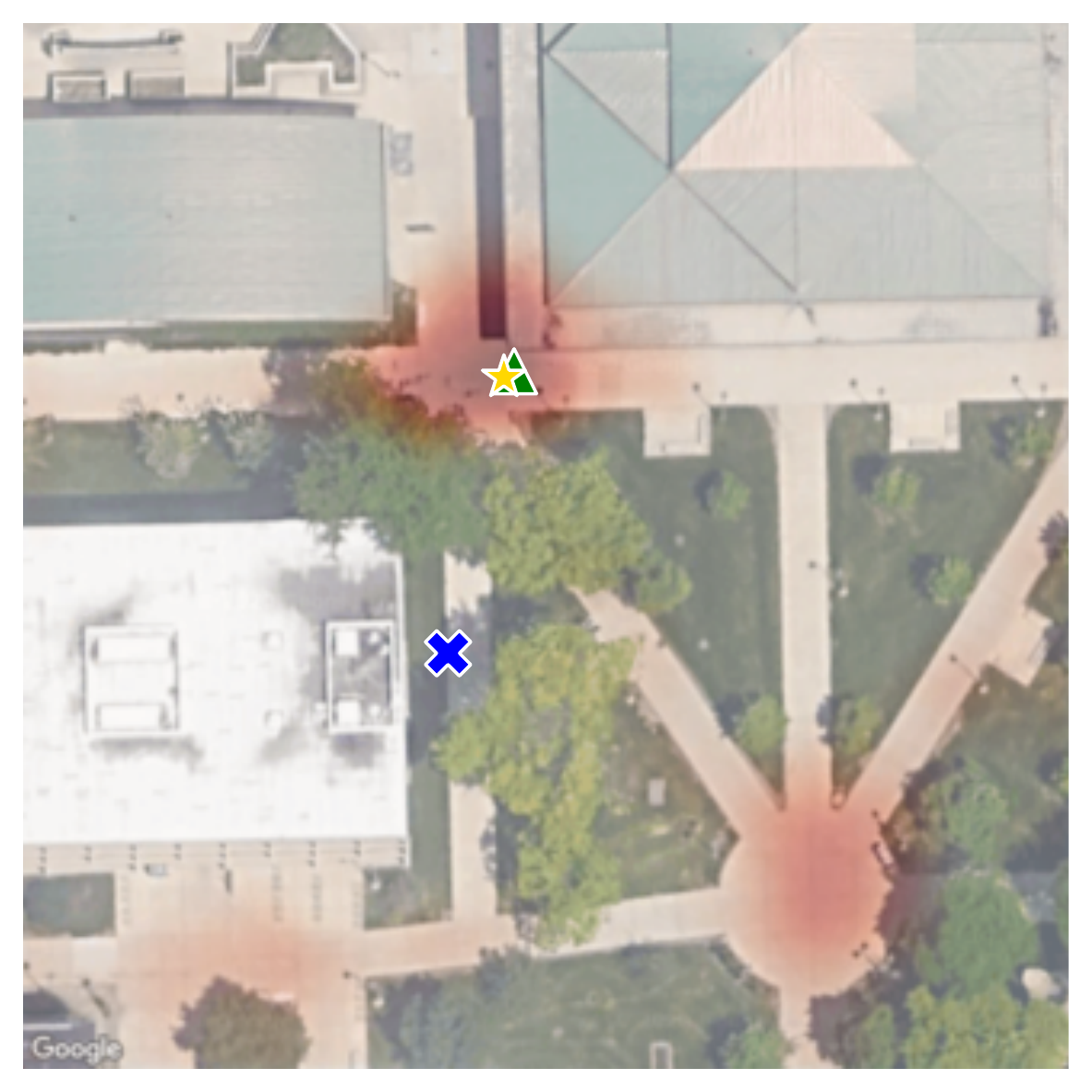}
    % \\
    % \includegraphics[width=0.25\textwidth]{figures_sup/samearea_good/same_CVR_heatmap_47179.pdf}\hspace*{-0.3em}
    % \includegraphics[width=0.25\textwidth]{figures_sup/samearea_good/same_CVR_heatmap_48803.pdf}\hspace*{-0.3em}
    % \includegraphics[width=0.25\textwidth]{figures_sup/samearea_good/same_CVR_heatmap_49240.pdf}\hspace*{-0.3em}
    % \includegraphics[width=0.25\textwidth]{figures_sup/samearea_good/same_CVR_heatmap_49895.pdf}
    \caption{Qualitative results on VIGOR dataset same-area split. 
    Top: input ground images and matching score maps at the model bottleneck, bottom: input satellite image overlayed with outputs from CVR and our method.
    }
    \label{fig:VIGOR_samearea}
\end{figure}

\begin{figure}[t]
    \centering    
    \hspace*{-0.5em}
    \includegraphics[width=0.17\textwidth]{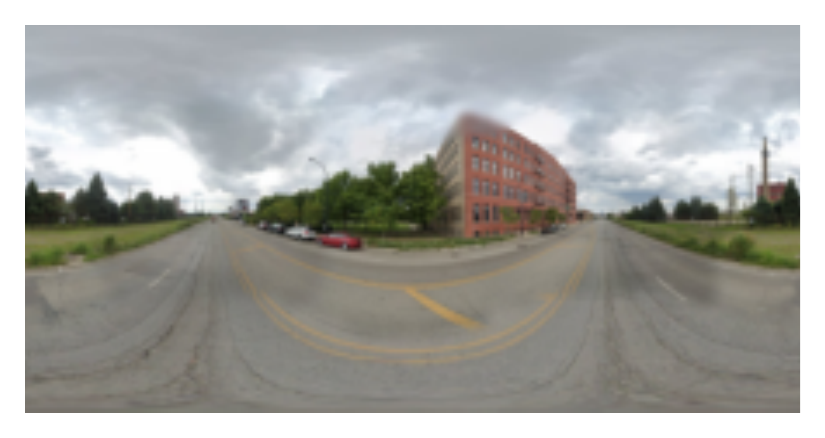}\hspace*{-0.4em}
    \includegraphics[width=0.085\textwidth]{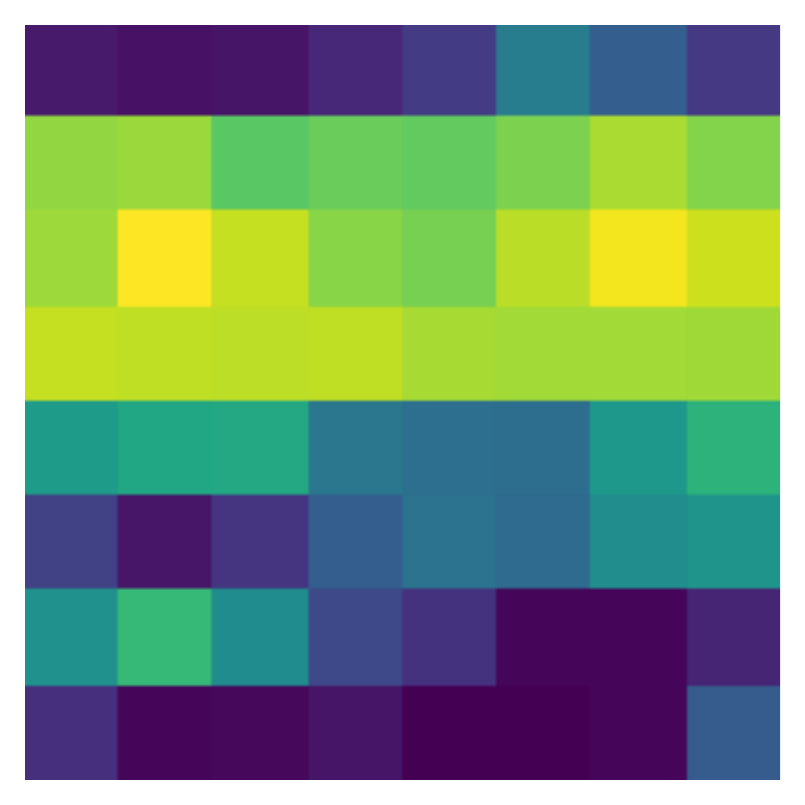}\hspace*{-0.4em}
    \includegraphics[width=0.17\textwidth]{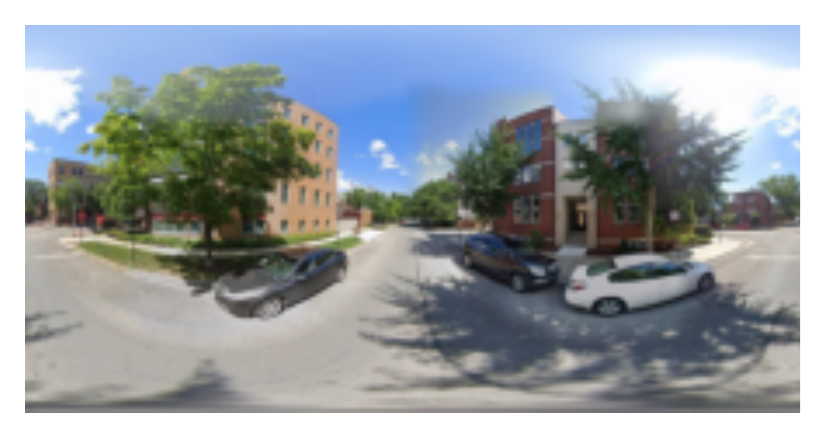}\hspace*{-0.4em}
     \includegraphics[width=0.085\textwidth]{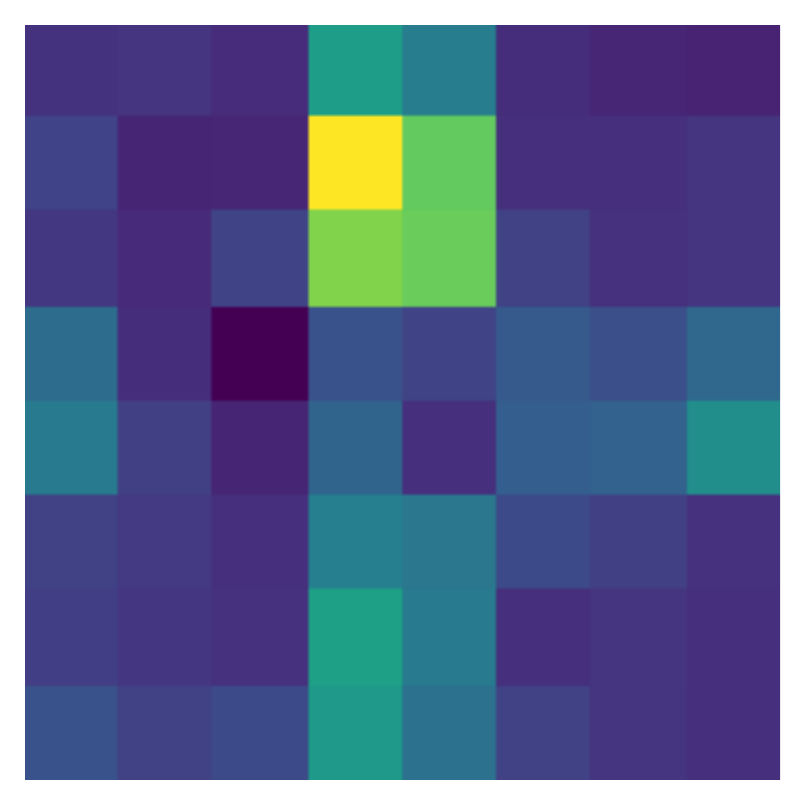}\hspace*{-0.4em}
    \includegraphics[width=0.17\textwidth]{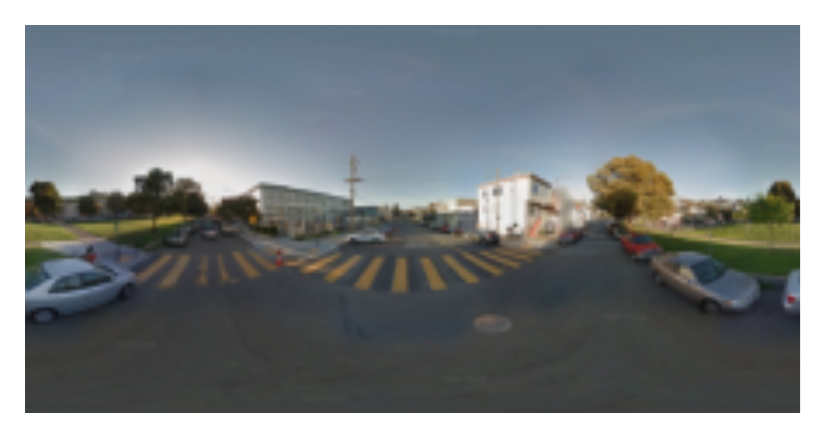}\hspace*{-0.4em}
     \includegraphics[width=0.085\textwidth]{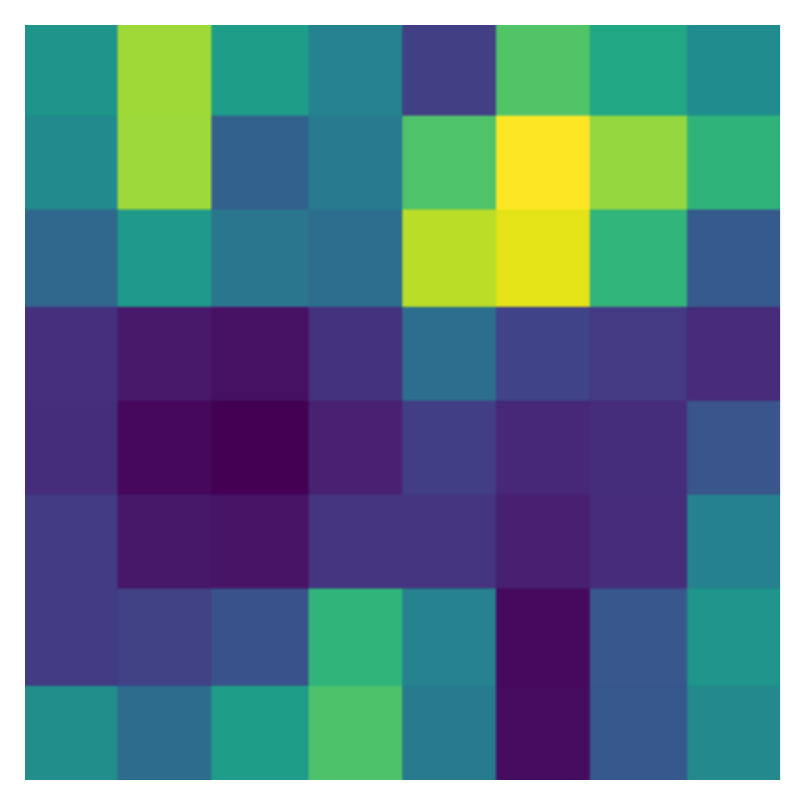}\hspace*{-0.4em}
    \includegraphics[width=0.17\textwidth]{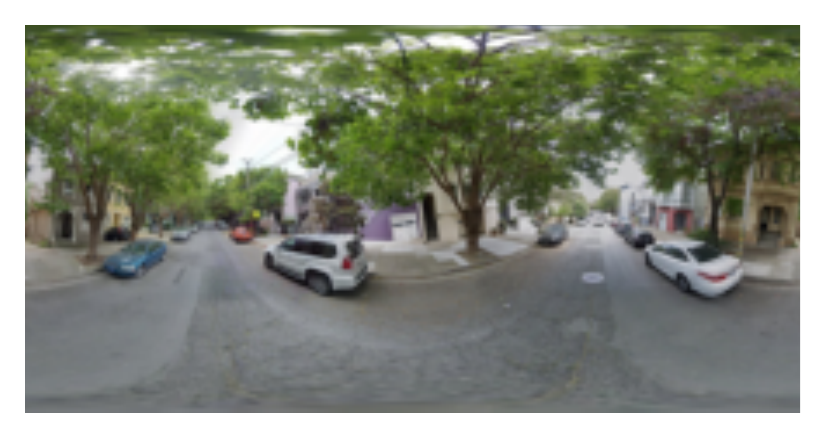}\hspace*{-0.4em}
     \includegraphics[width=0.085\textwidth]{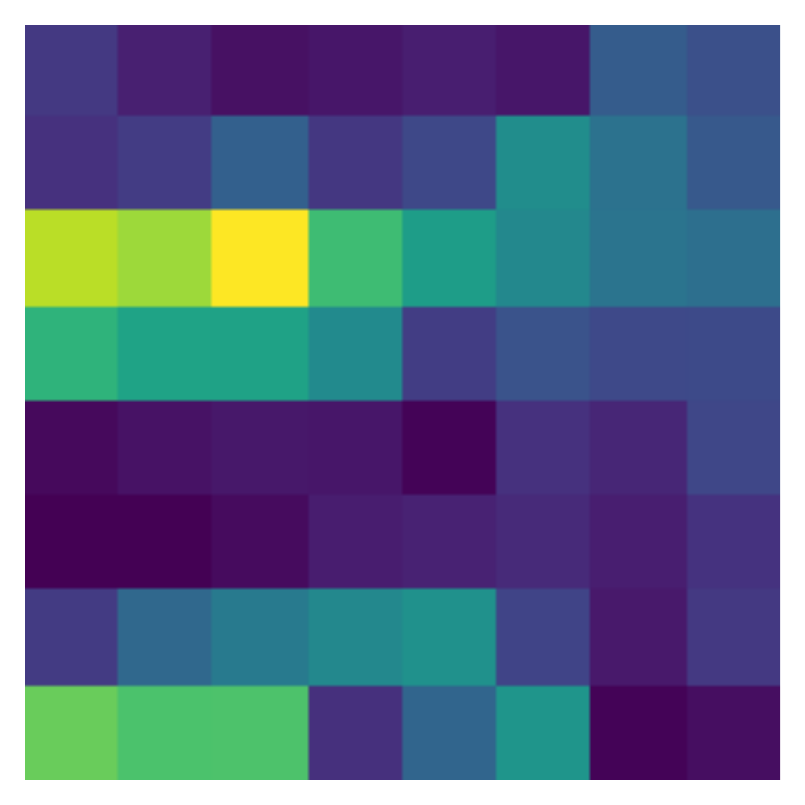}
    \\%\vspace*{-0.3em}
    \includegraphics[width=0.25\textwidth]{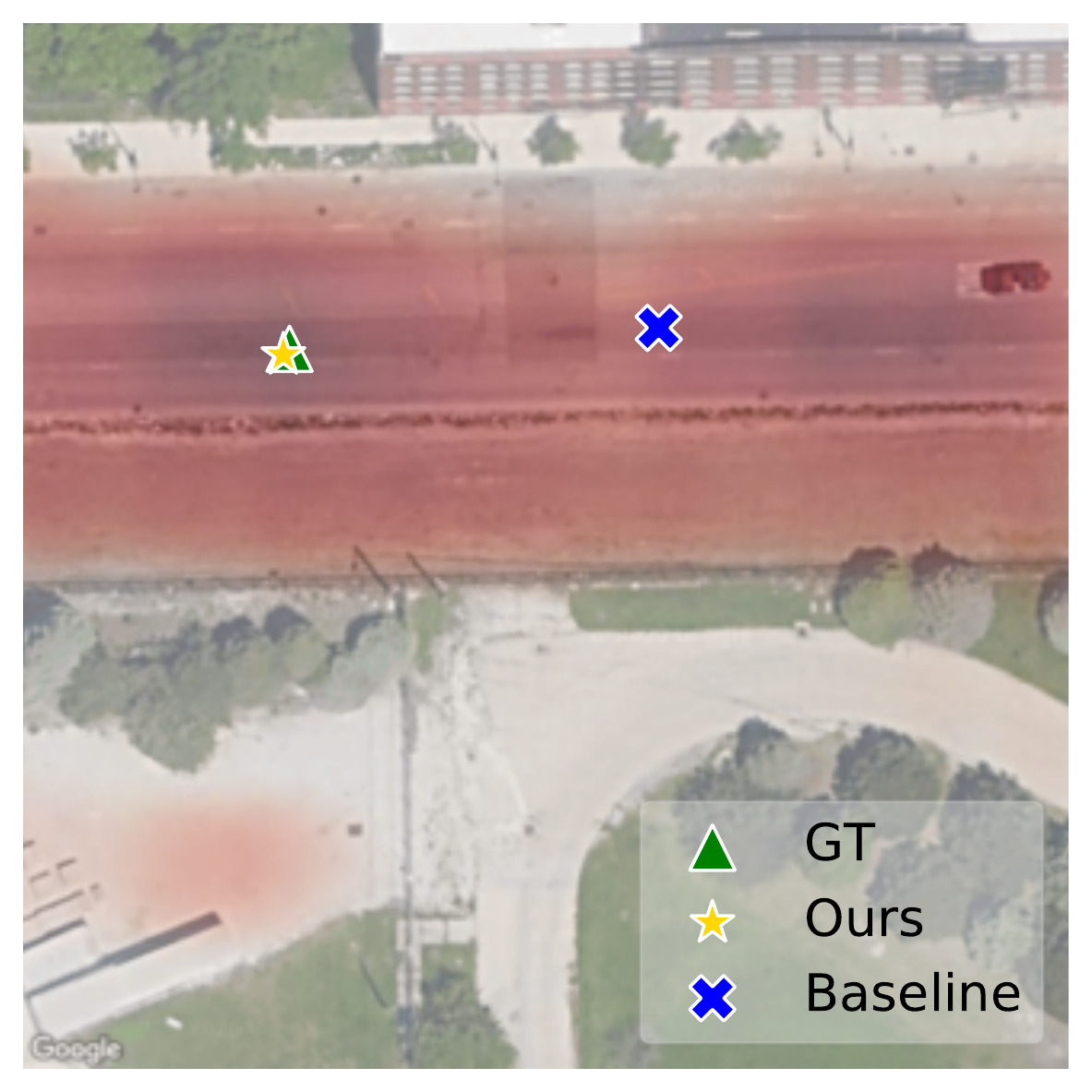}\hspace*{-0.3em}
    \includegraphics[width=0.25\textwidth]{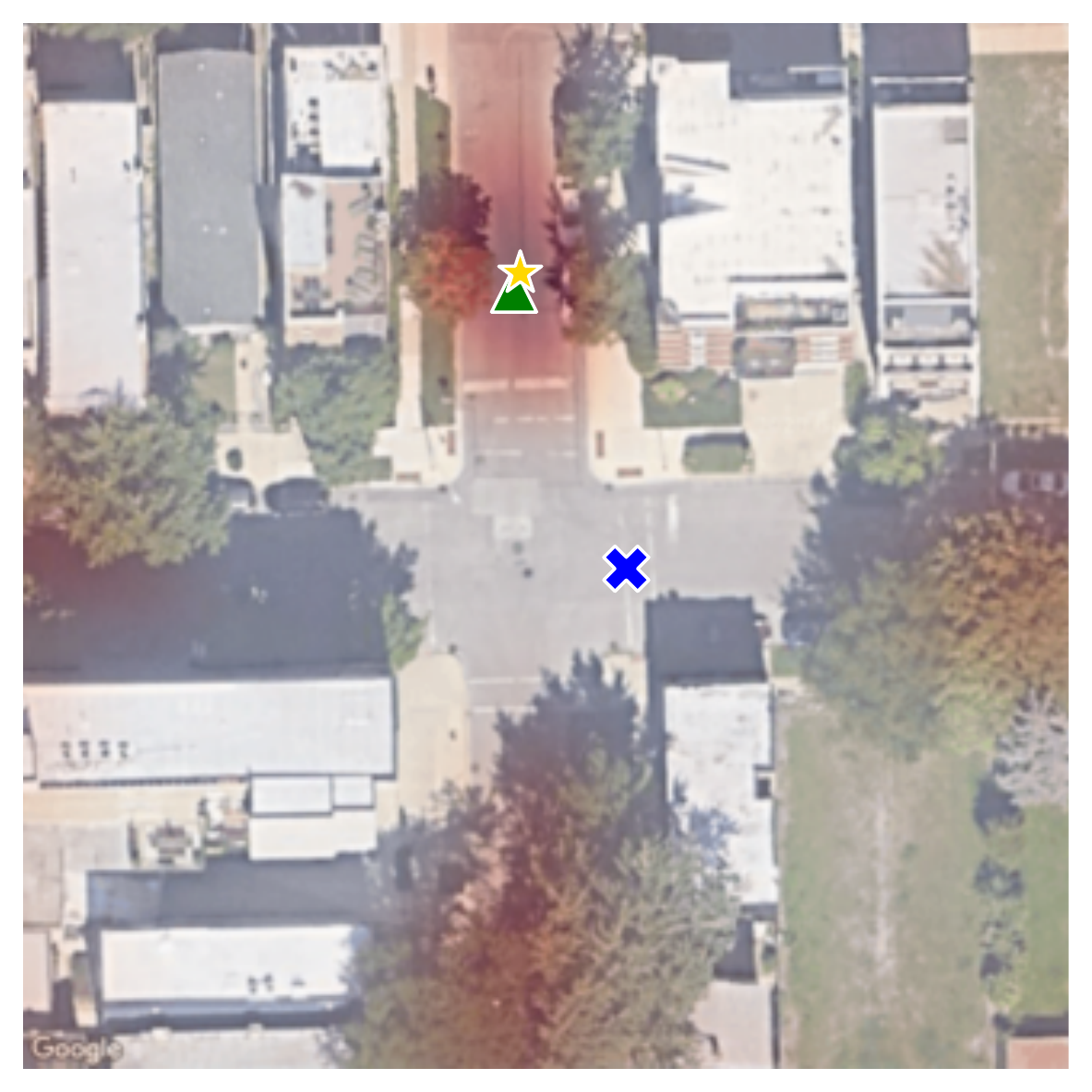}\hspace*{-0.3em}
    \includegraphics[width=0.25\textwidth]{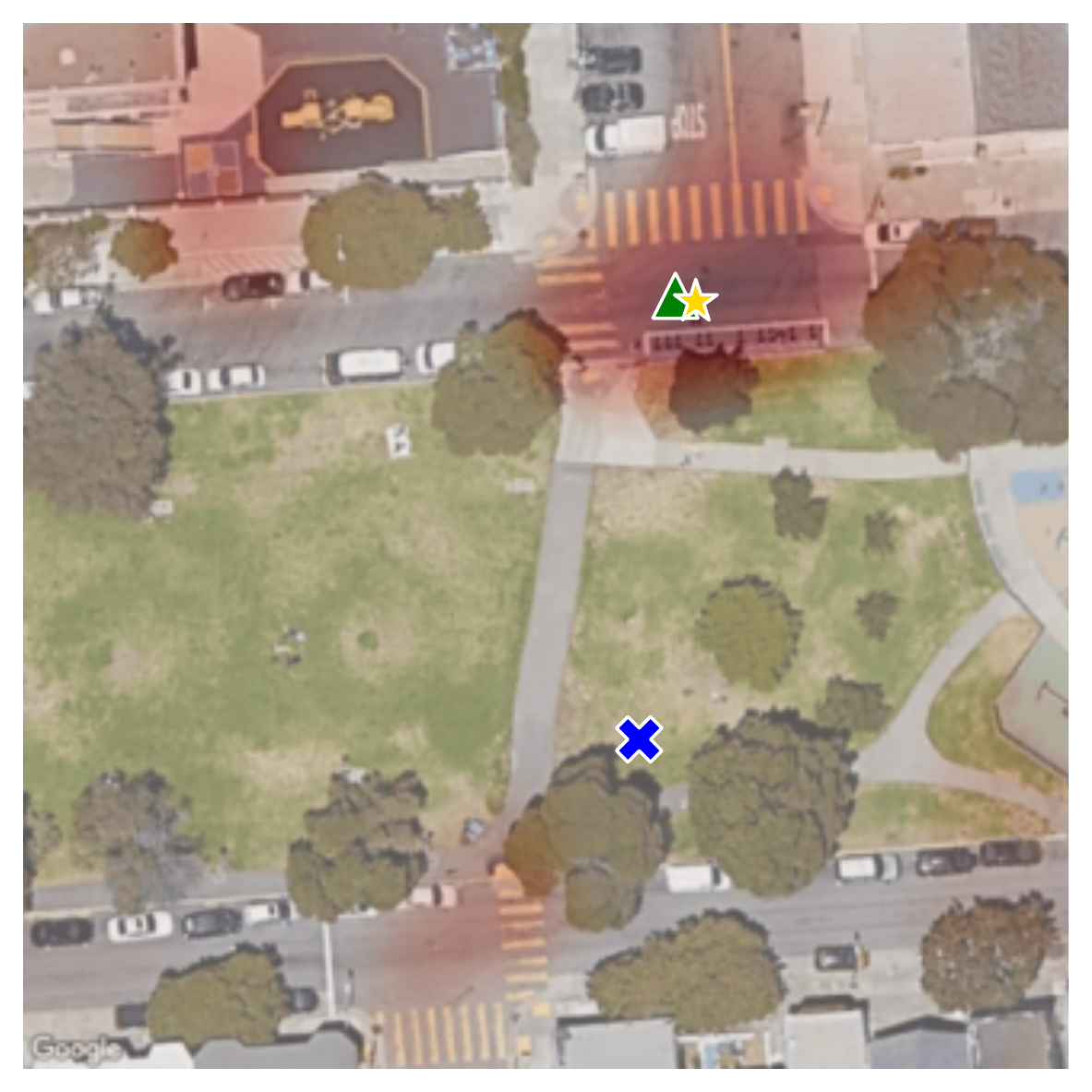}\hspace*{-0.3em}
    \includegraphics[width=0.25\textwidth]{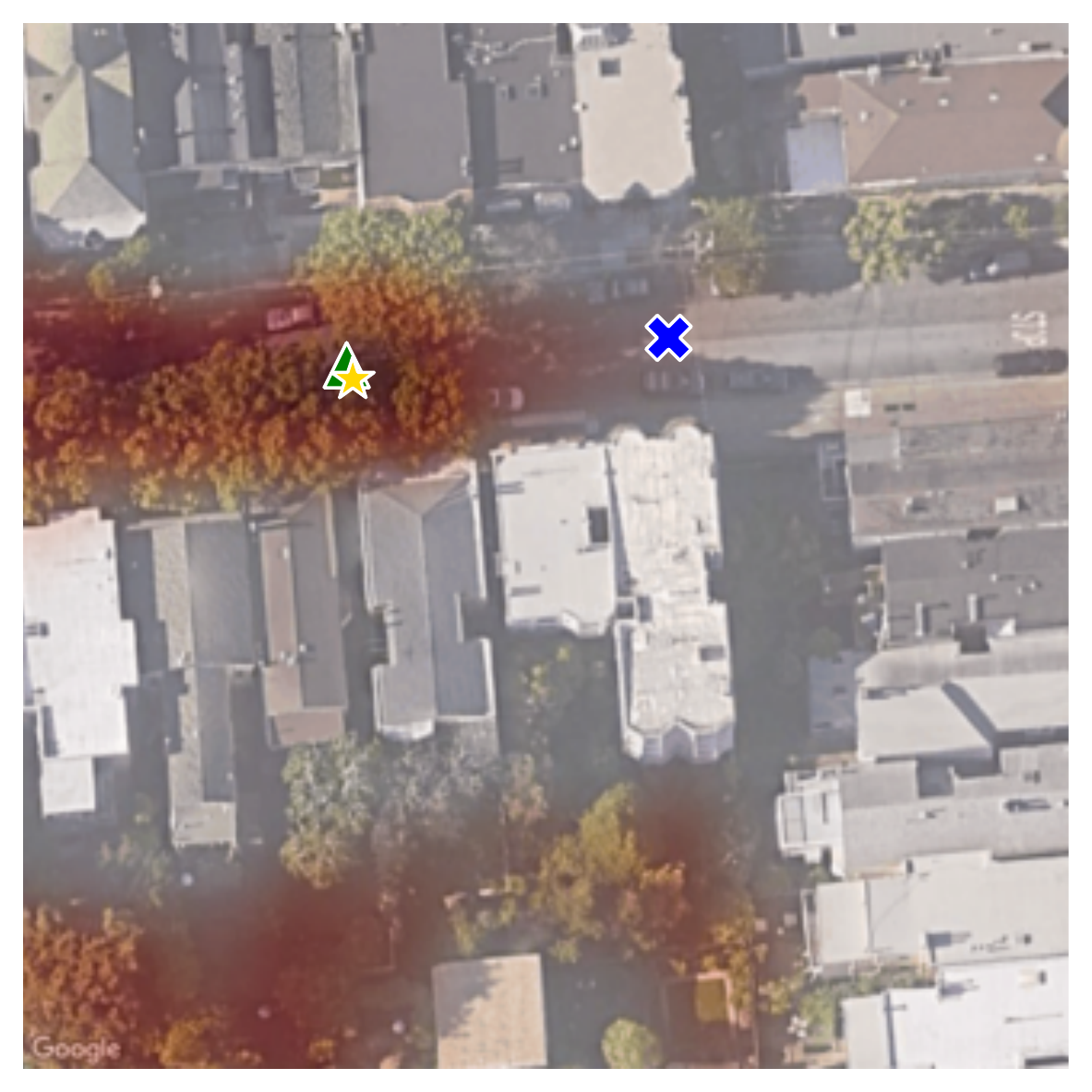}
    \caption{Qualitative results on VIGOR dataset cross-area split. 
    Top: input ground images and matching score maps at the model bottleneck, bottom: input satellite image overlayed with outputs from CVR and our method.
    }
    \label{fig:VIGOR_crossarea}
\end{figure}

In addition, we provide more qualitative cross-view metric localization results in the same-area and cross-area splits in Figure~\ref{fig:VIGOR_samearea} and~\ref{fig:VIGOR_crossarea}.

\begin{figure}[t]
    \centering    
    \hspace*{-0.5em}
    \includegraphics[width=0.17\textwidth]{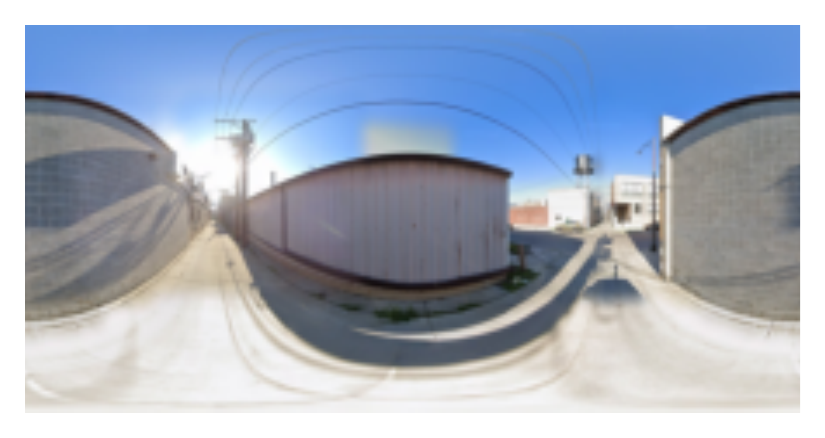}\hspace*{-0.4em}
    \includegraphics[width=0.085\textwidth]{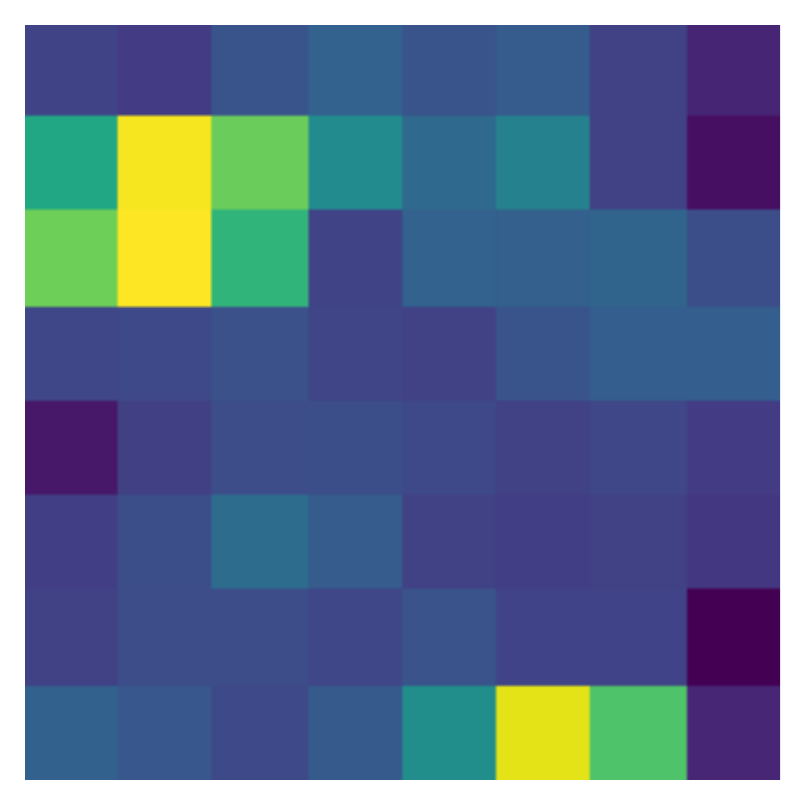}\hspace*{-0.4em}
    \includegraphics[width=0.17\textwidth]{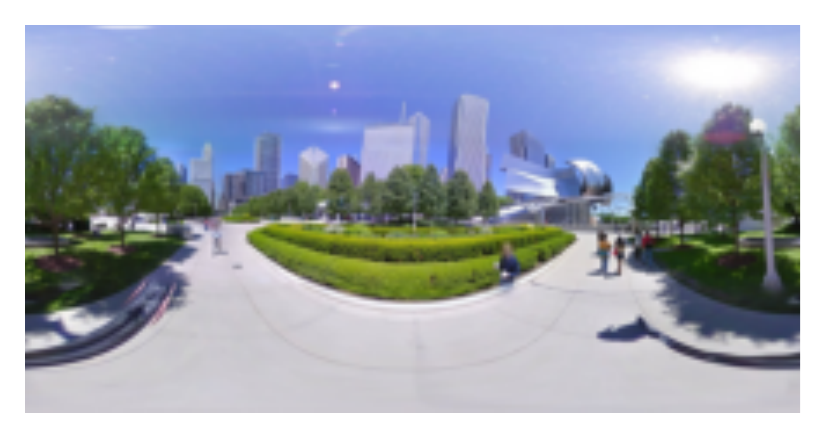}\hspace*{-0.4em}
     \includegraphics[width=0.085\textwidth]{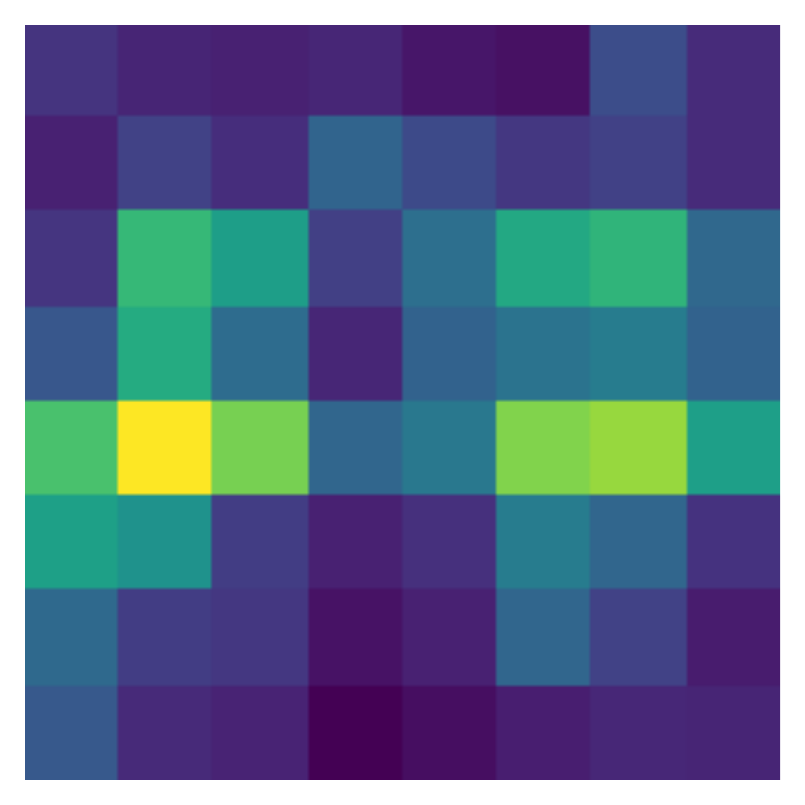}\hspace*{-0.4em}
    \includegraphics[width=0.17\textwidth]{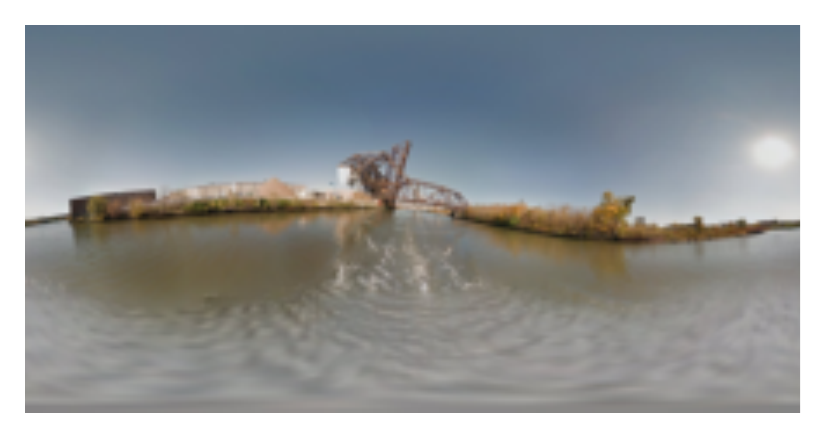}\hspace*{-0.4em}
     \includegraphics[width=0.085\textwidth]{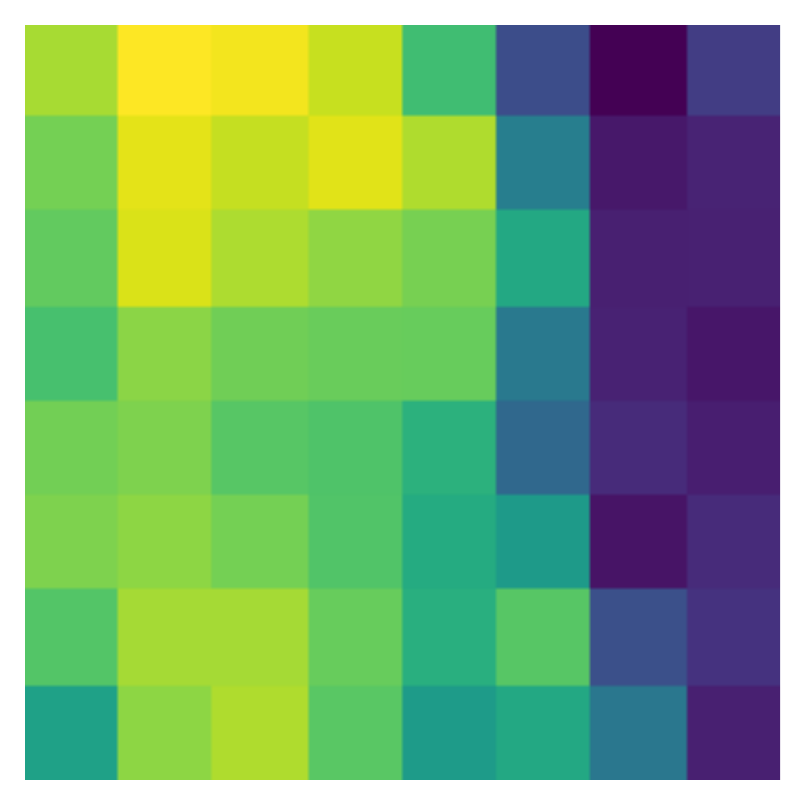}\hspace*{-0.4em}
    \includegraphics[width=0.17\textwidth]{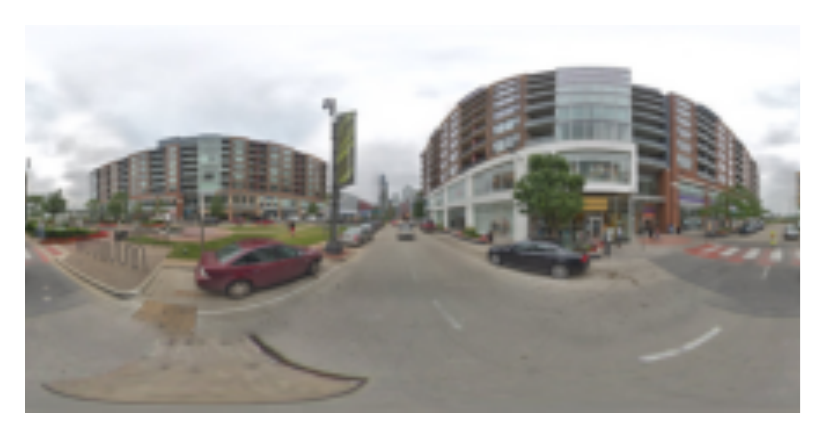}\hspace*{-0.4em}
     \includegraphics[width=0.085\textwidth]{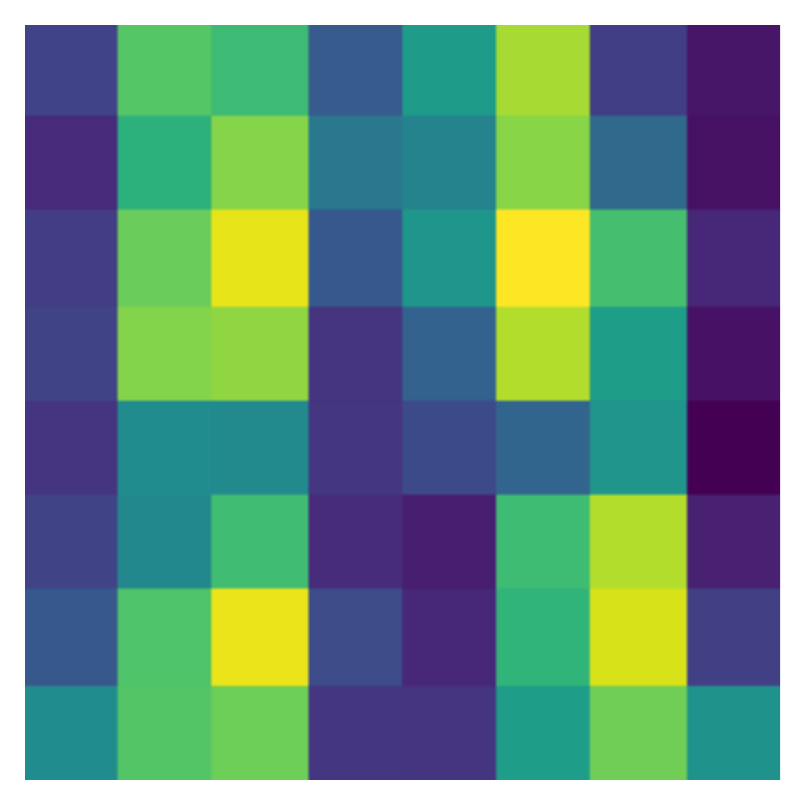}
    \\%\vspace*{-0.3em}
    \includegraphics[width=0.25\textwidth]{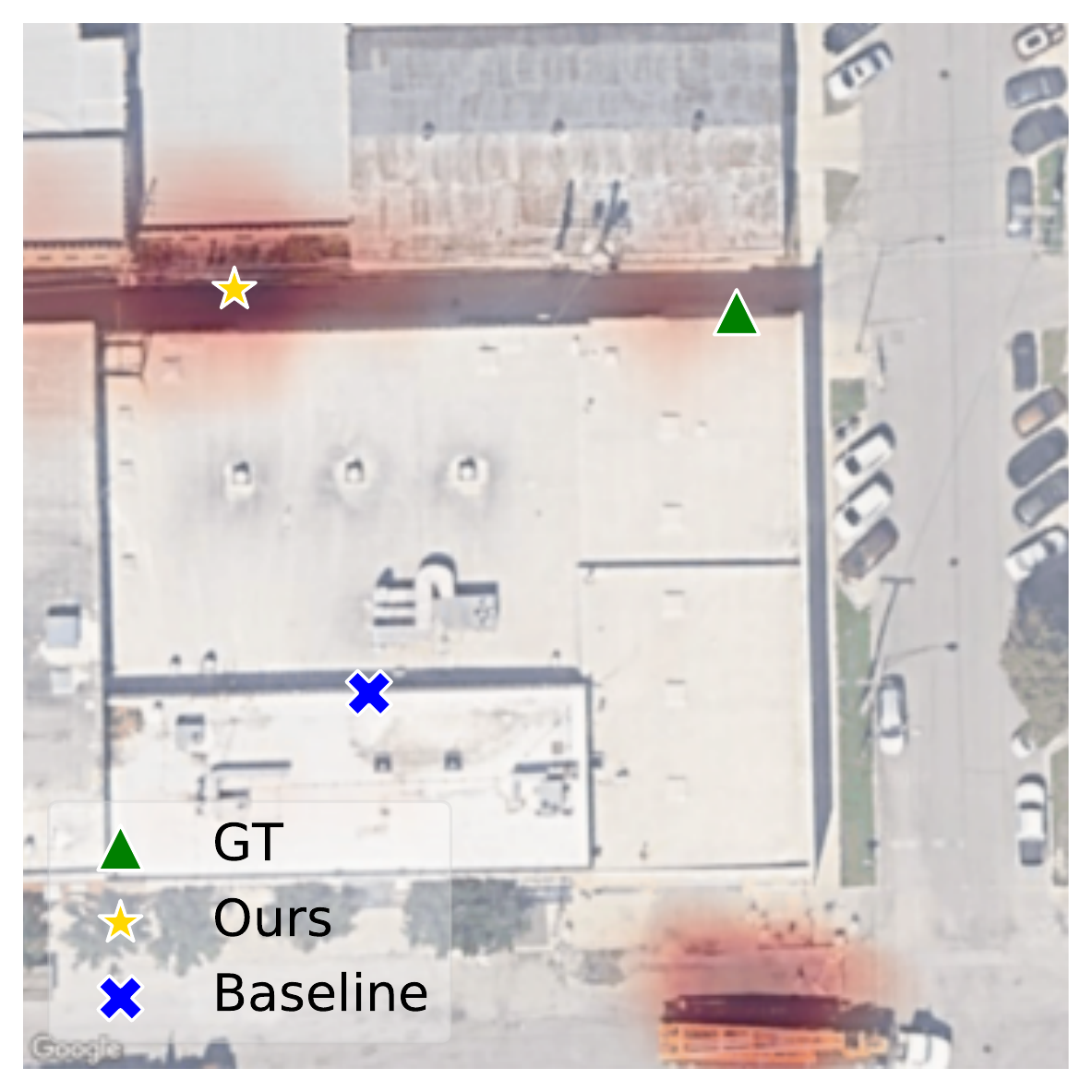}\hspace*{-0.3em}
    \includegraphics[width=0.25\textwidth]{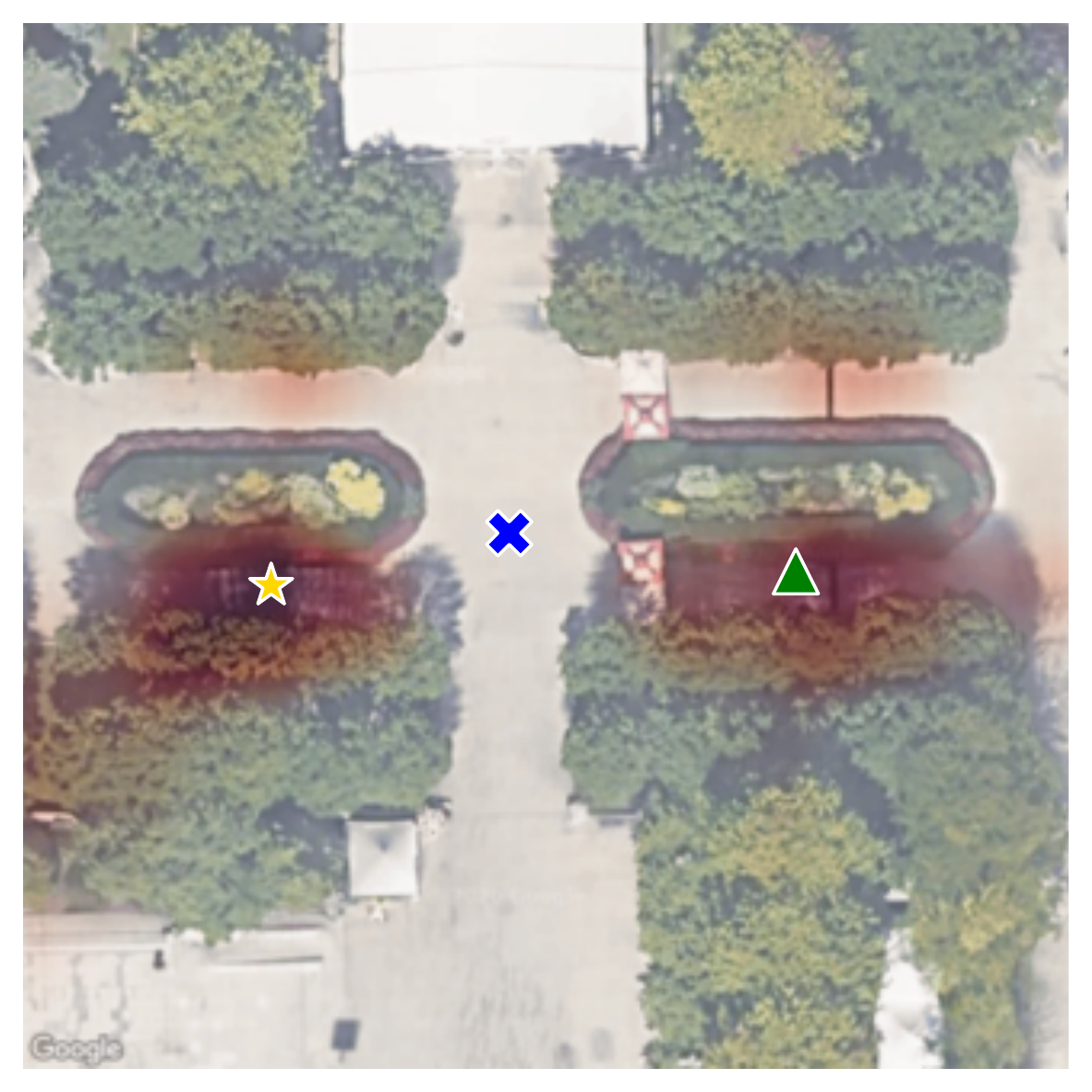}\hspace*{-0.3em}
    \includegraphics[width=0.25\textwidth]{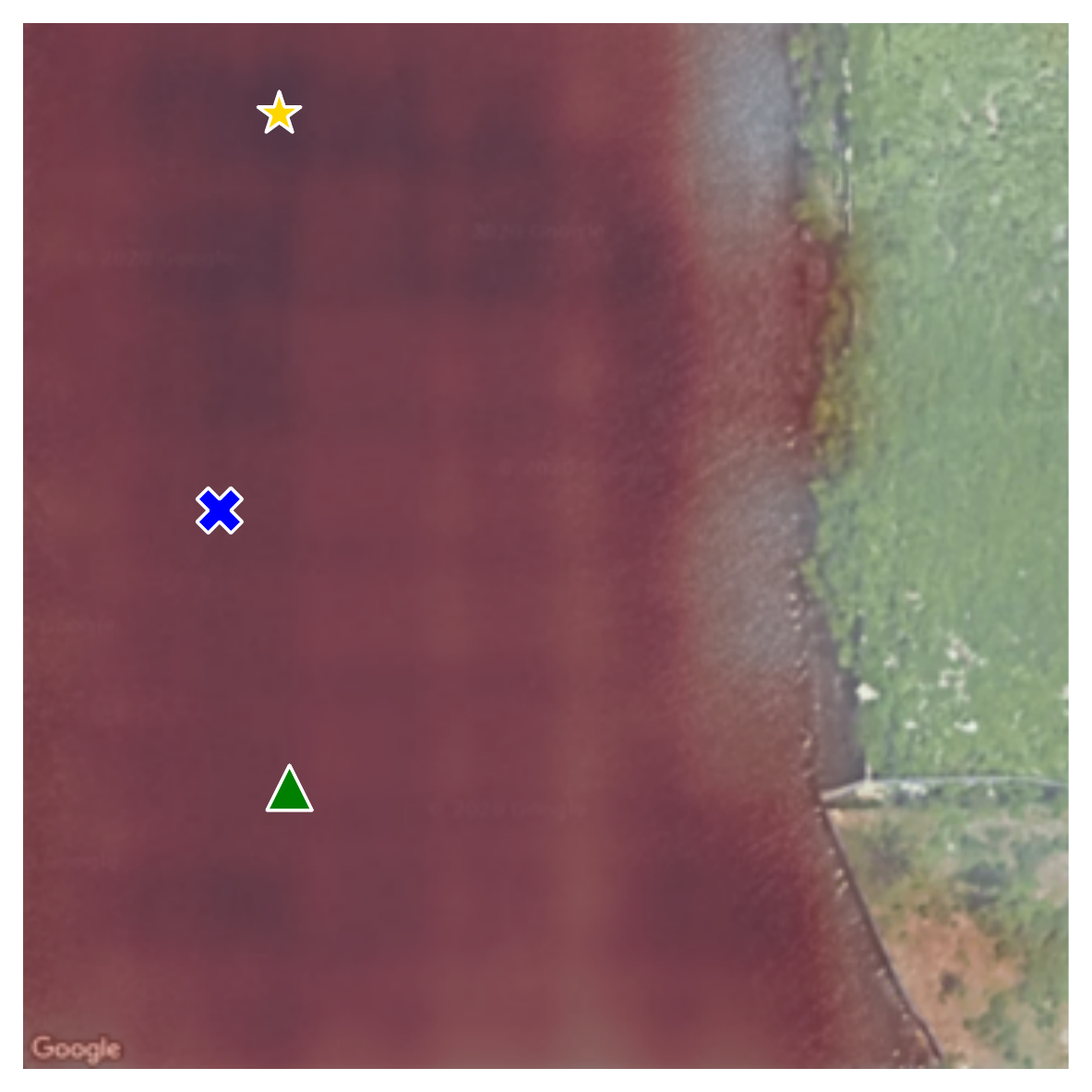}\hspace*{-0.3em}
    \includegraphics[width=0.25\textwidth]{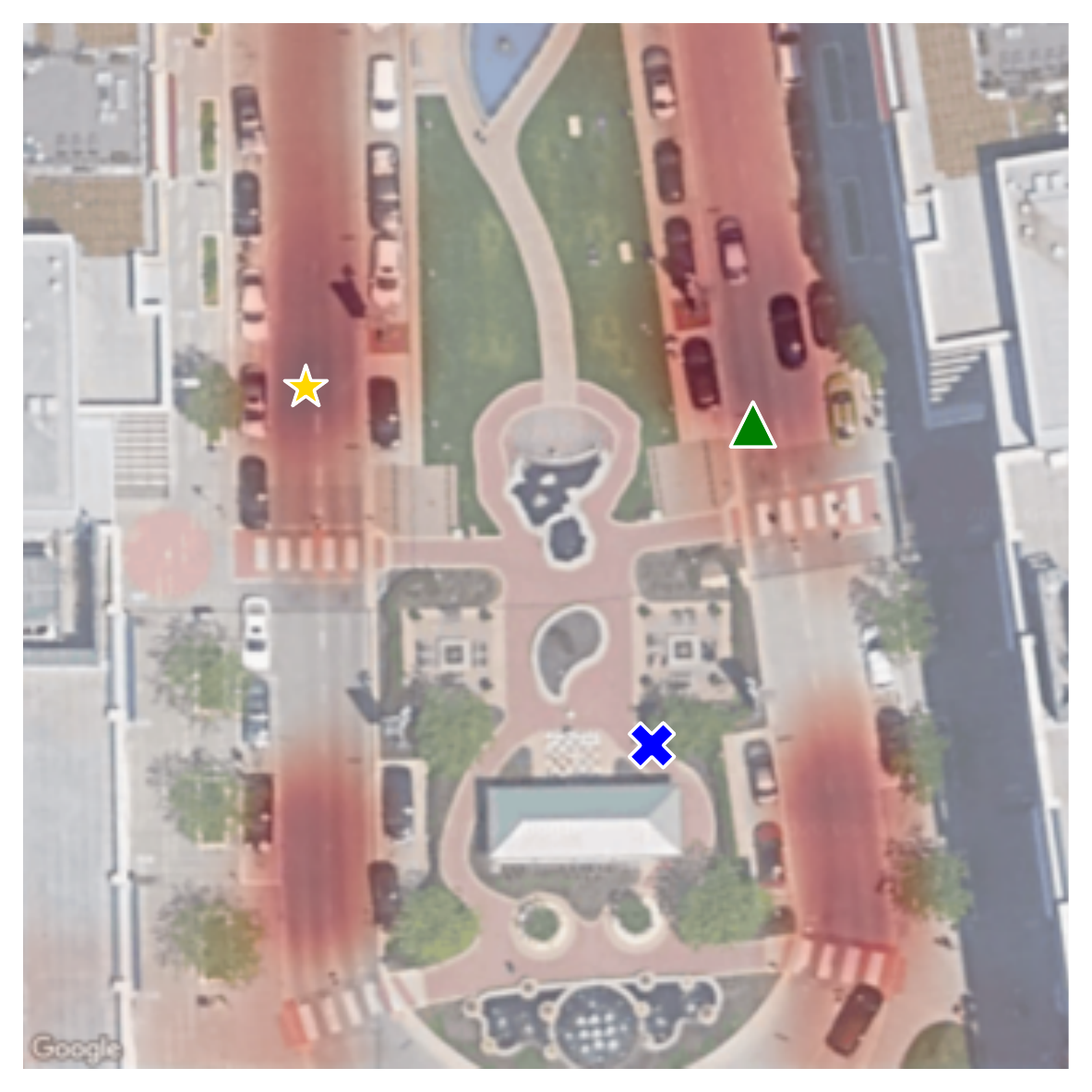}
    \caption{Failure cases on VIGOR dataset. 
    Top: input ground images and matching score maps at the model bottleneck, bottom: input satellite image overlayed with outputs from CVR and our method.
    }
    \label{fig:VIGOR_failure}
\end{figure}

As pointed out in the \textbf{main paper Section 4.4.}, our method has slightly more outliers than the CVR.
In Figure~\ref{fig:VIGOR_failure}, we show some of the failure cases where our localization error is higher than CVR.
Importantly, as we stated in the main paper, instead of regressing to a location closer to the ground truth location as CVR, our method expresses the underlying uncertainty.
This property is desirable in real-world applications since the ground truth location is also captured plus our prediction expresses the model's underlying uncertainty.

\begin{figure}
    \centering
    \hspace*{-0.5em}
    \includegraphics[width=0.17\textwidth]{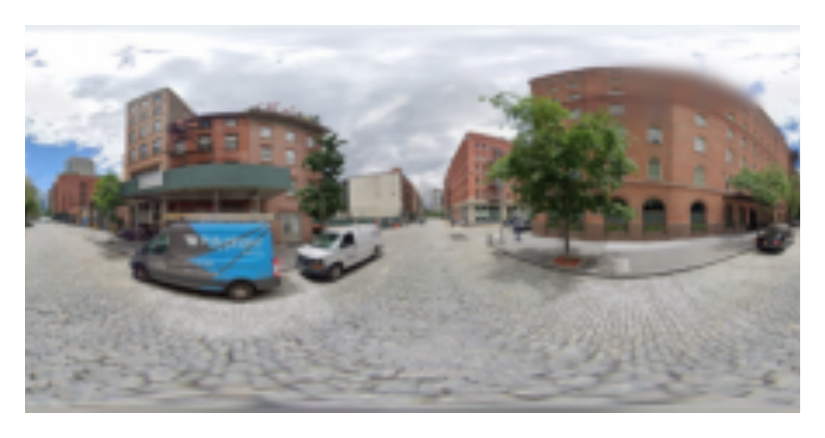}\hspace*{-0.4em}
    \includegraphics[width=0.085\textwidth]{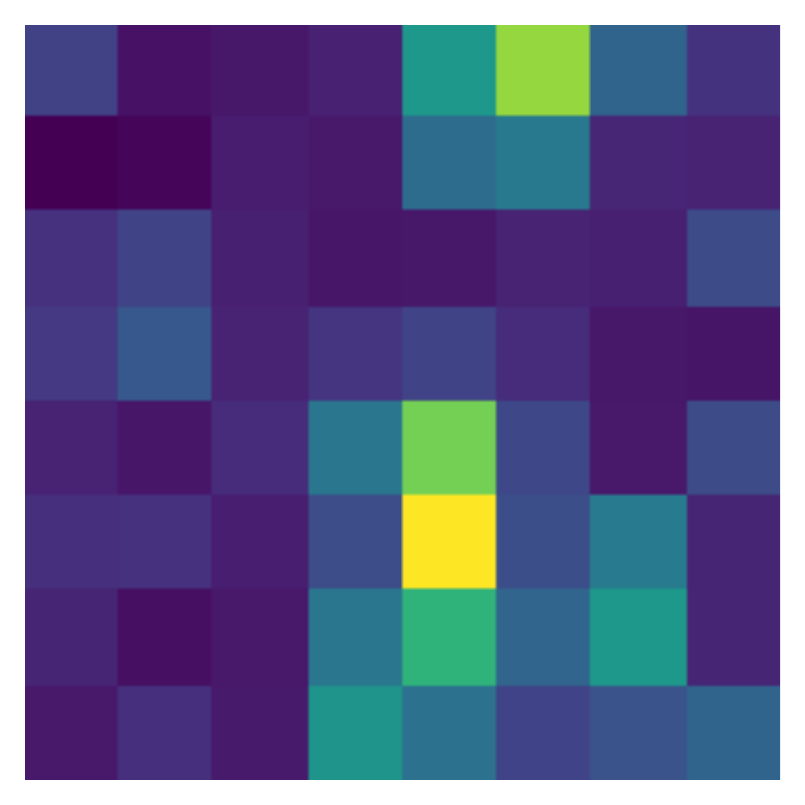}\hspace*{-0.4em}
    \includegraphics[width=0.17\textwidth]{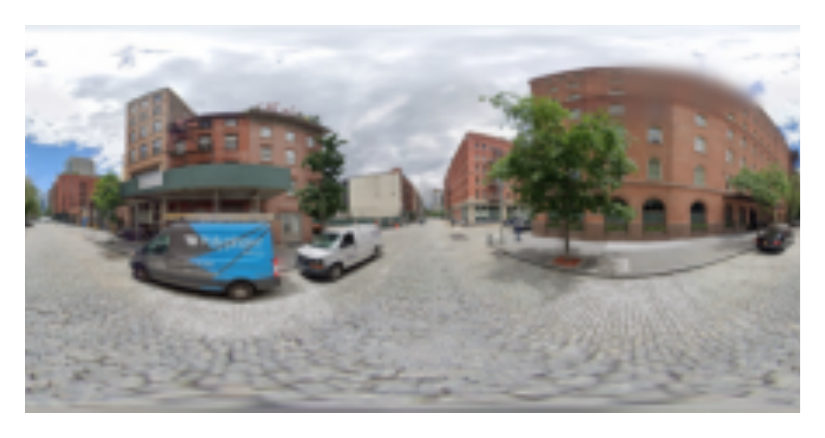}\hspace*{-0.4em}
     \includegraphics[width=0.085\textwidth]{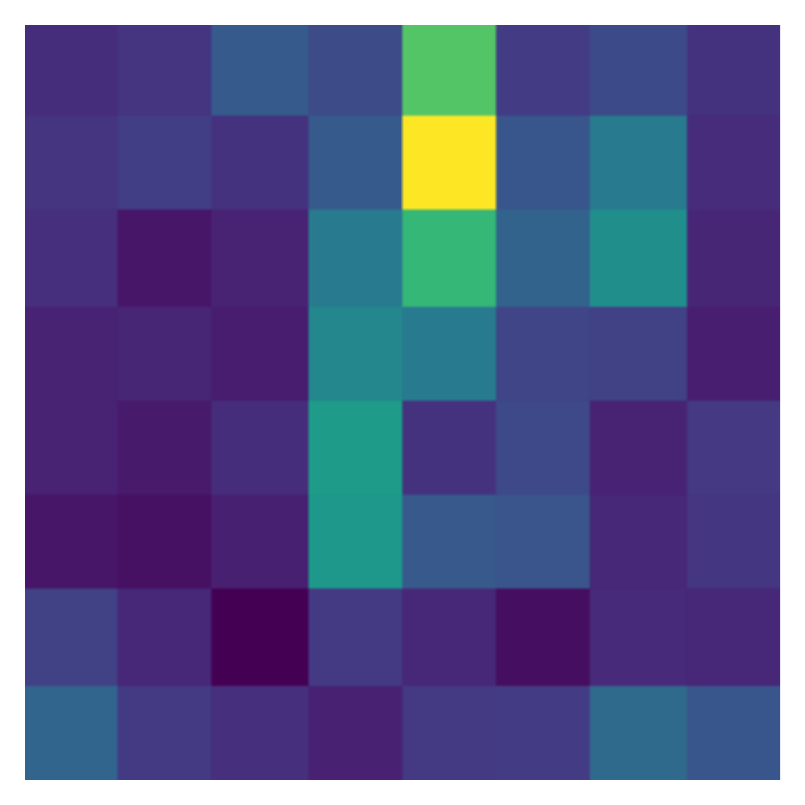}\hspace*{-0.4em}
    \includegraphics[width=0.17\textwidth]{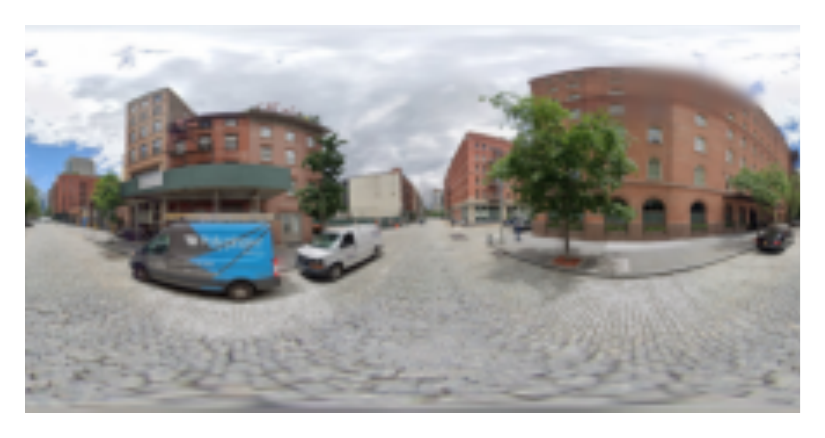}\hspace*{-0.4em}
     \includegraphics[width=0.085\textwidth]{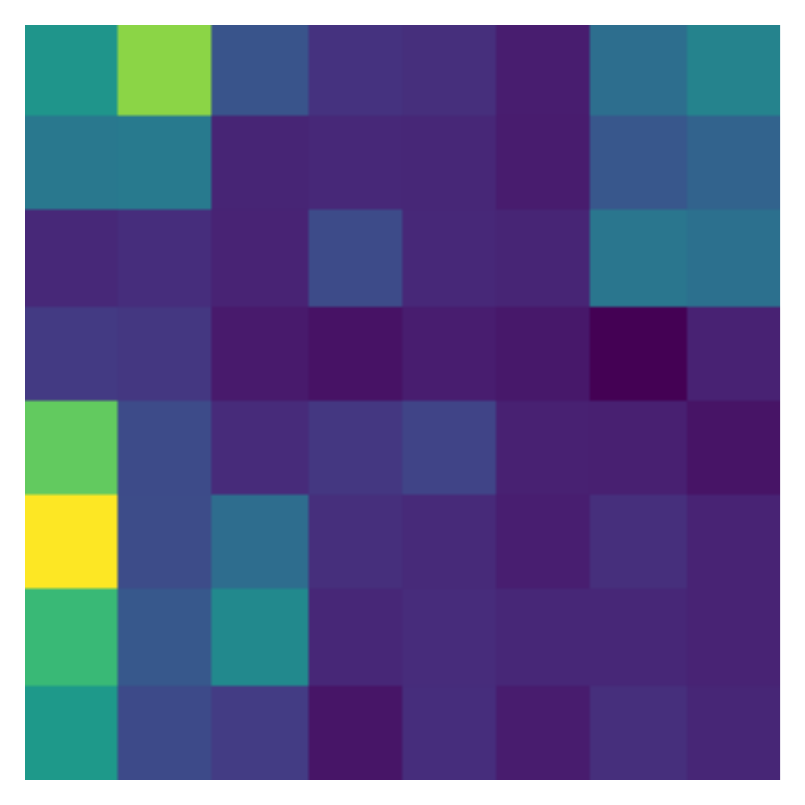}\hspace*{-0.4em}
    \includegraphics[width=0.17\textwidth]{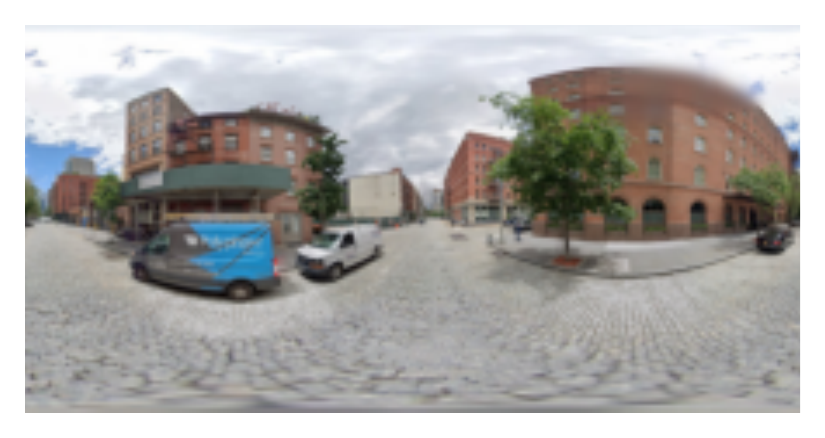}\hspace*{-0.4em}
     \includegraphics[width=0.085\textwidth]{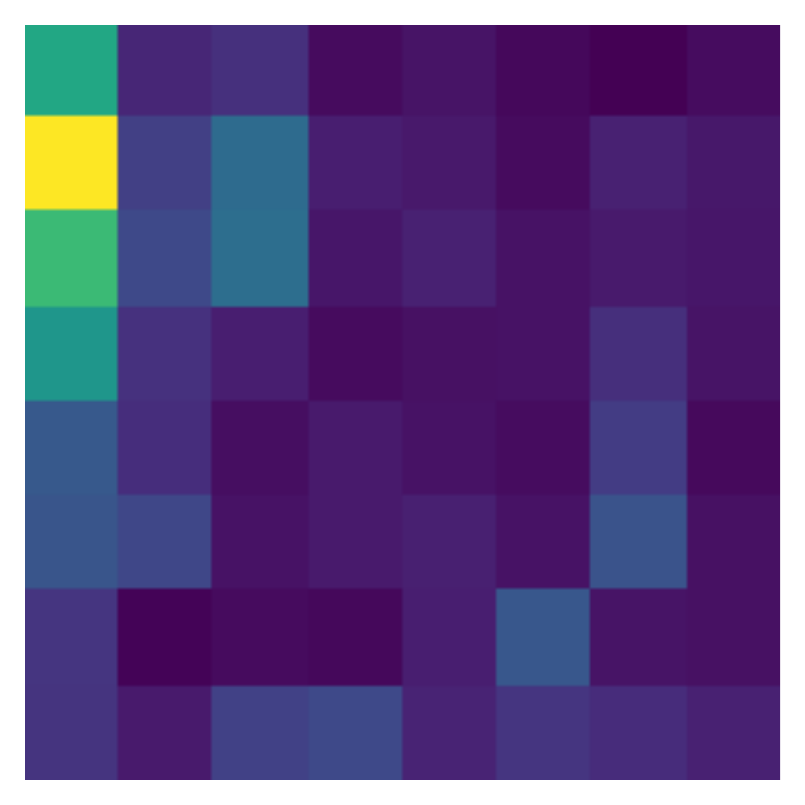}
    \\%\vspace*{-0.3em}
    \includegraphics[width=0.25\textwidth]{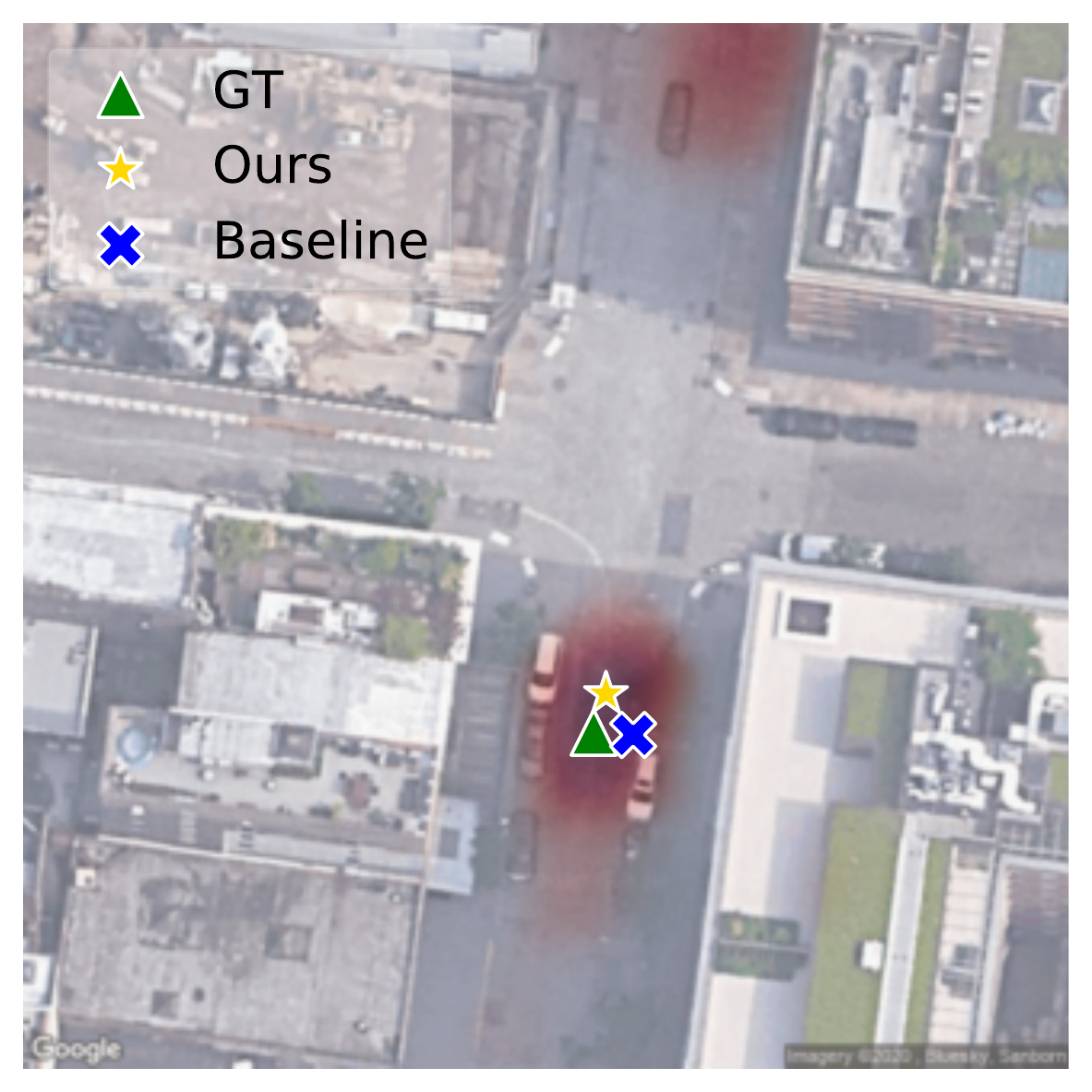}\hspace*{-0.3em}
    \includegraphics[width=0.25\textwidth]{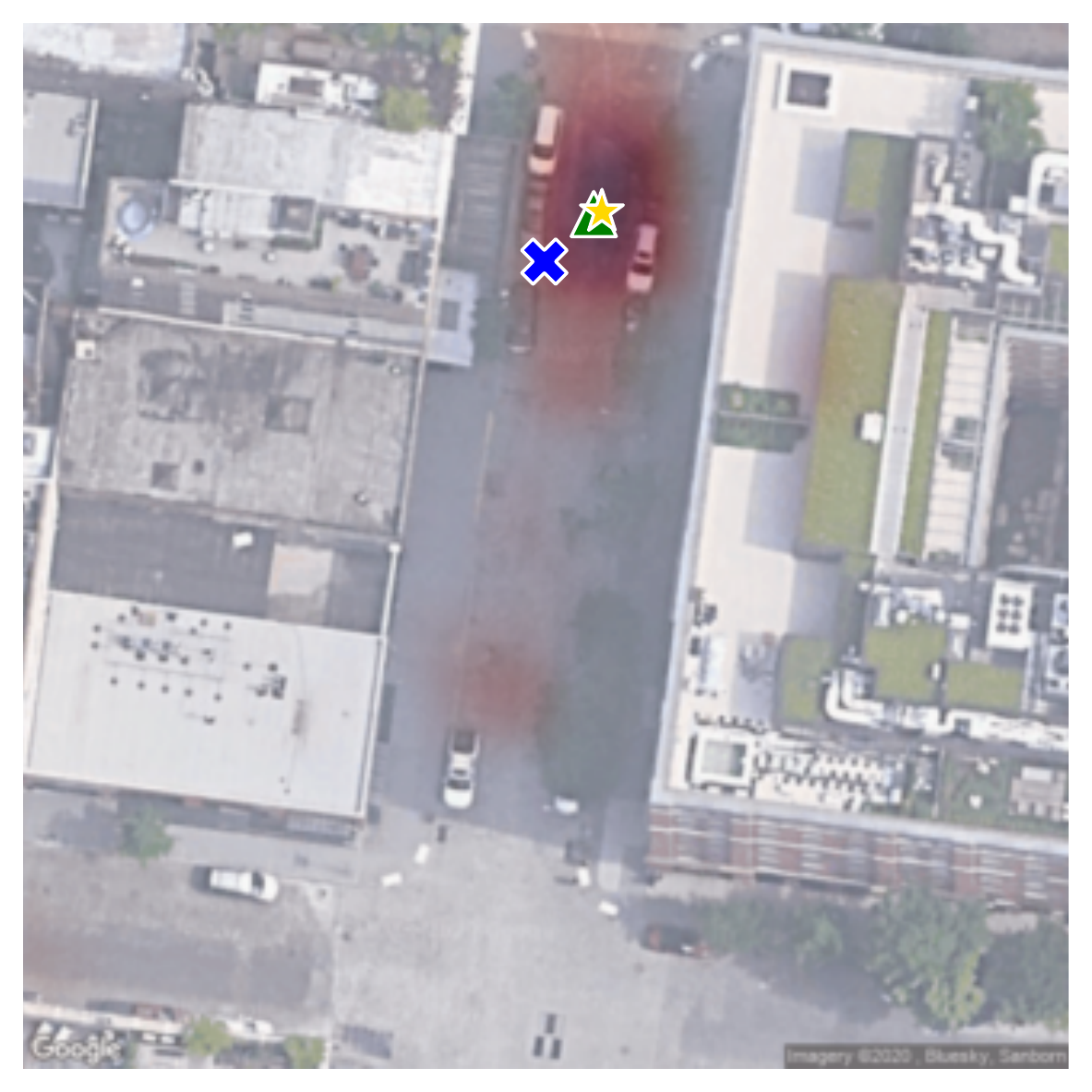}\hspace*{-0.3em}
    \includegraphics[width=0.25\textwidth]{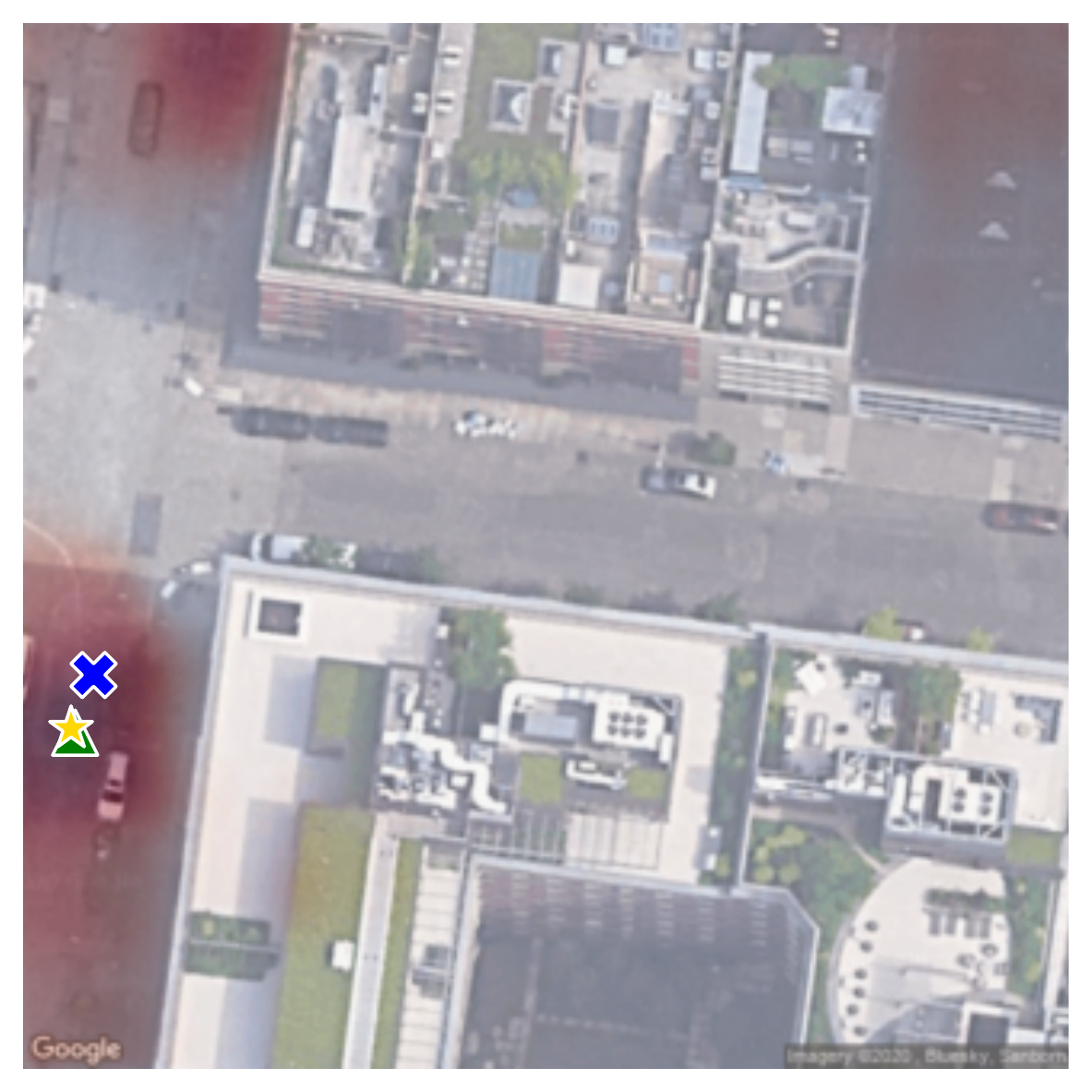}\hspace*{-0.3em}
    \includegraphics[width=0.25\textwidth]{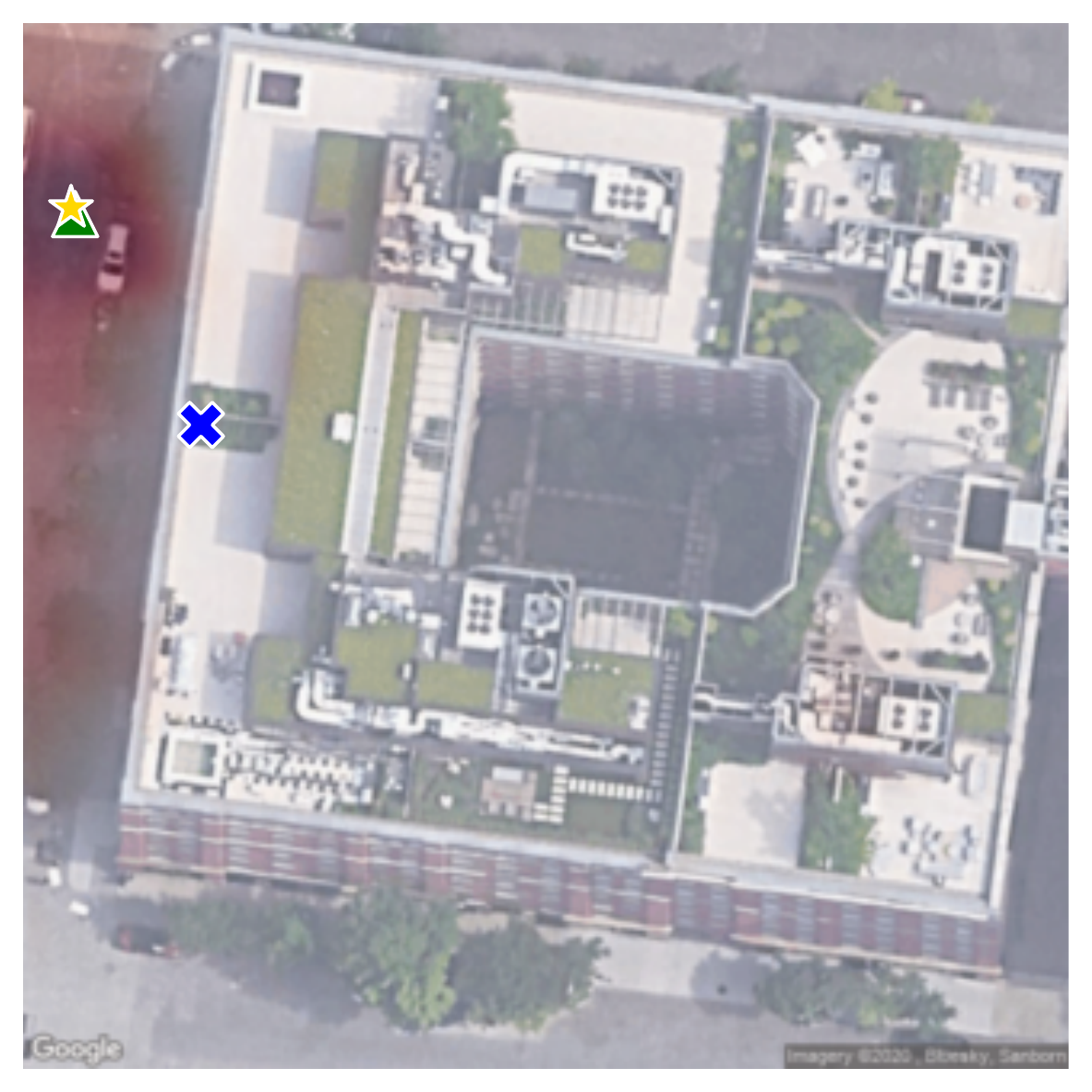}
    \\\hspace*{-0.5em}
    \includegraphics[width=0.17\textwidth]{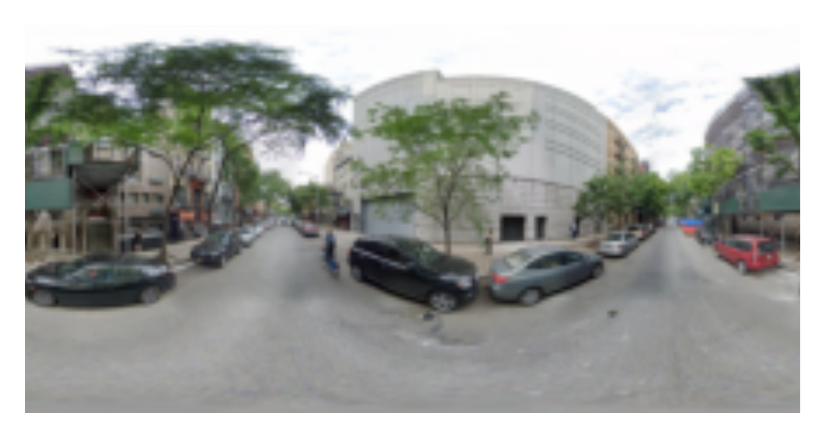}\hspace*{-0.4em}
    \includegraphics[width=0.085\textwidth]{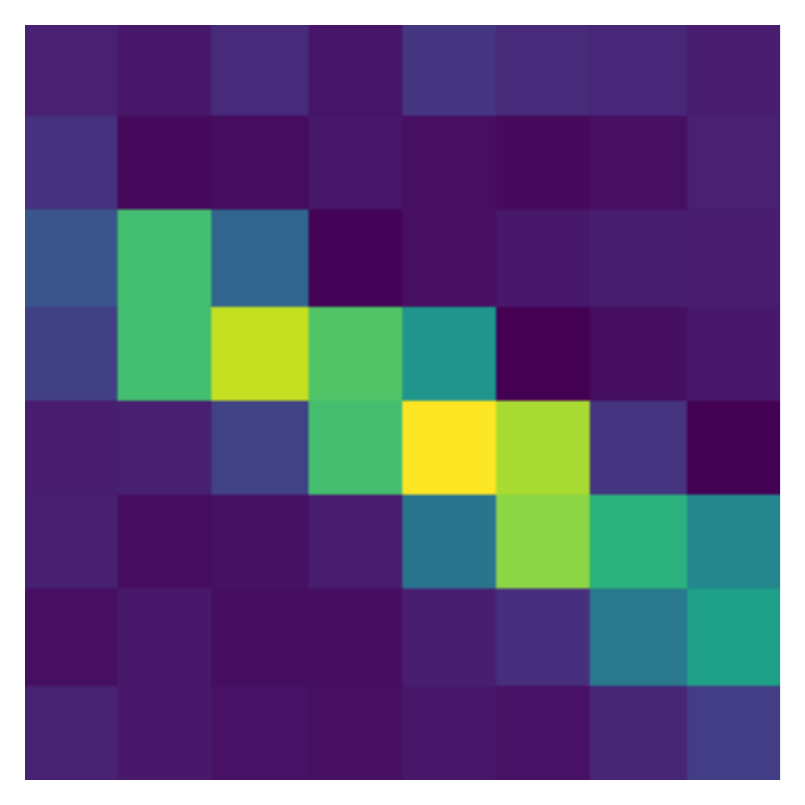}\hspace*{-0.4em}
    \includegraphics[width=0.17\textwidth]{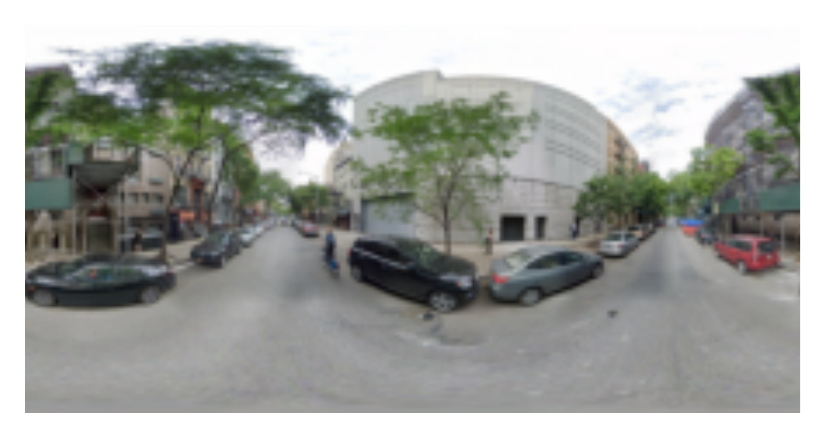}\hspace*{-0.4em}
     \includegraphics[width=0.085\textwidth]{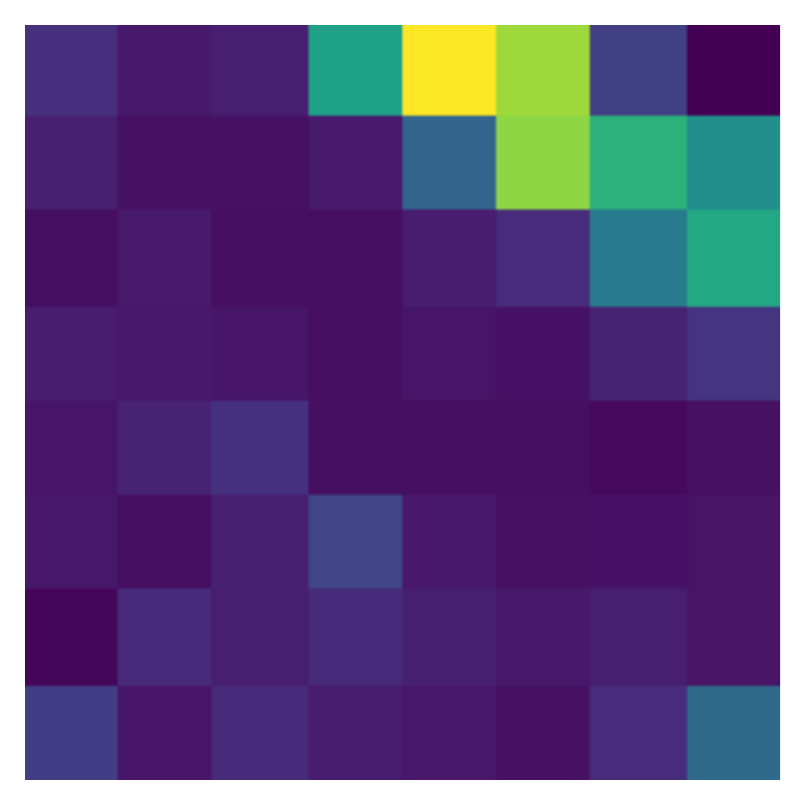}\hspace*{-0.4em}
    \includegraphics[width=0.17\textwidth]{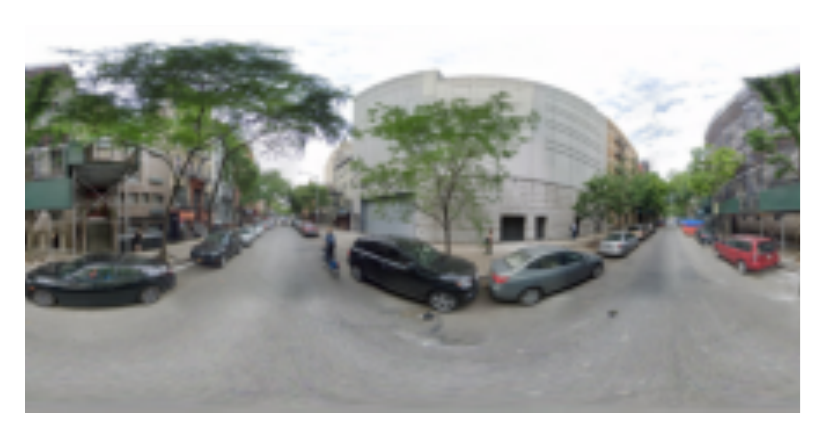}\hspace*{-0.4em}
     \includegraphics[width=0.085\textwidth]{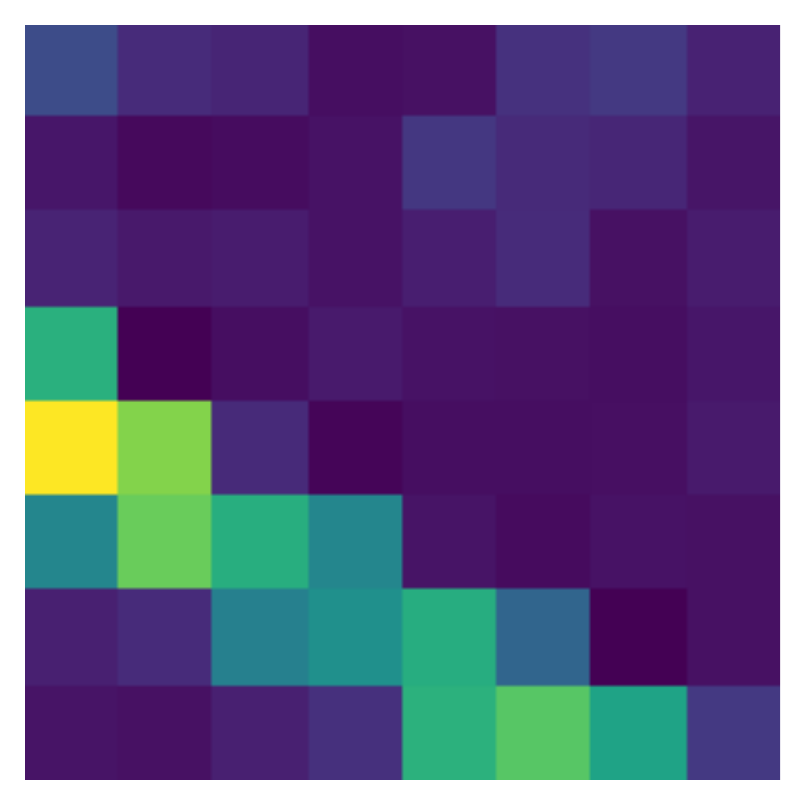}\hspace*{-0.4em}
    \includegraphics[width=0.17\textwidth]{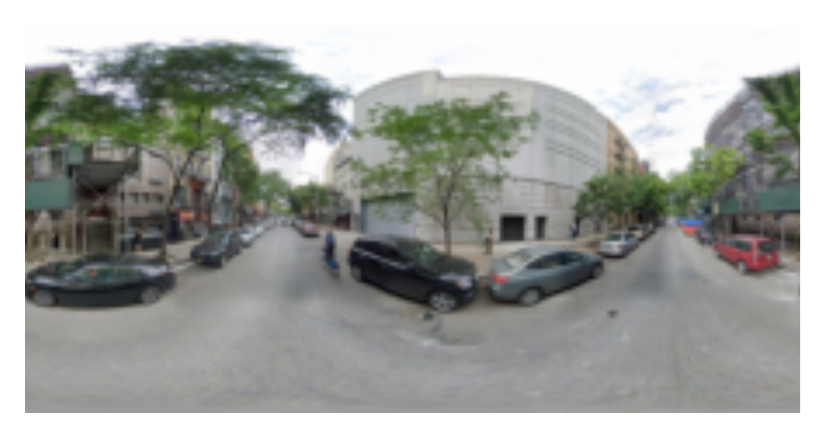}\hspace*{-0.4em}
     \includegraphics[width=0.085\textwidth]{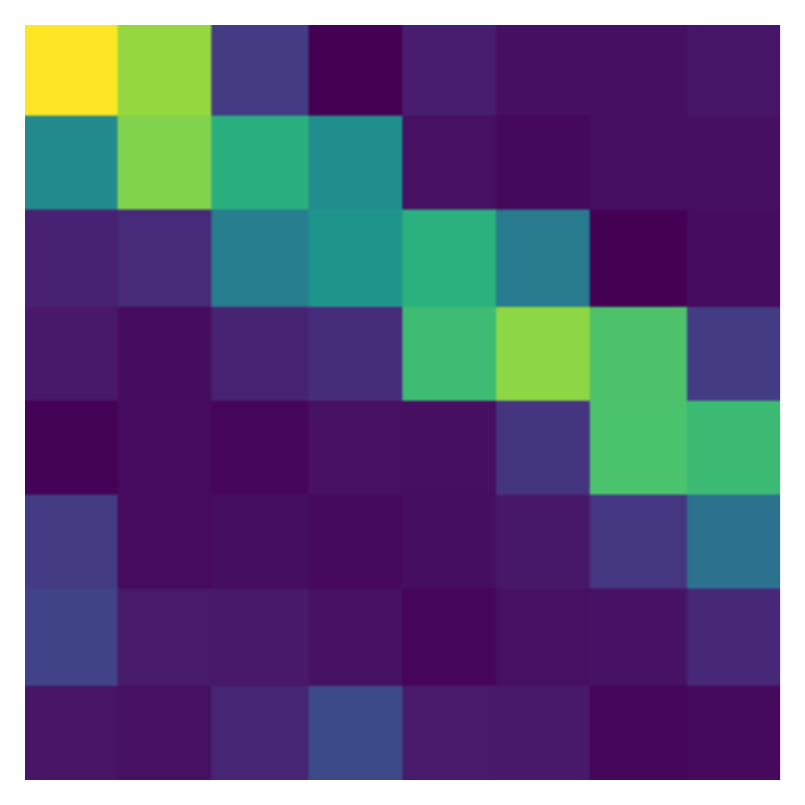}
    \\%\vspace*{-0.3em}
    \includegraphics[width=0.25\textwidth]{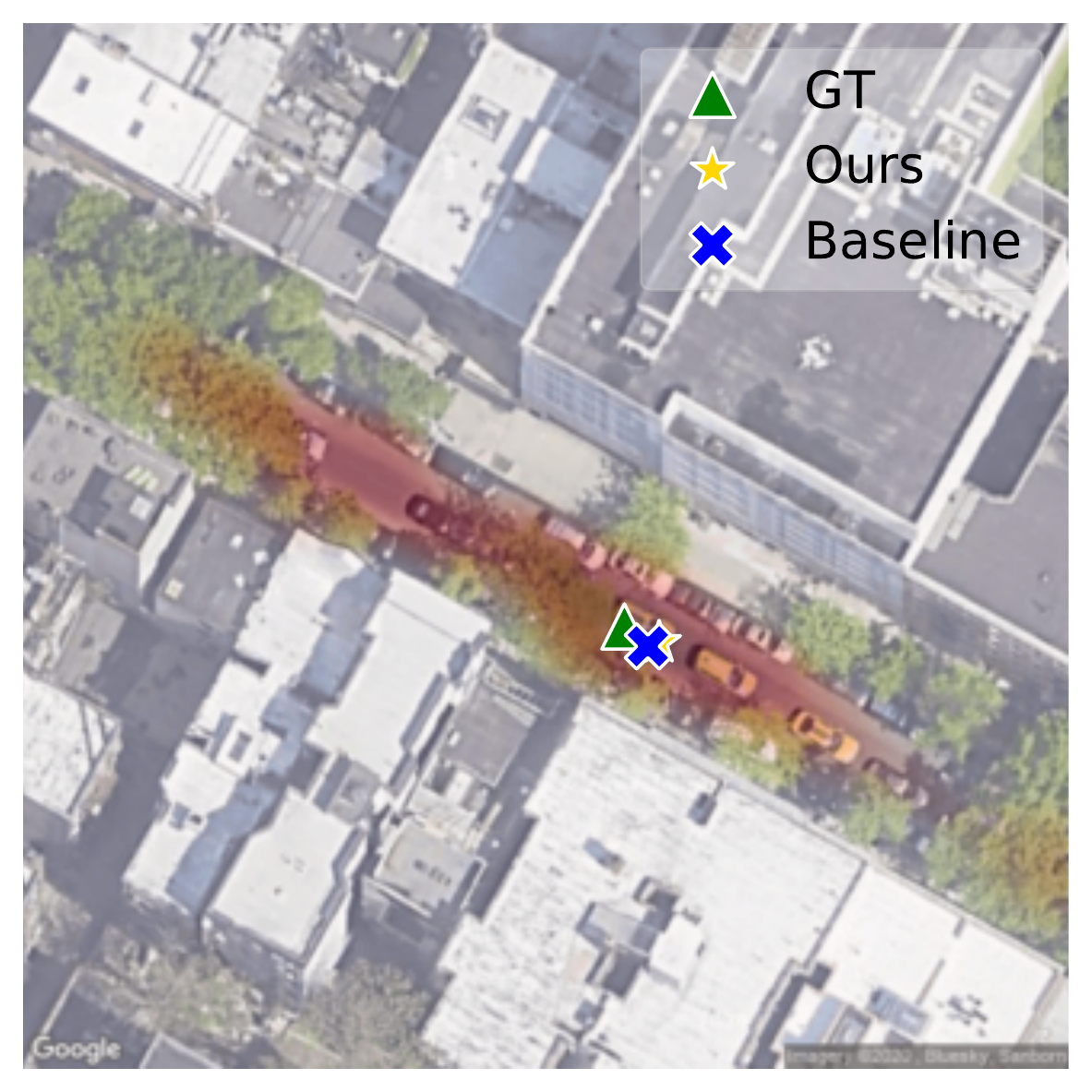}\hspace*{-0.3em}
    \includegraphics[width=0.25\textwidth]{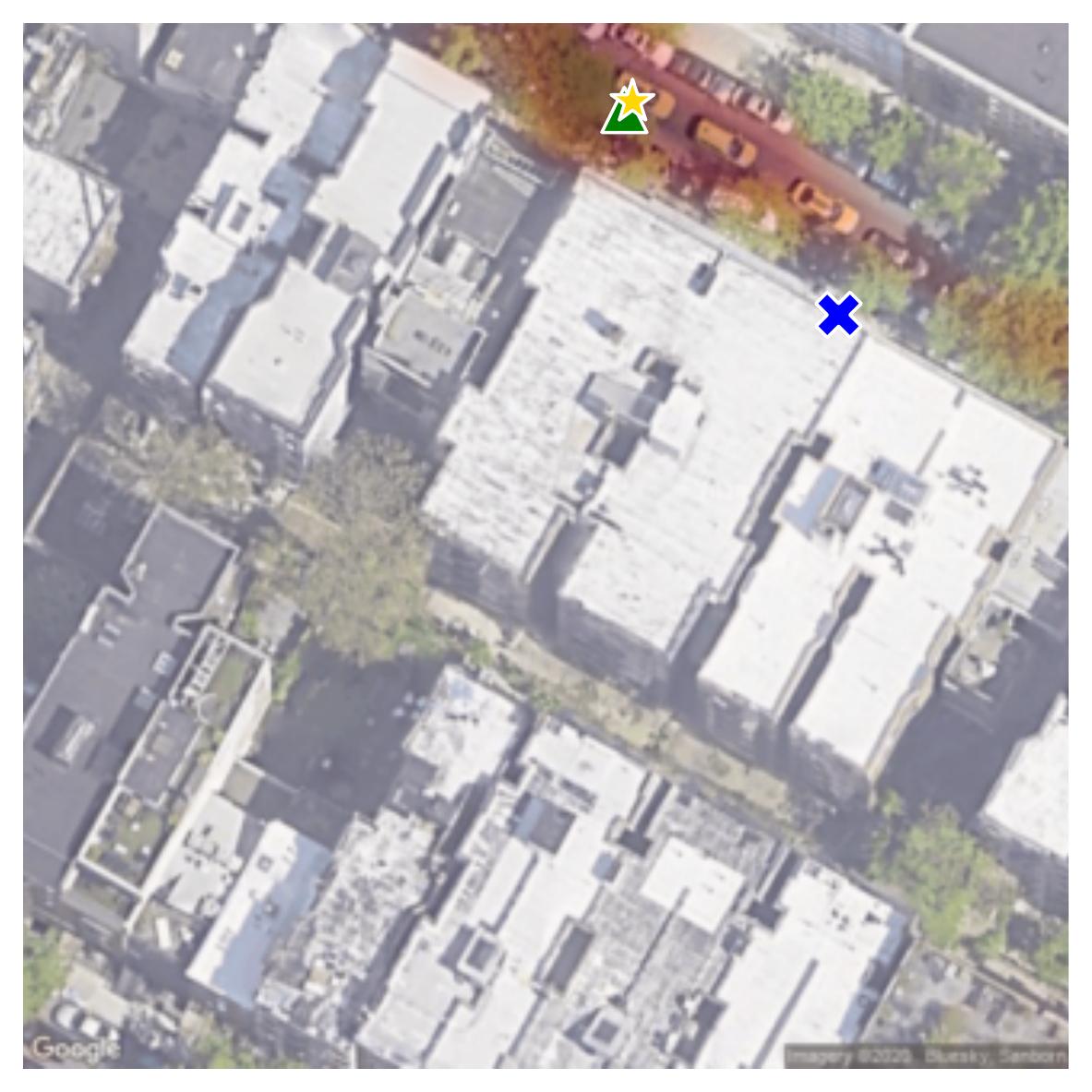}\hspace*{-0.3em}
    \includegraphics[width=0.25\textwidth]{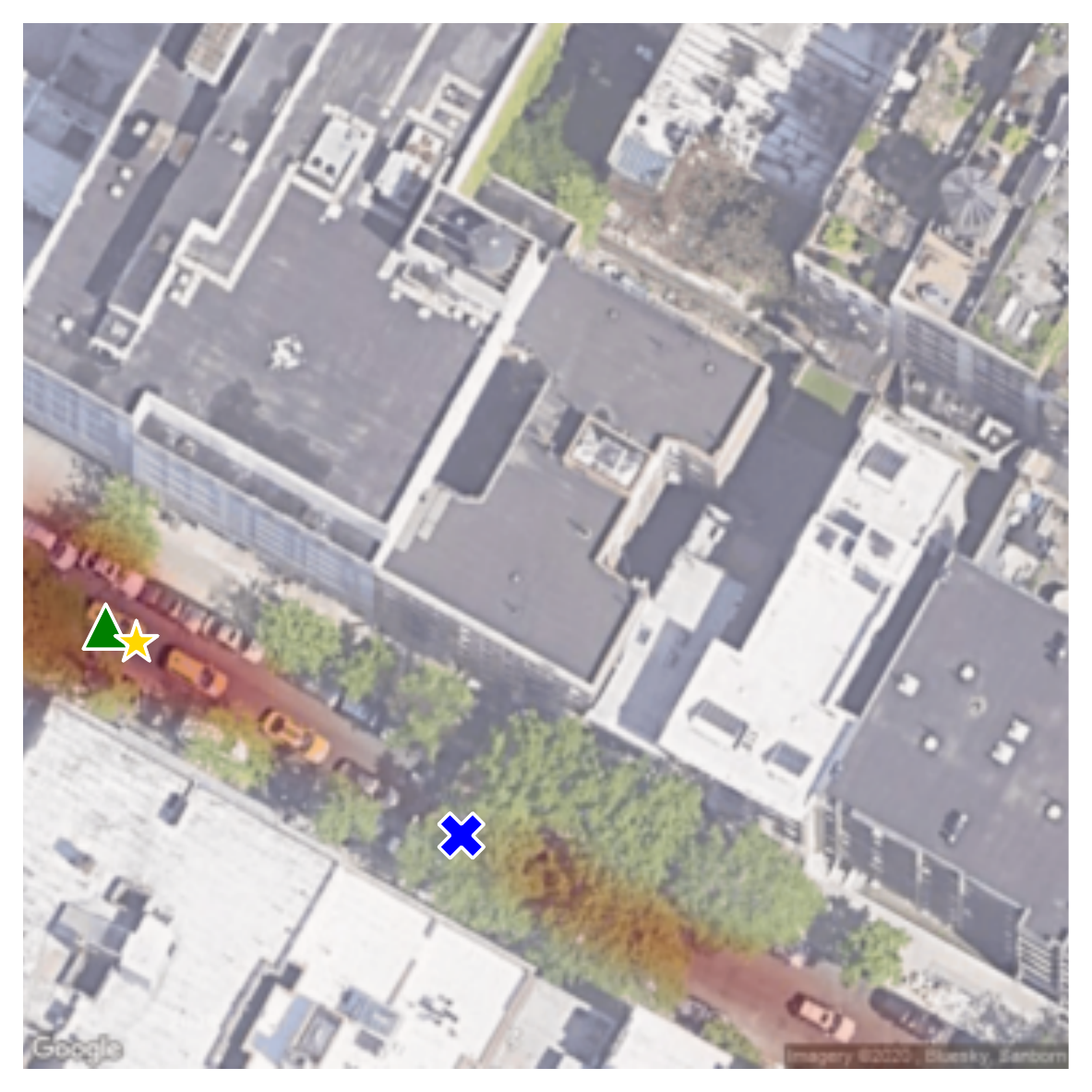}\hspace*{-0.3em}
    \includegraphics[width=0.25\textwidth]{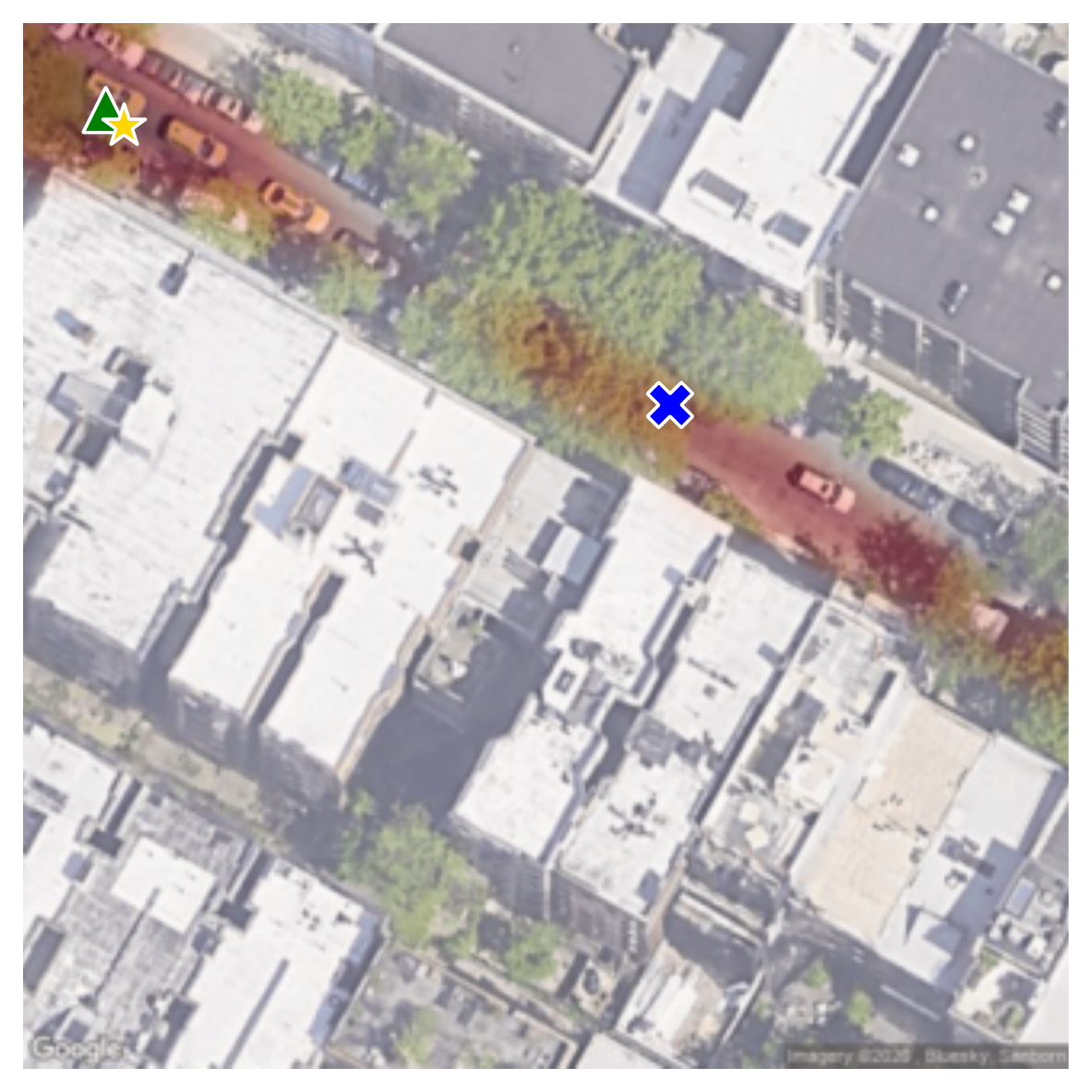}
    \\\hspace*{-0.5em}
    \includegraphics[width=0.17\textwidth]{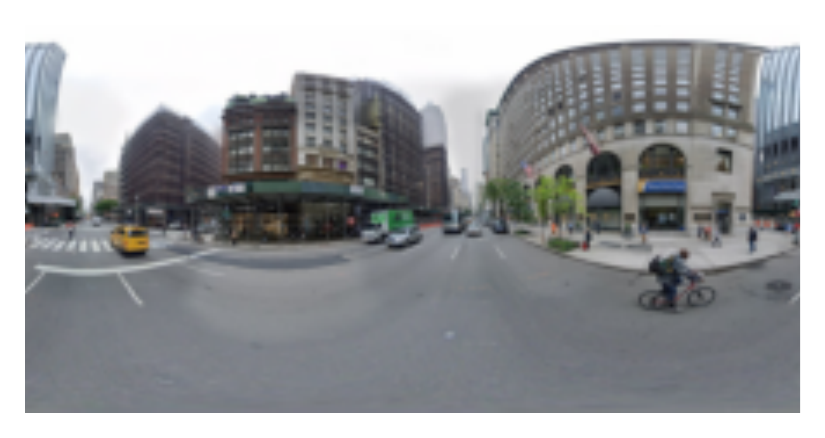}\hspace*{-0.4em}
    \includegraphics[width=0.085\textwidth]{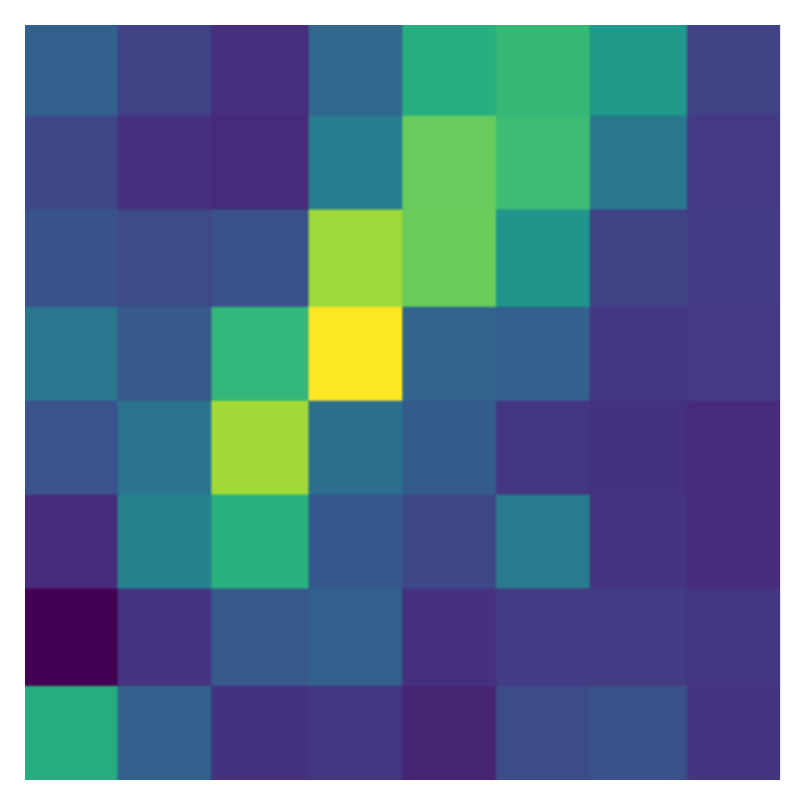}\hspace*{-0.4em}
    \includegraphics[width=0.17\textwidth]{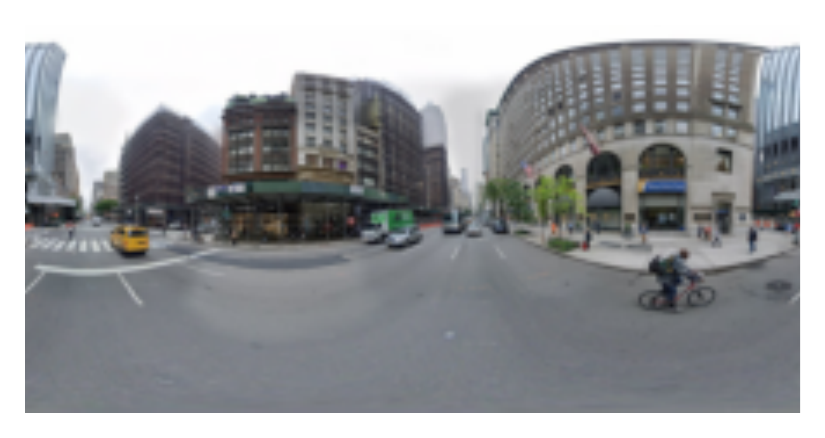}\hspace*{-0.4em}
     \includegraphics[width=0.085\textwidth]{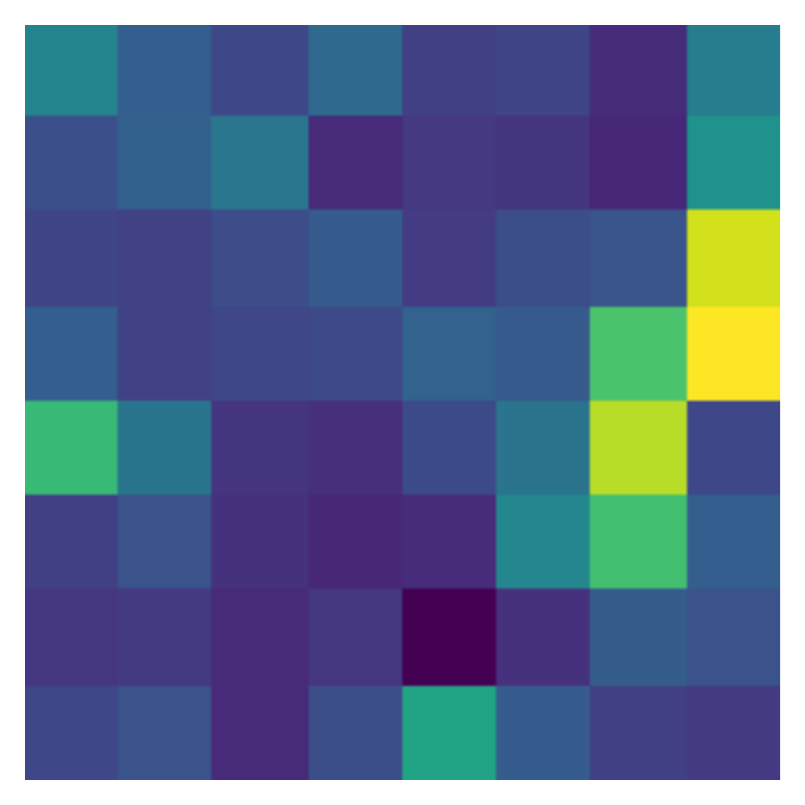}\hspace*{-0.4em}
    \includegraphics[width=0.17\textwidth]{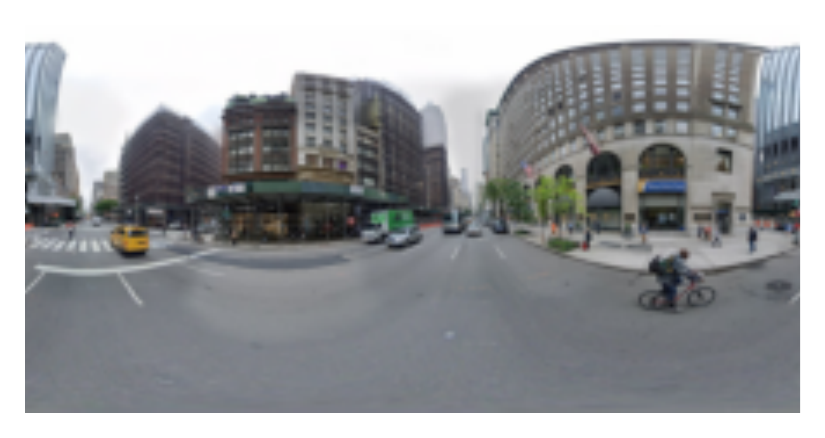}\hspace*{-0.4em}
     \includegraphics[width=0.085\textwidth]{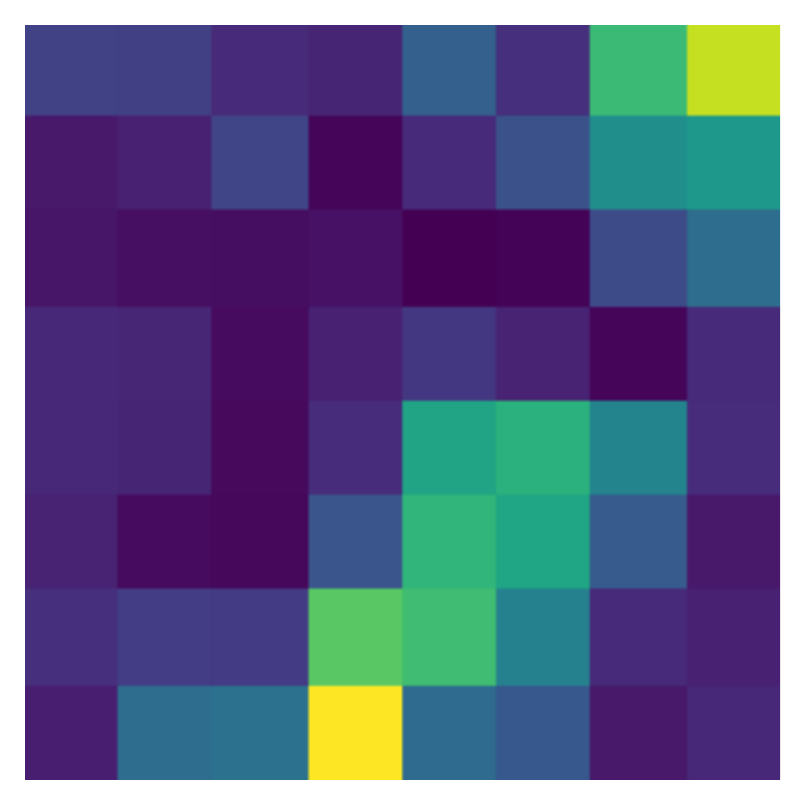}\hspace*{-0.4em}
    \includegraphics[width=0.17\textwidth]{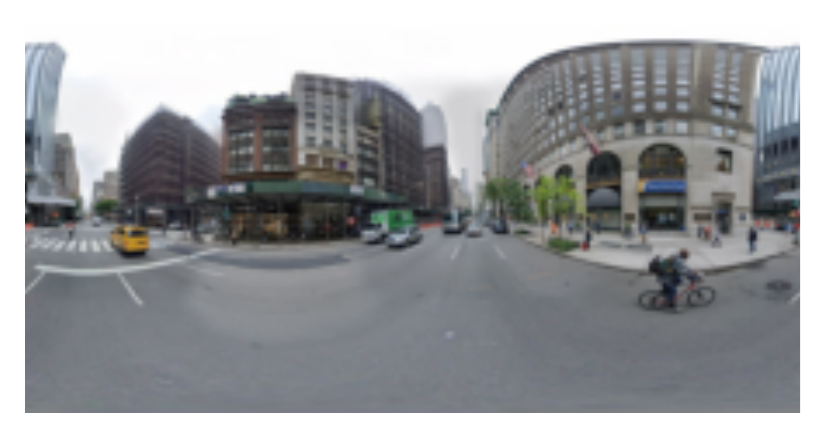}\hspace*{-0.4em}
     \includegraphics[width=0.085\textwidth]{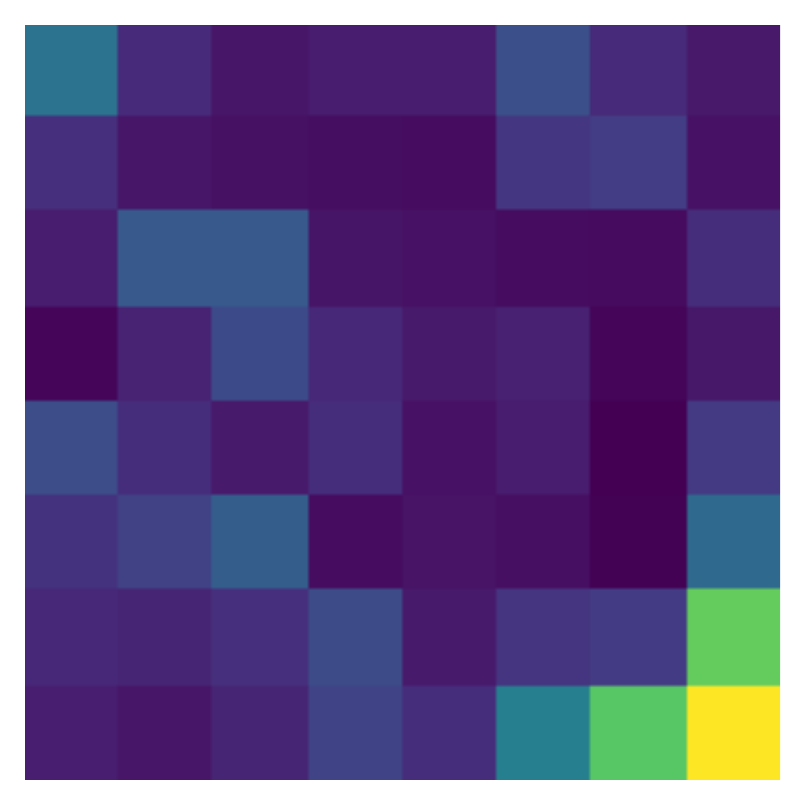}
    \\%\vspace*{-0.3em}
    \includegraphics[width=0.25\textwidth]{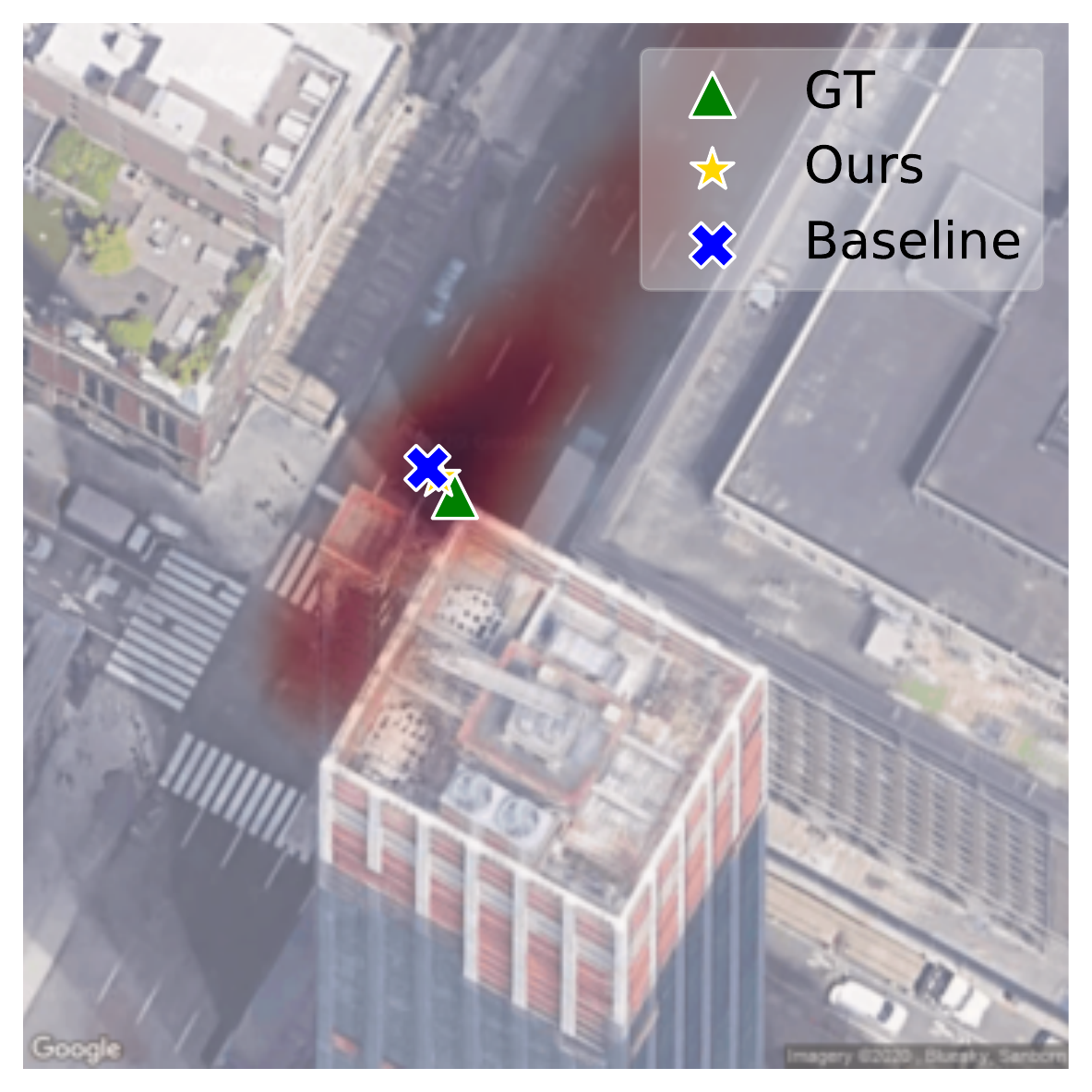}\hspace*{-0.3em}
    \includegraphics[width=0.25\textwidth]{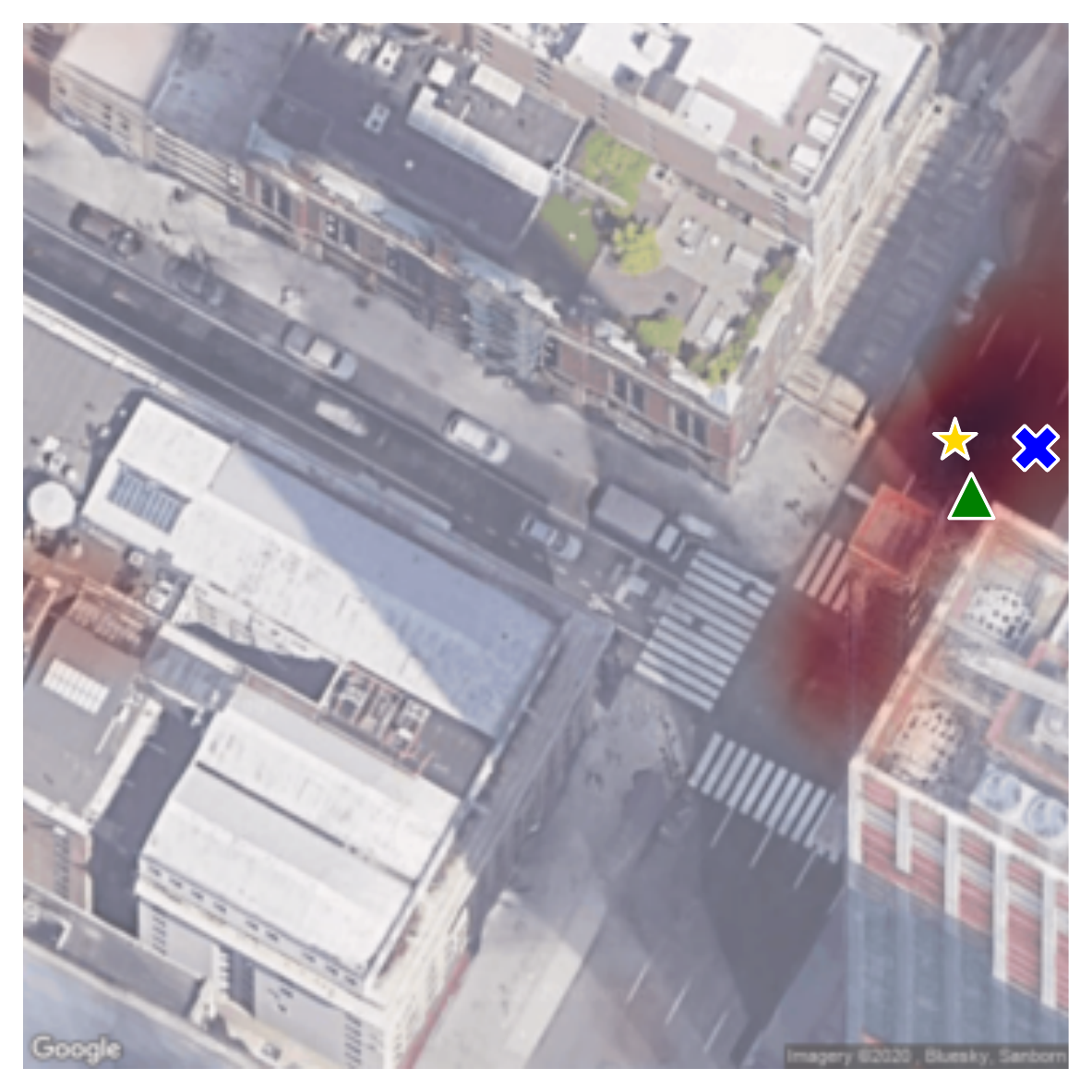}\hspace*{-0.3em}
    \includegraphics[width=0.25\textwidth]{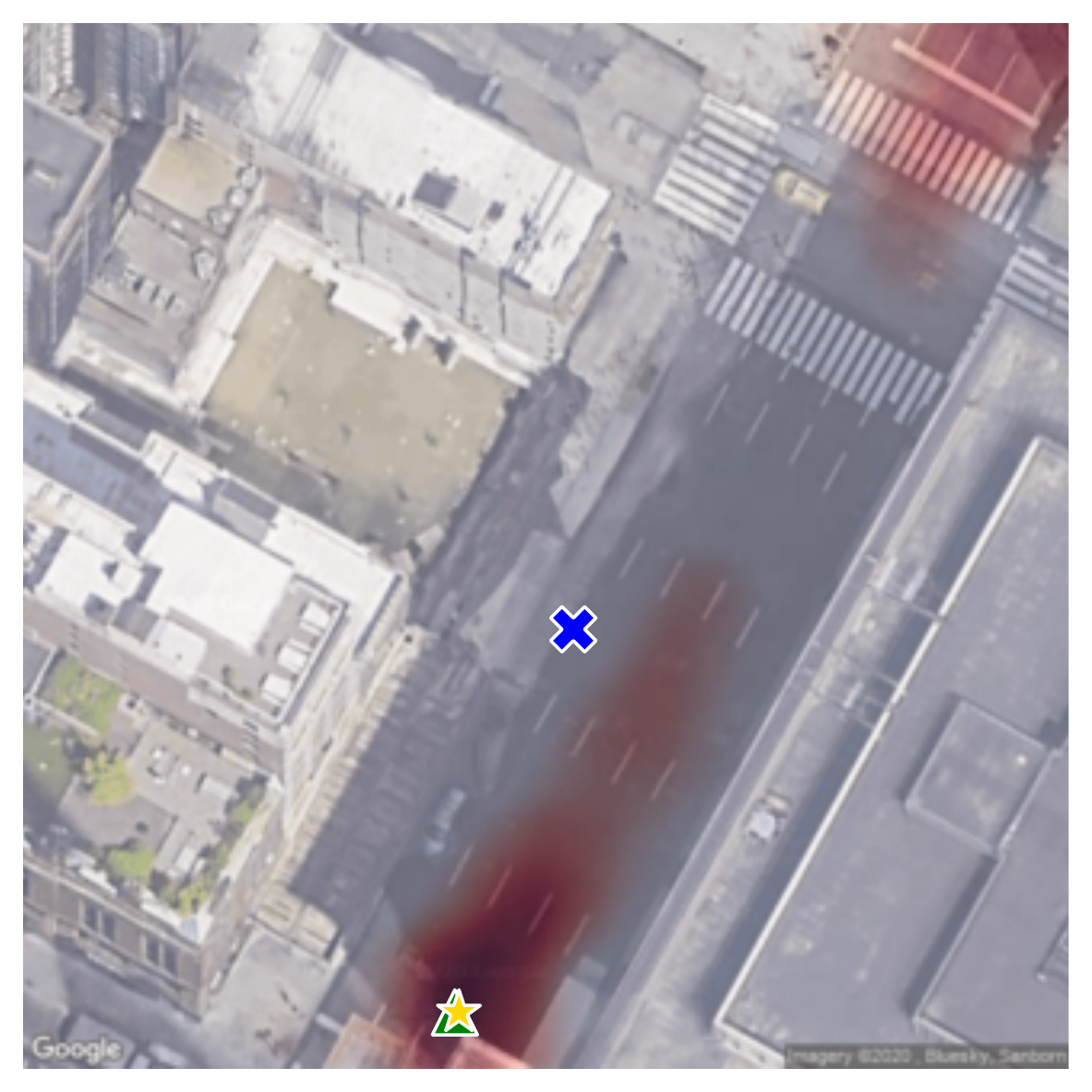}\hspace*{-0.3em}
    \includegraphics[width=0.25\textwidth]{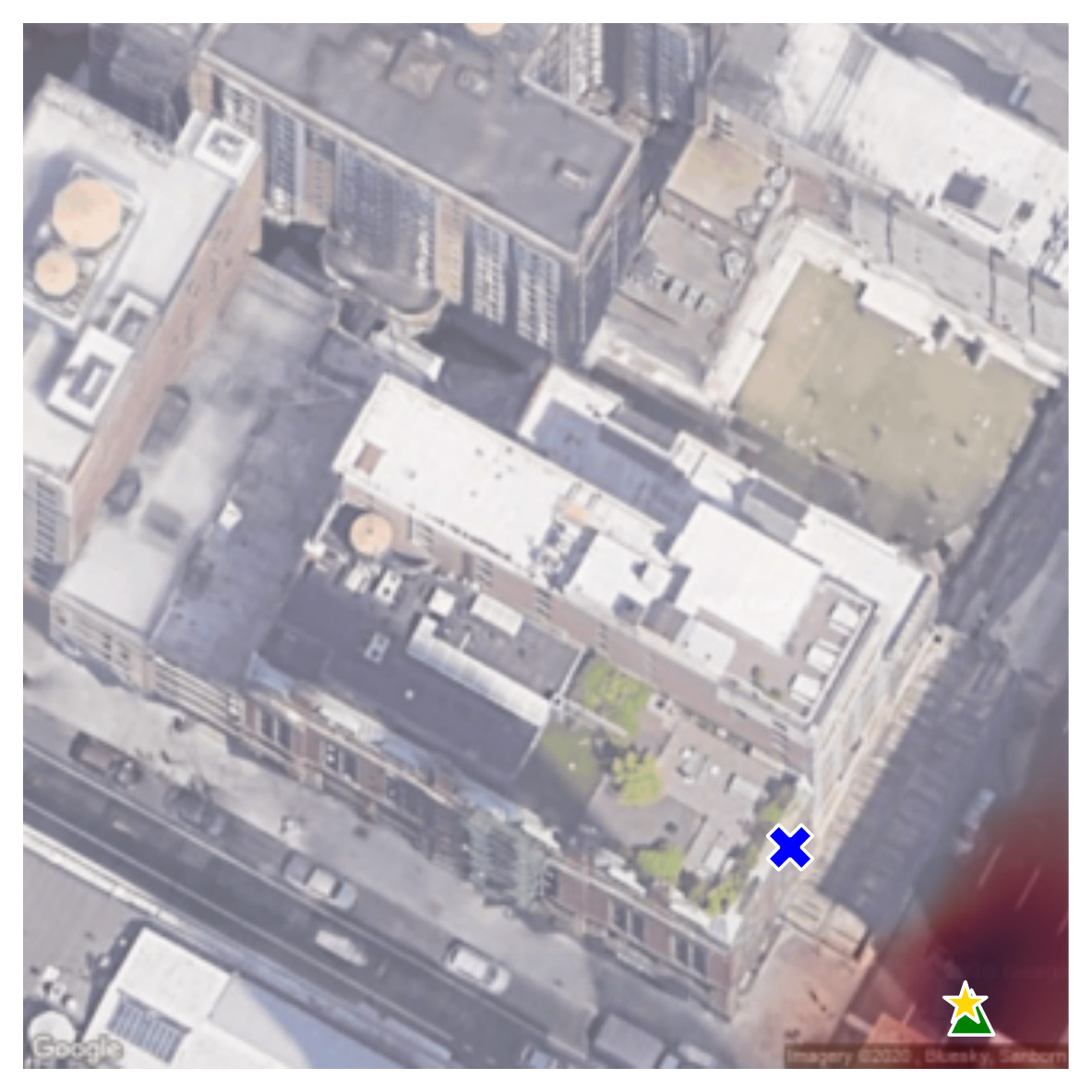}
    \caption{Matching a ground image to the positive and all semi-positive (shifted) satellite patches on the VIGOR dataset. Each example includes an input ground image, a matching score map at our model bottleneck, and an input satellite image overlayed with outputs from CVR and our method.
    The first (from left to right) example in each row is the matching between the ground image and positive. The other three show the matching between the same ground image and 3 semi-positive satellite patches.
    % Best viewed in color.
    % When matching the same ground image to positive and semi-positive satellite patches, our predicted location is more consistent than that of the baseline.
    }
    \label{fig:VIGOR_samearea_pos_semipos}

\end{figure}

Finally, we show qualitative results of localizing the ground image inside both positive and semi-positive satellite patches in Figure~\ref{fig:VIGOR_samearea_pos_semipos}.
The predicted location from our model is more consistent than that from CVR.
This confirms our benefit of localizing against semi-positive satellite patches shown in the \textbf{main paper Section 4.4 Table 2}.

\section{Additional results, generalization across time}
Next, we provide more results for \textbf{main paper Section 4.5}.

\subsection{Metric error and probability evaluation}
As a supplement to the average metric localization error and average probability at the ground truth pixel over three Oxford RobotCar test traversals in the main paper, here, we provide the separate test results on each traversal in Table~\ref{table:Oxford_localization}~and~\ref{table:Oxford_probability} for reference.

\subsection{Orientation}
Similar to Section~\ref{sec:orientation_and_metric_error_VIGOR}, we also report the localization error when jointly evaluating both localization and orientation estimation.
We keep the same setting as in \textbf{main paper Section 4.5} that the satellite patches are rotated 16 times
with $22.5^{\circ}$, starting at $0^{\circ}$ where north points in the vertically up direction.
In this case, the mean/median error increases from 1.42/1.10 (Table~\ref{table:Oxford_localization}, Test 1) to 5.46/1.98, 1.95/1.33 (Table~\ref{table:Oxford_localization}, Test 2) to 6.05/2.44, and 1.94/1.29 (Table~\ref{table:Oxford_localization}, Test 3) to 5.77/2.35 meters on three test traversals.

\begin{table}[t]
\centering
\caption{Localization error on Oxford RobotCar test traversals. Last column: Average over all traversals}
\label{table:Oxford_localization}
\begin{tabular}{ |p{2.5cm}|p{1.4cm}|p{1.4cm}|p{1.4cm}|p{1.8cm}|}
\hline
\textit{Error (meters)}& Test 1 & Test 2 & Test 3 & Average \\
\hline
CVR mean & 1.88 & 2.64 & 2.35 & 2.29 $\pm$ 0.31\\
Ours mean & \textbf{1.42} & \textbf{1.95} & \textbf{1.94} & \textbf{1.77$\pm$0.25}\\
\hline
CVR median & 1.47 & 1.99 & 1.71 & 1.72 $\pm$ 0.21\\
Ours median & \textbf{1.10} & \textbf{1.33} & \textbf{1.29} & \textbf{1.24$\pm$0.10}\\
\hline
\end{tabular}
\end{table}

\begin{table}[t]
\centering
\caption{Probabilities at the ground truth pixel on Oxford RobotCar. Last column: Average over all traversals. For reference, the probability in a $512\times512$ uniform map is 3.81e-6. Best in bold}
\label{table:Oxford_probability}
\begin{tabular}{ |p{2.5cm}|p{1.8cm}|p{1.8cm}|p{1.8cm}|p{1.8cm}|}
\hline
\textit{Prob.~at GT}& Test 1 & Test 2 & Test 3 & Average \\
\hline
CVR mean & $1.78\times10^{-4}$ & $1.59\times10^{-4}$ & $1.65\times10^{-4}$ & $1.67\times10^{-4}$\\
Ours mean & $\mathbf{1.76\times10^{-3}}$ & $\mathbf{1.40\times10^{-3}}$ & $\mathbf{1.46\times10^{-3}}$ & $\mathbf{1.54\times10^{-3}}$ \\
\hline
CVR median & $1.96\times10^{-4}$ & $1.81\times10^{-4}$ & $1.89\times10^{-4}$ & $1.89\times10^{-4}$\\
Ours median & $\mathbf{1.69\times10^{-3}}$ & $\mathbf{1.18\times10^{-3}}$ & $\mathbf{1.28\times10^{-3}}$ & $\mathbf{1.38\times10^{-3}}$ \\
\hline
\end{tabular}
\end{table}

\begin{figure}[t]
    \centering
    \includegraphics[width=0.7\textwidth]{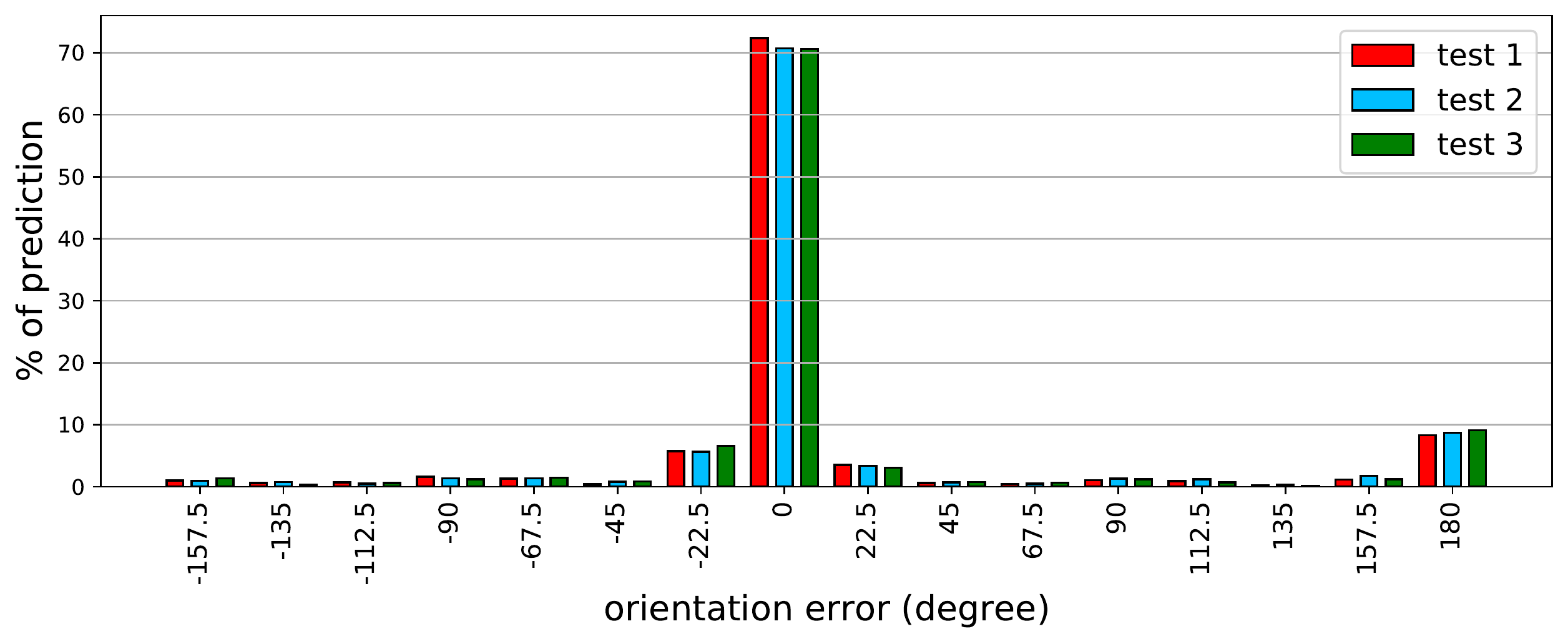}
    \caption{An overview of orientation classification results on three test traversals from Oxford RobotCar.
    }
    \label{fig:Oxford_orientation_histogram}
\end{figure}

\begin{figure}[t]
    \centering
    \includegraphics[width=0.24\textwidth]{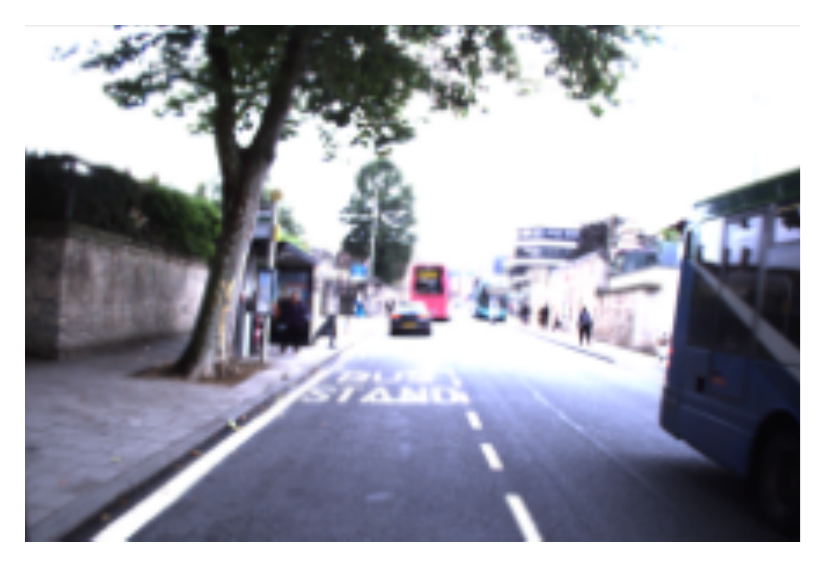}\\
    \includegraphics[width=0.24\textwidth]{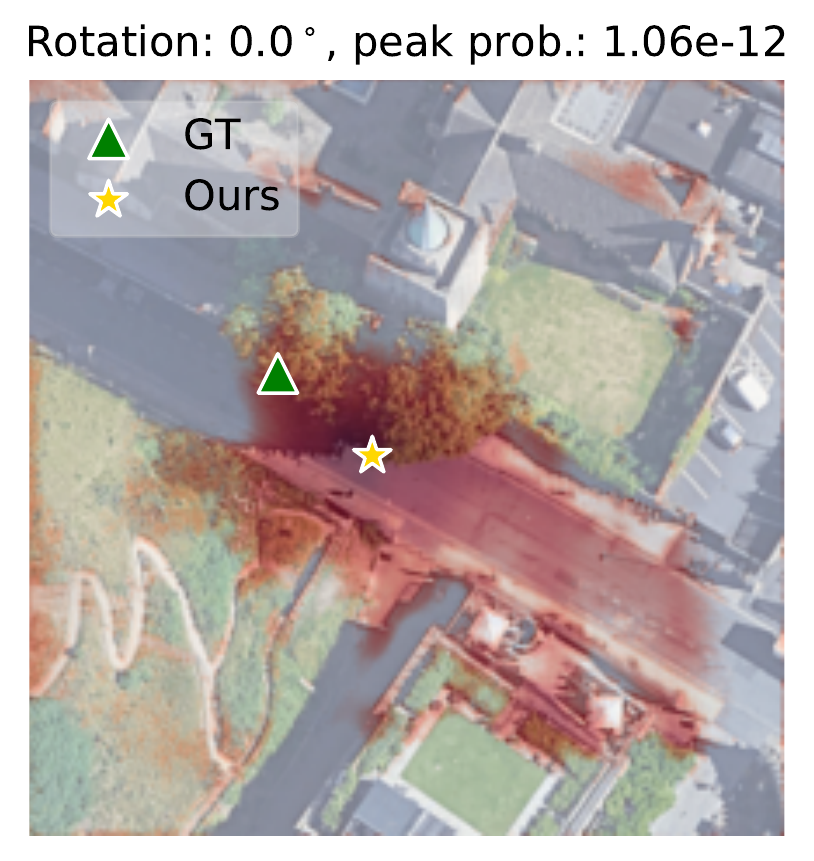}
    \hspace*{-0.5em}
    \includegraphics[width=0.24\textwidth]{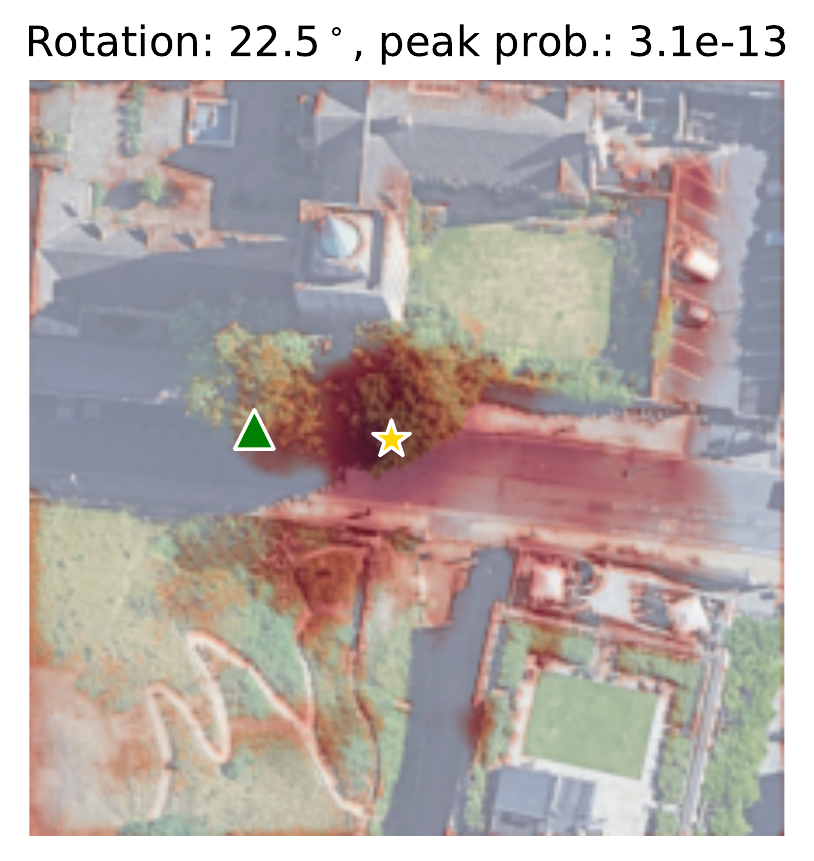}
    \hspace*{-0.5em}
    \includegraphics[width=0.24\textwidth]{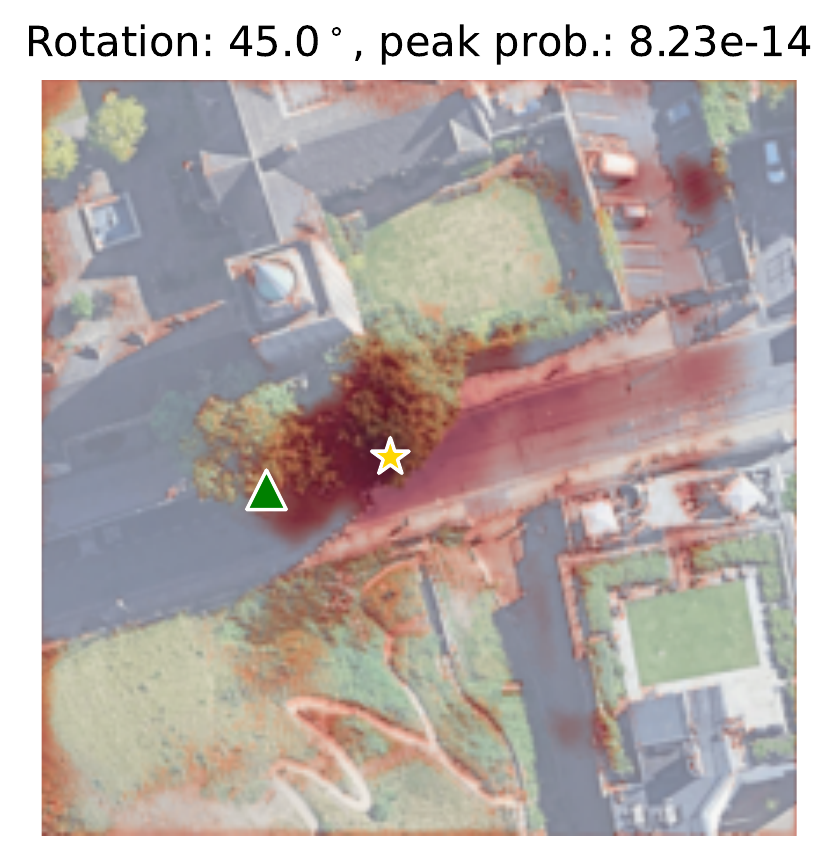}
    \hspace*{-0.5em}
    \includegraphics[width=0.24\textwidth]{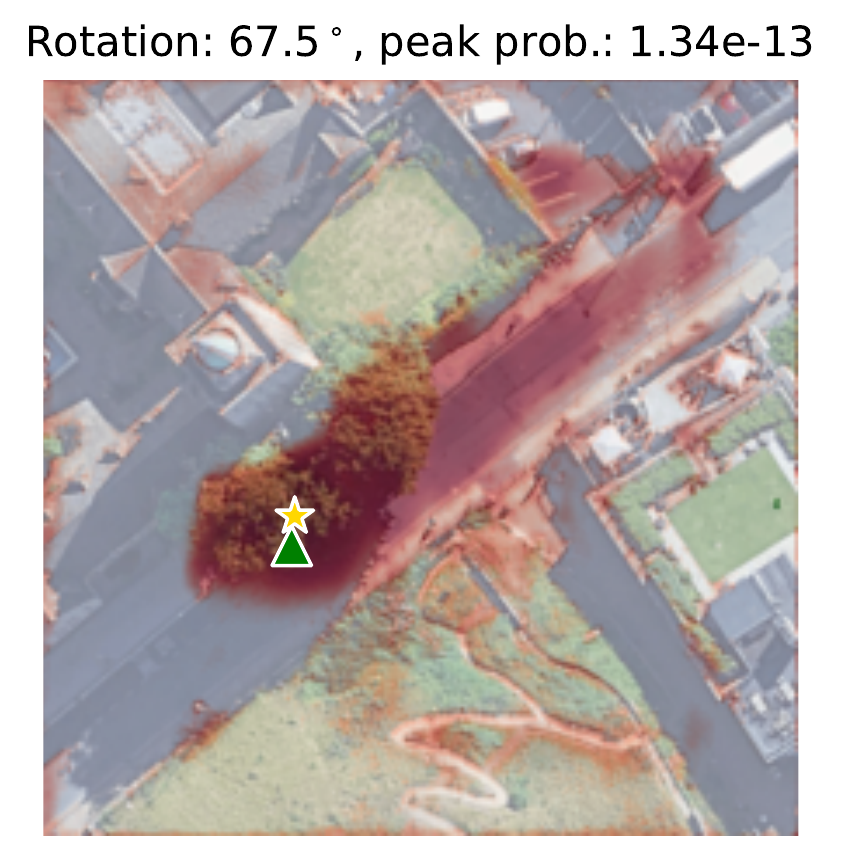}\\
    \hspace*{-0.5em}
    \includegraphics[width=0.24\textwidth]{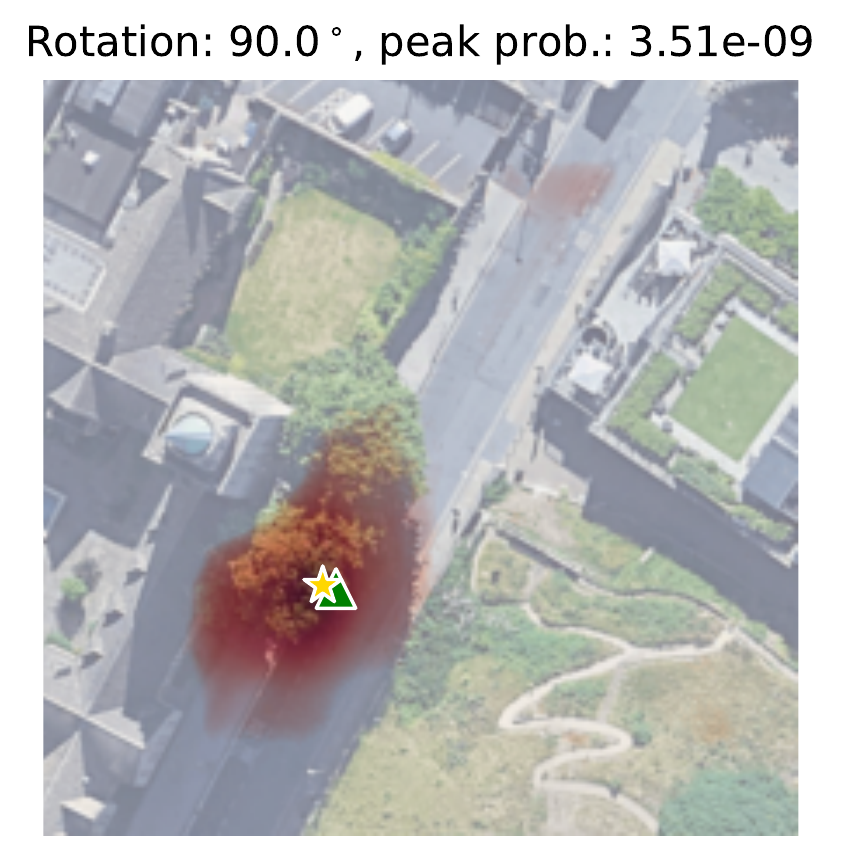}
    \hspace*{-0.5em}
    \includegraphics[width=0.24\textwidth]{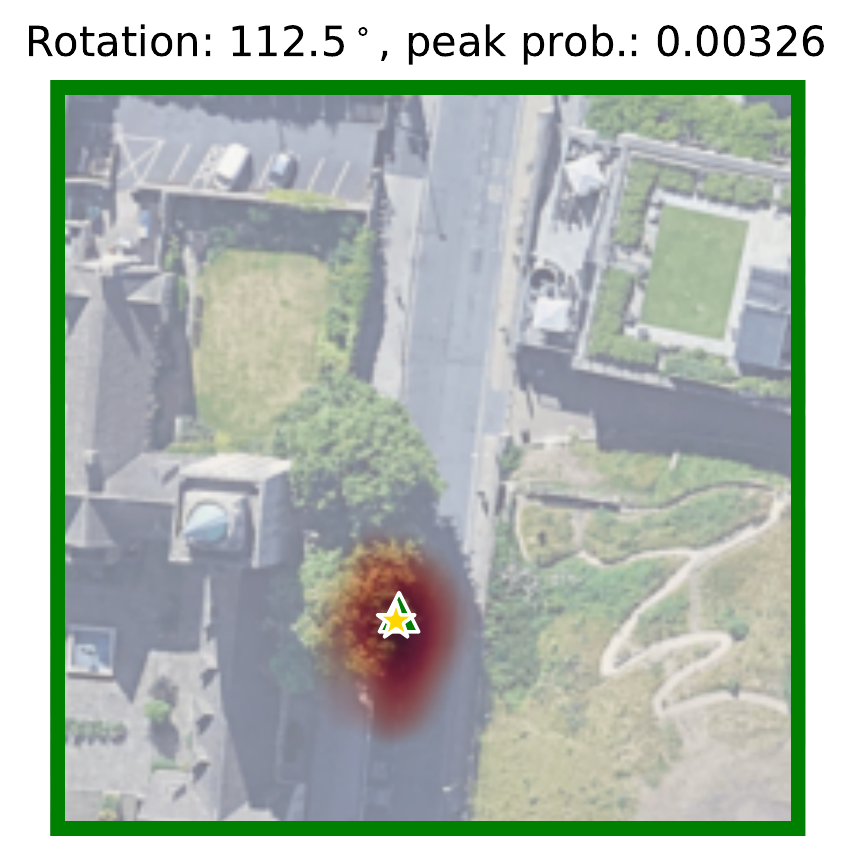}
    \hspace*{-0.5em}
    \includegraphics[width=0.24\textwidth]{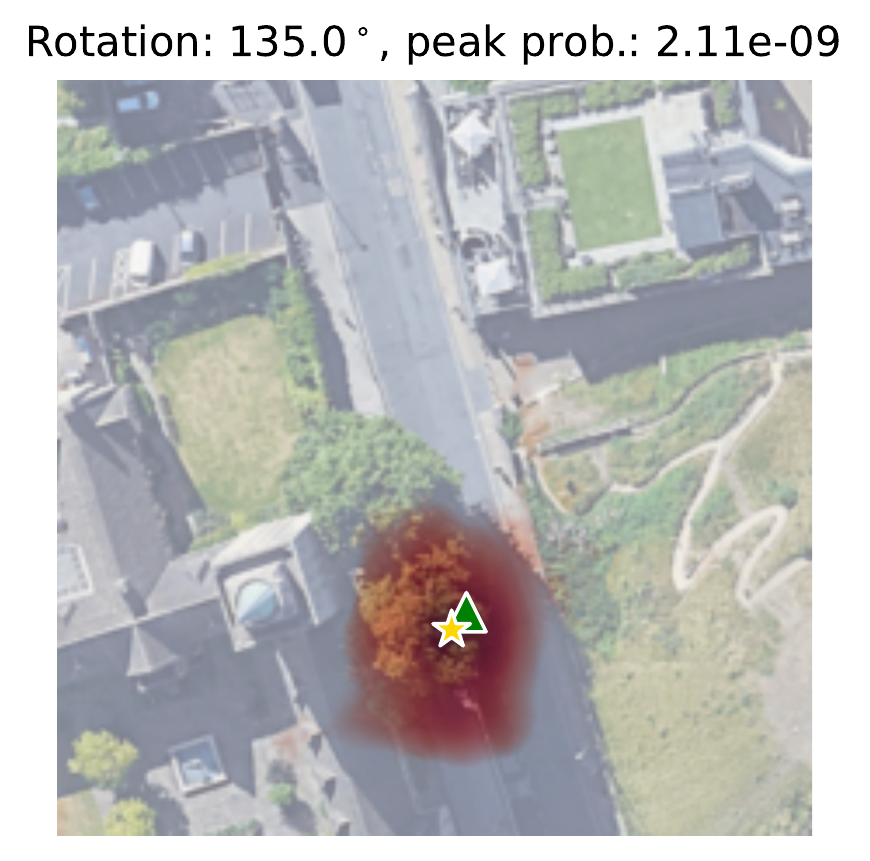}
    \hspace*{-0.5em}
    \includegraphics[width=0.24\textwidth]{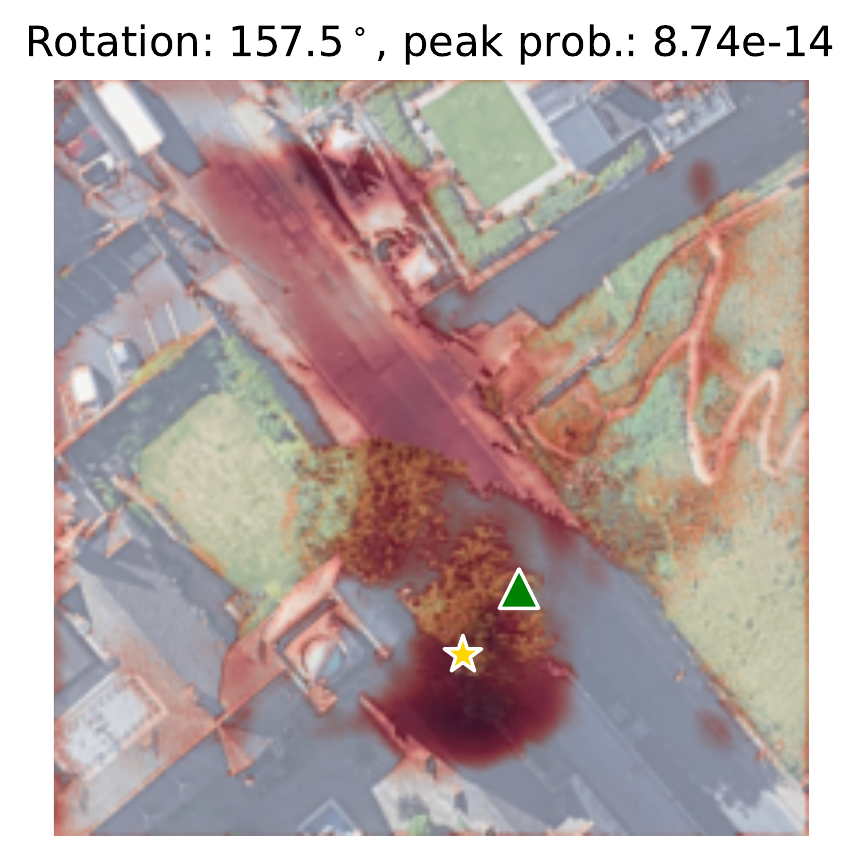}\\
    \hspace*{-0.5em}
    \includegraphics[width=0.24\textwidth]{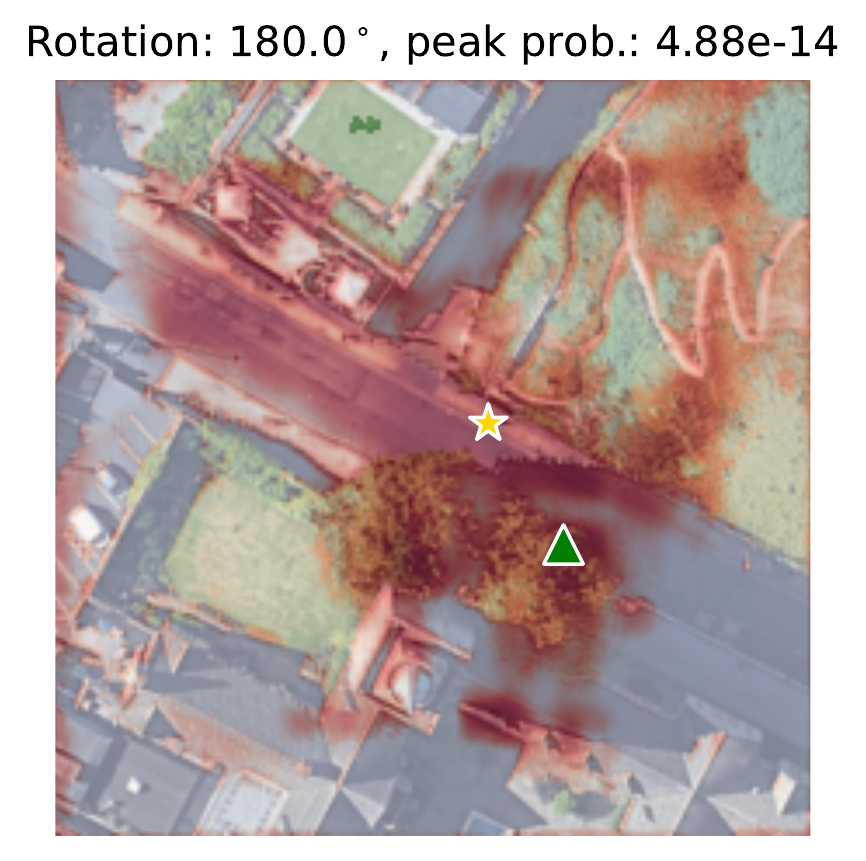}
    \hspace*{-0.5em}
    \includegraphics[width=0.24\textwidth]{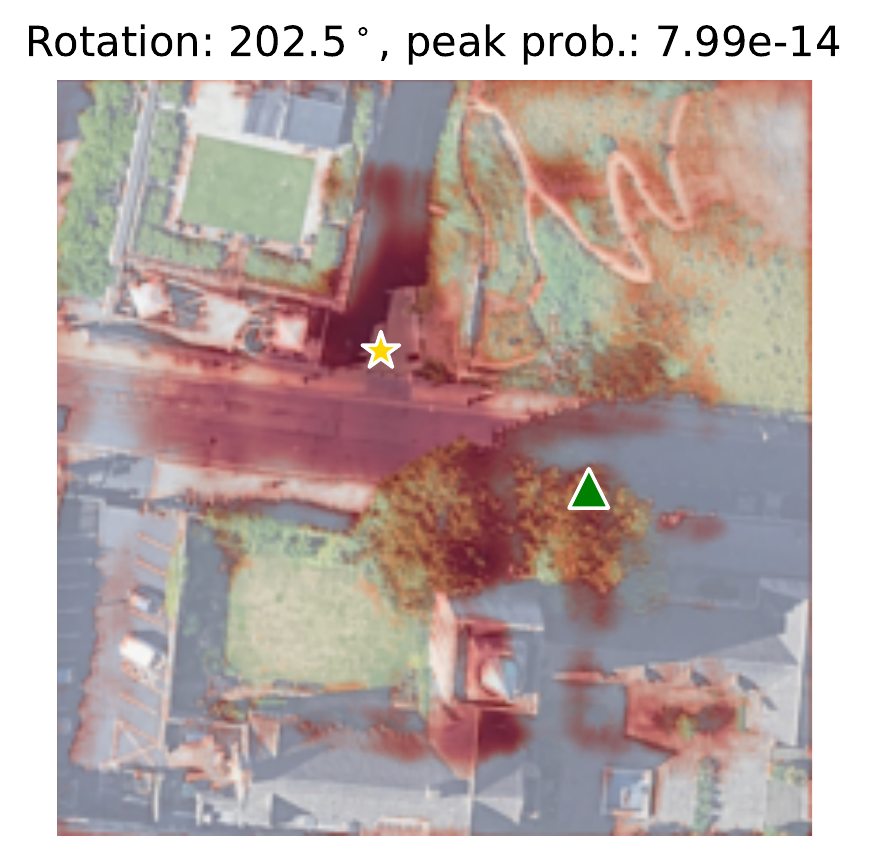}
    \hspace*{-0.5em}
    \includegraphics[width=0.24\textwidth]{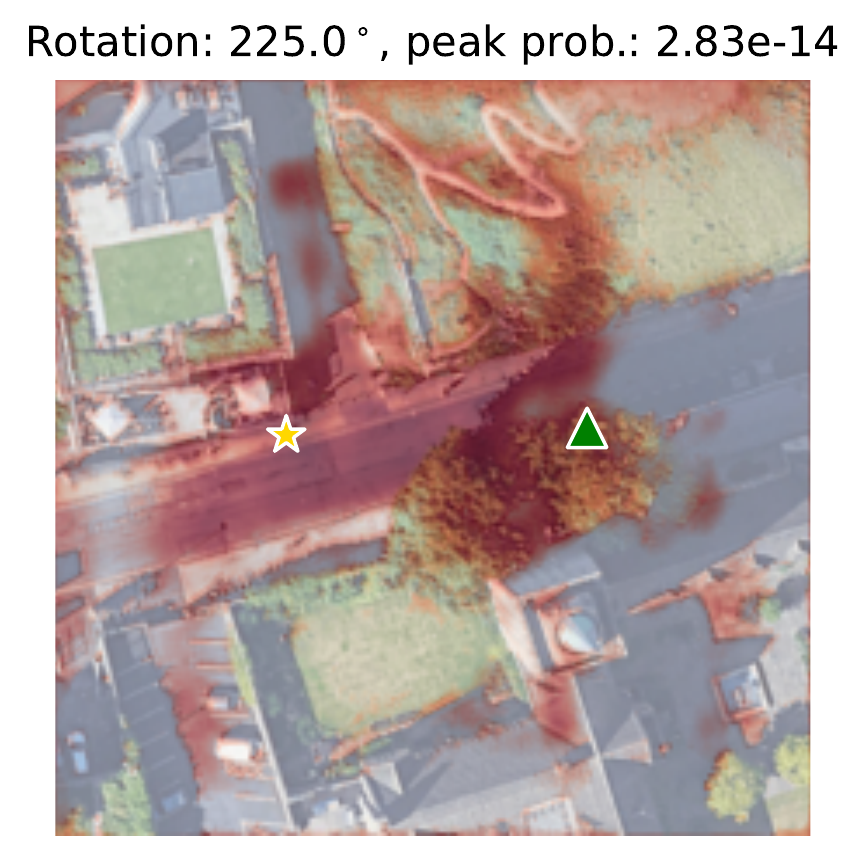}
    \hspace*{-0.5em}
    \includegraphics[width=0.24\textwidth]{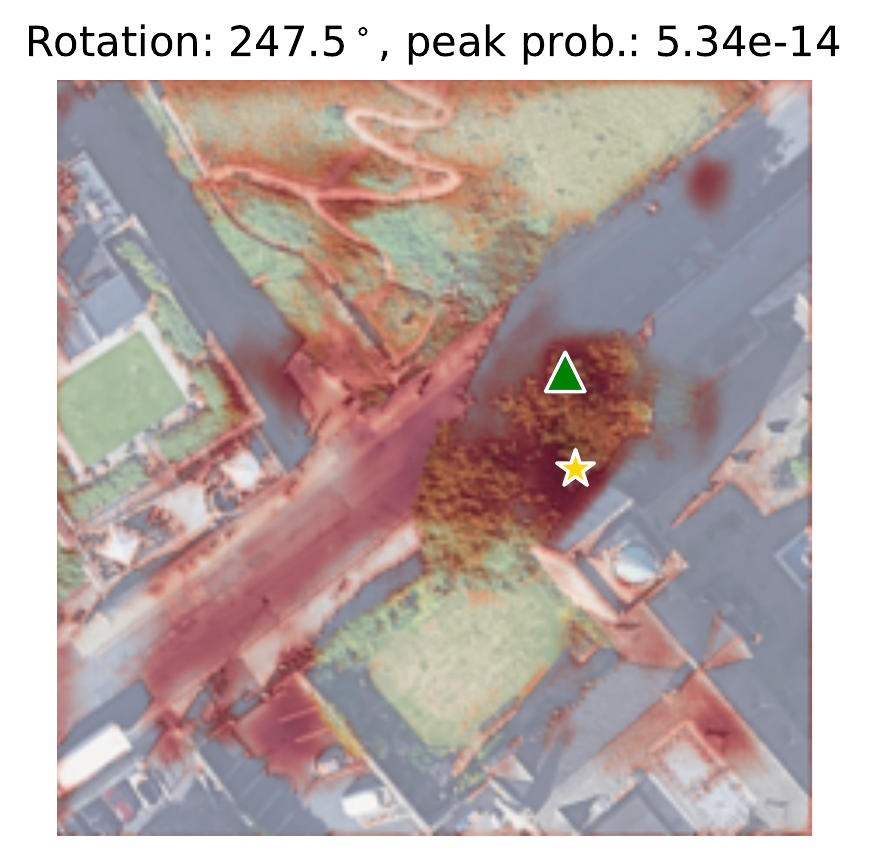}\\
    \hspace*{-0.5em}
    \includegraphics[width=0.24\textwidth]{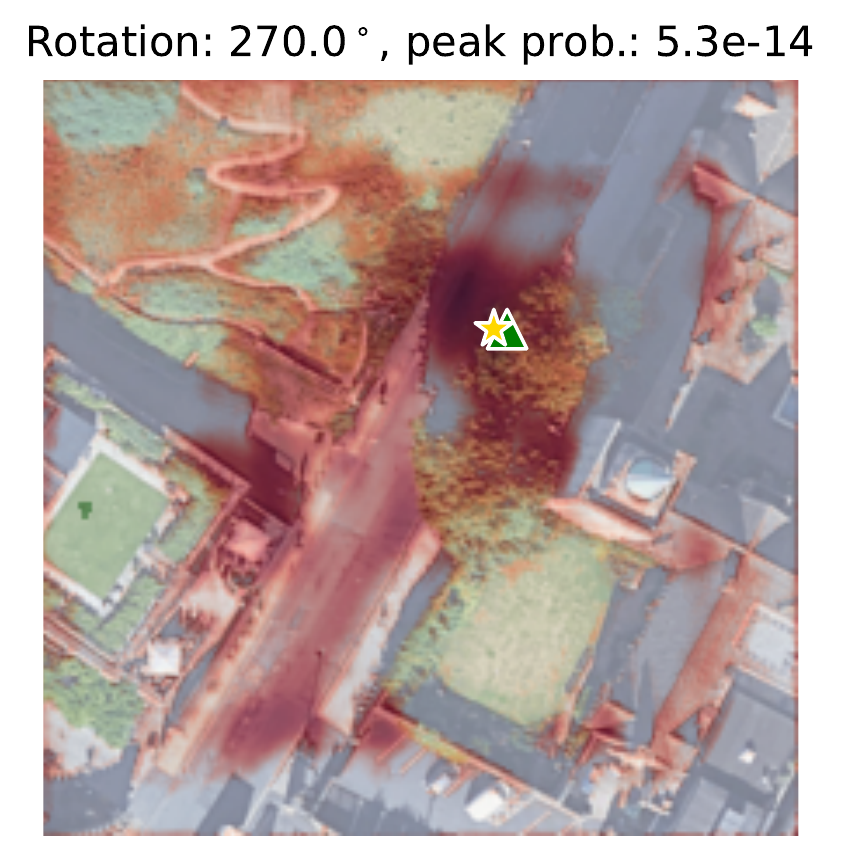}
    \includegraphics[width=0.24\textwidth]{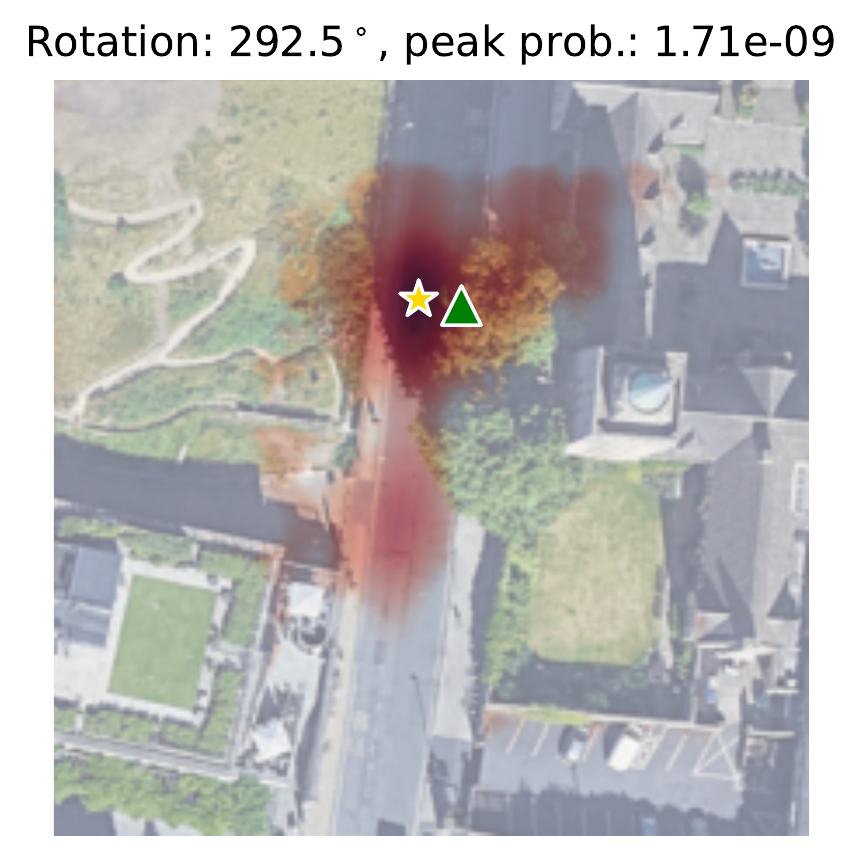}
    \hspace*{-0.5em}
    \includegraphics[width=0.24\textwidth]{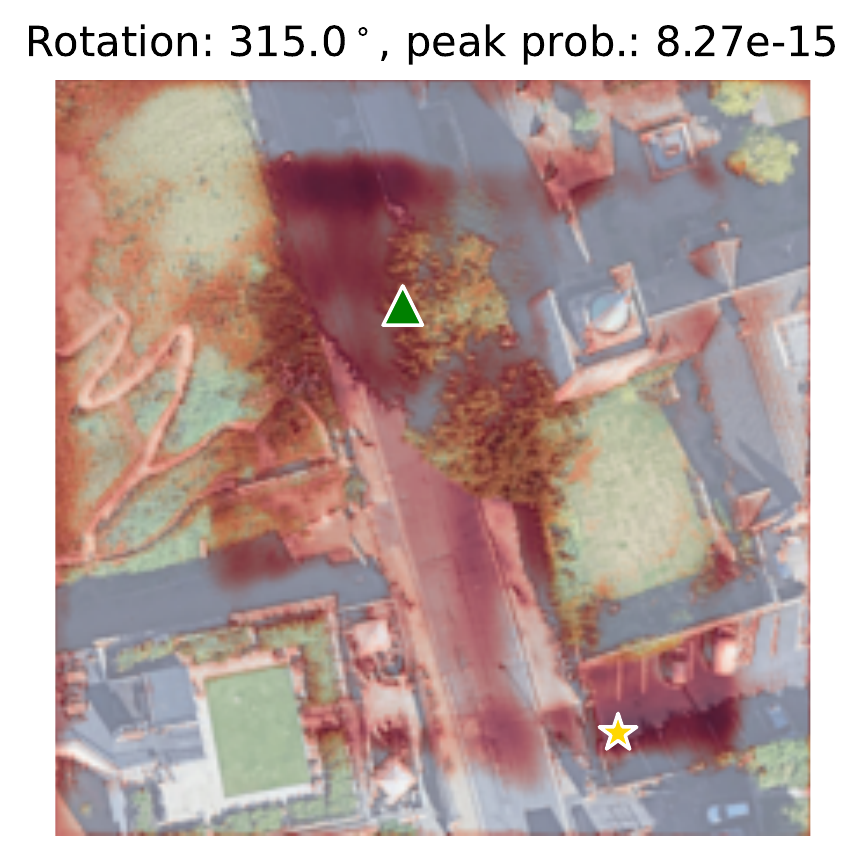}\hspace*{-0.5em}
    \includegraphics[width=0.24\textwidth]{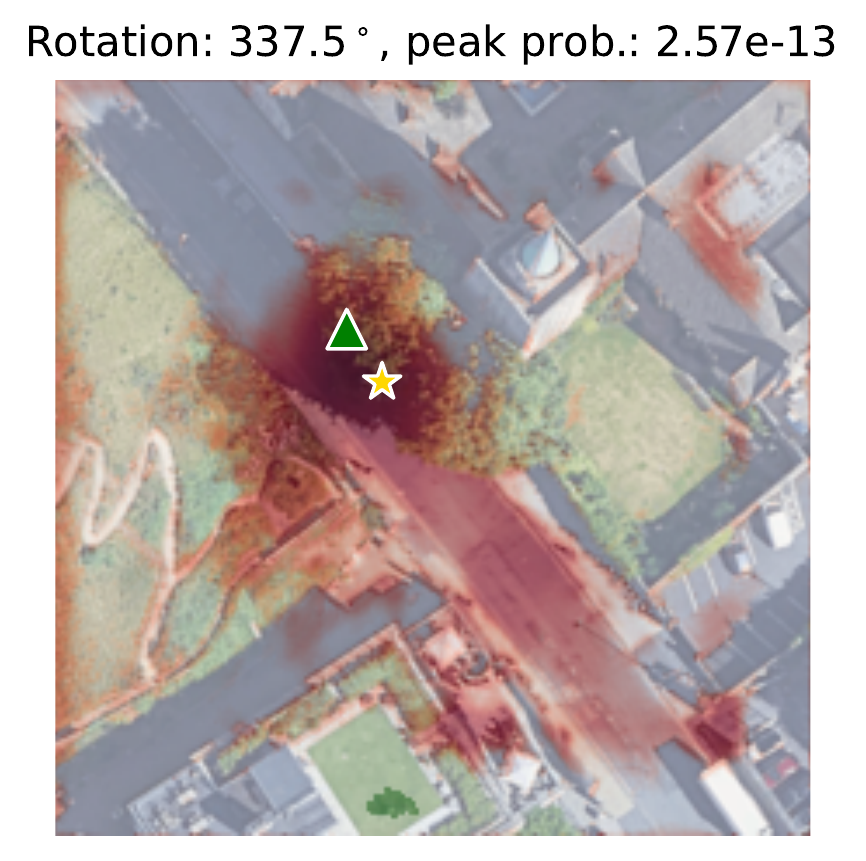}
    \caption{Oxford RobotCar, the same ground image matched to 16 different rotated satellite patches. 
    The ground truth orientation is $118^{\circ}$.
    The satellite patch with $0^{\circ}$ rotation is the north-aligned satellite patch.
    The patch with a green box ($112.5^{\circ}$) is the one prediction with the highest peak probability.
    The probabilities are generated by the softmax operation over a batch of inputs.
    For visualization, we re-scale the probability in each heat map.
    }
    \label{fig:Oxford_orientation}
\end{figure}

Akin to the evaluation on VIGOR, \textbf{main paper Figure 6 right}, we also provide the histogram of orientation classification results on Oxford RobotCar. 
As shown in Figure~\ref{fig:Oxford_orientation_histogram}, the most frequent wrong class is $180^{\circ}$.
We show qualitative results of the orientation classification experiment in Figure~\ref{fig:Oxford_orientation}.
The model is trained with orientation-aligned satellite patches, which means if the ground image's sight is along the road, the corresponding road in the satellite patch is along the vertical direction.
In inference, our model exploits this information and assigns low probability when the road is not vertical.
Importantly, our model is not a vertical road detector, since it also tries to match other objects, e.g. trees, across views, plus it can also differentiate the satellite patch rotated $180^{\circ}$ from the correct orientation class in most cases.
As shown in Figure~\ref{fig:Oxford_orientation}, when the ground image indicates the vehicle is under the tree, our model tries to find the vegetation in the satellite view no matter the rotation angle.

\subsection{Localization qualitative results}
\begin{figure}[t]
    \centering
    % \hspace*{-0.5em}
    % \includegraphics[width=0.17\textwidth]{figures_sup/oxford/T1/Oxford_ground_538.pdf}\hspace*{-0.4em}
    % \includegraphics[width=0.085\textwidth]{figures_sup/oxford/T1/Oxford_matching_538.pdf}\hspace*{-0.4em}
    % \includegraphics[width=0.17\textwidth]{figures_sup/oxford/T2/Oxford_ground_598.pdf}\hspace*{-0.4em}
    %  \includegraphics[width=0.085\textwidth]{figures_sup/oxford/T2/Oxford_matching_598.pdf}\hspace*{-0.4em}
    % \includegraphics[width=0.17\textwidth]{figures_sup/oxford/T3/Oxford_ground_658.pdf}\hspace*{-0.4em}
    %  \includegraphics[width=0.085\textwidth]{figures_sup/oxford/T3/Oxford_matching_658.pdf}\hspace*{-0.4em}
    % \includegraphics[width=0.17\textwidth]{figures_sup/oxford/T2/Oxford_ground_121.pdf}\hspace*{-0.4em}
    %  \includegraphics[width=0.085\textwidth]{figures_sup/oxford/T2/Oxford_matching_121.pdf}
    % \\%\vspace*{-0.3em}
    \includegraphics[width=0.25\textwidth]{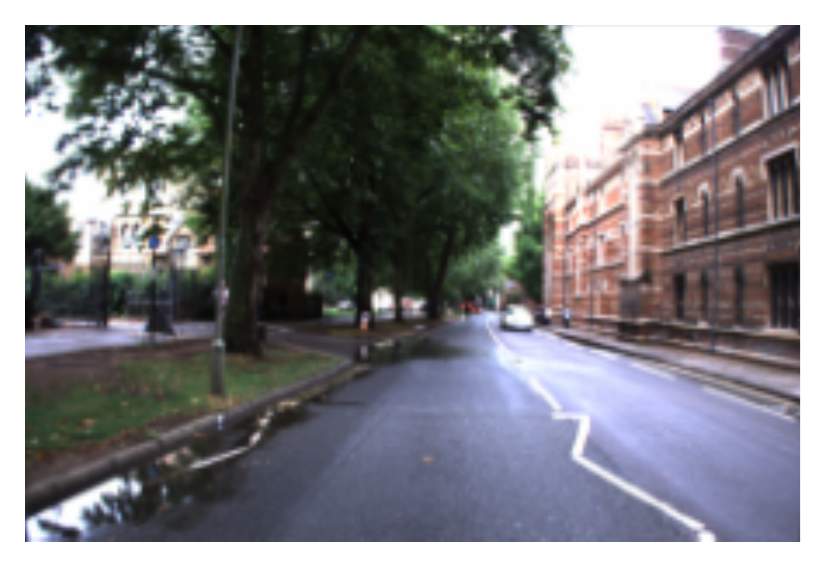}\hspace*{-0.4em}
    \includegraphics[width=0.25\textwidth]{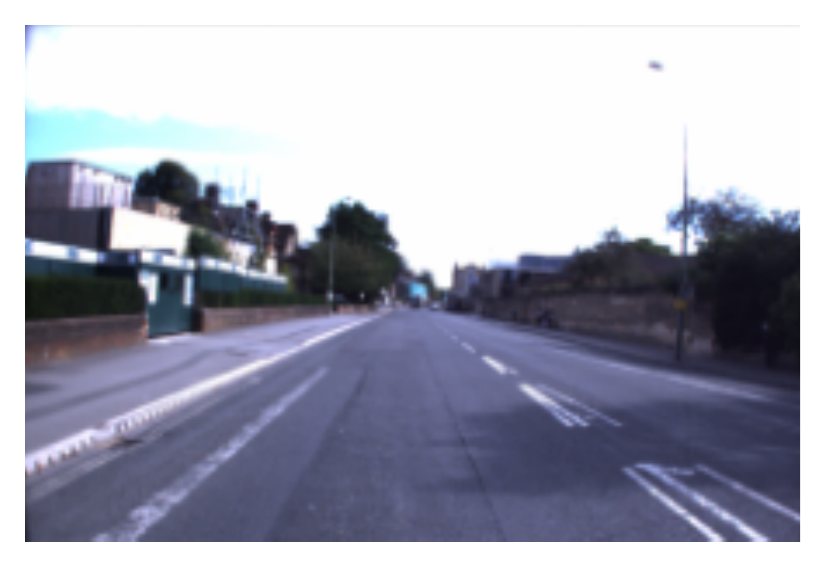}\hspace*{-0.4em}
    \includegraphics[width=0.25\textwidth]{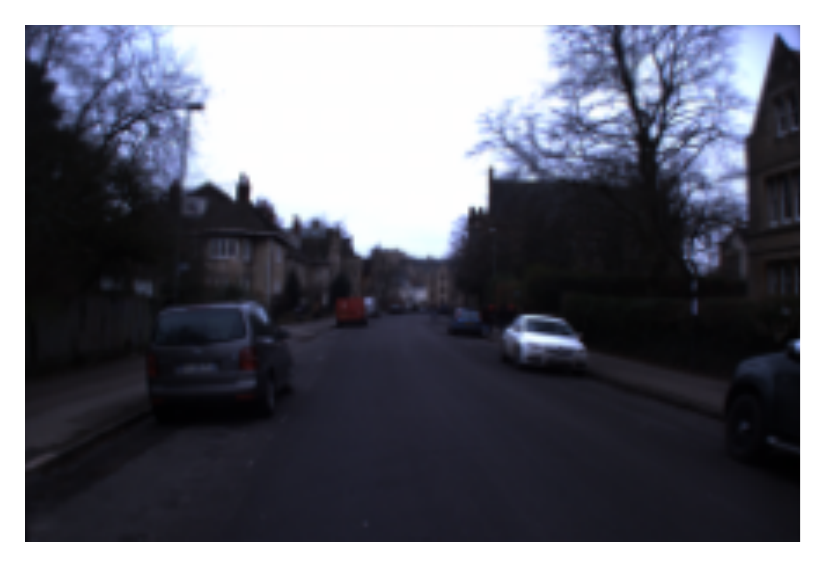}\hspace*{-0.4em}
    \includegraphics[width=0.25\textwidth]{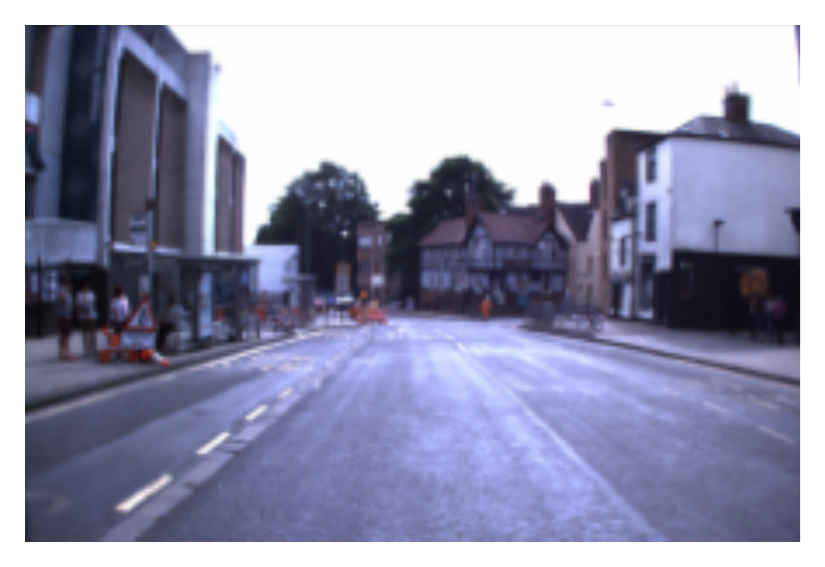}\hspace*{-0.4em}
    \\
    \includegraphics[width=0.25\textwidth]{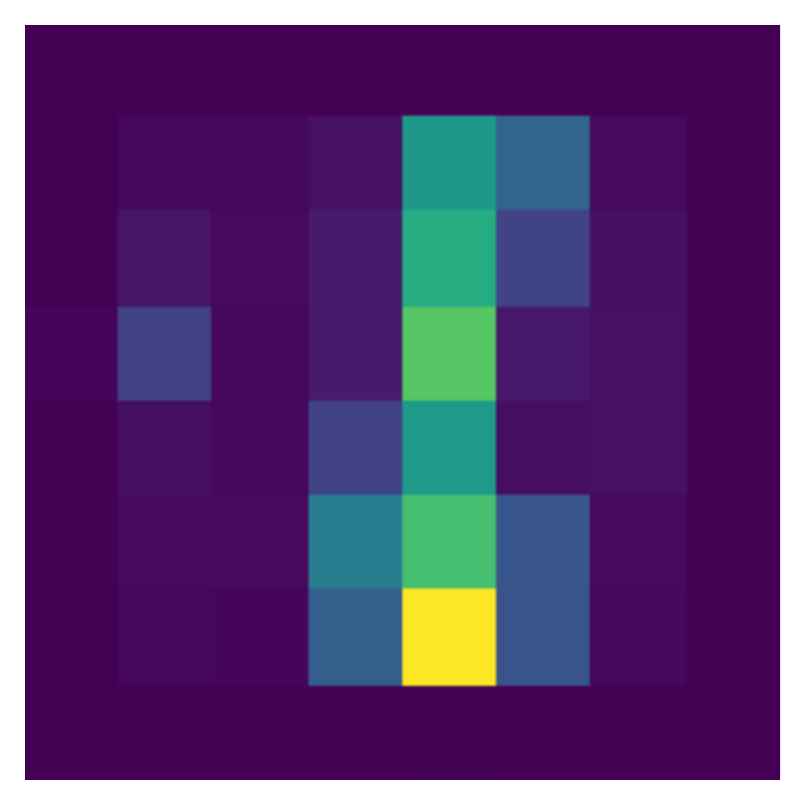}\hspace*{-0.4em}
     \includegraphics[width=0.25\textwidth]{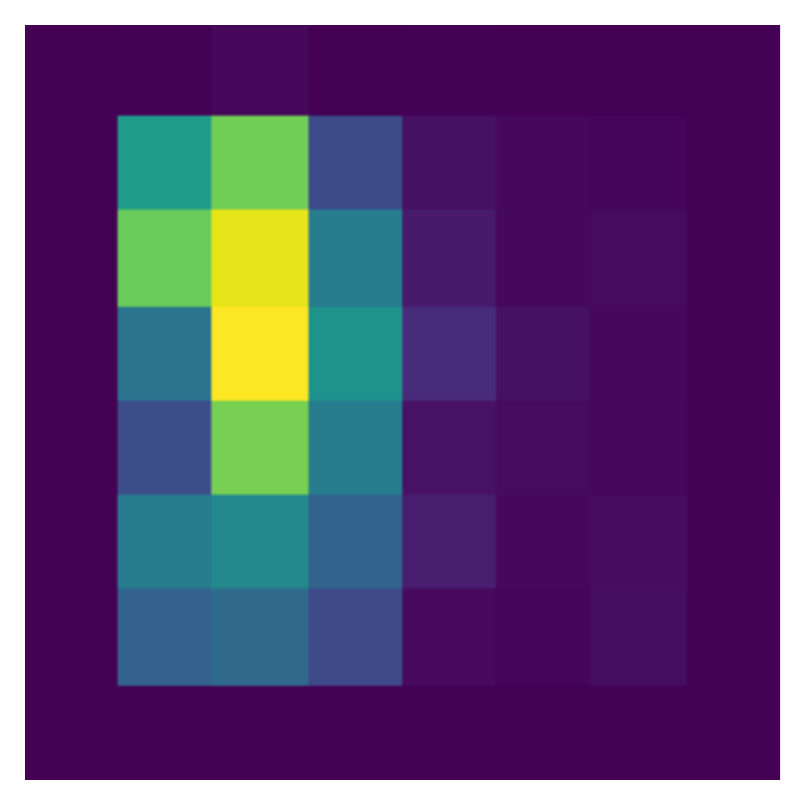}\hspace*{-0.4em}
     \includegraphics[width=0.25\textwidth]{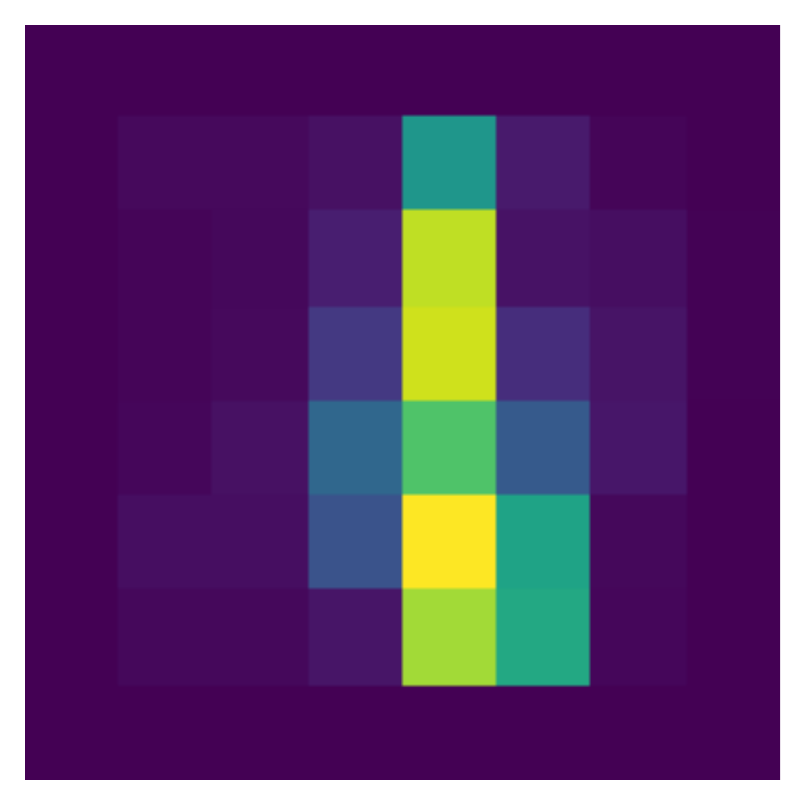}\hspace*{-0.4em}
     \includegraphics[width=0.25\textwidth]{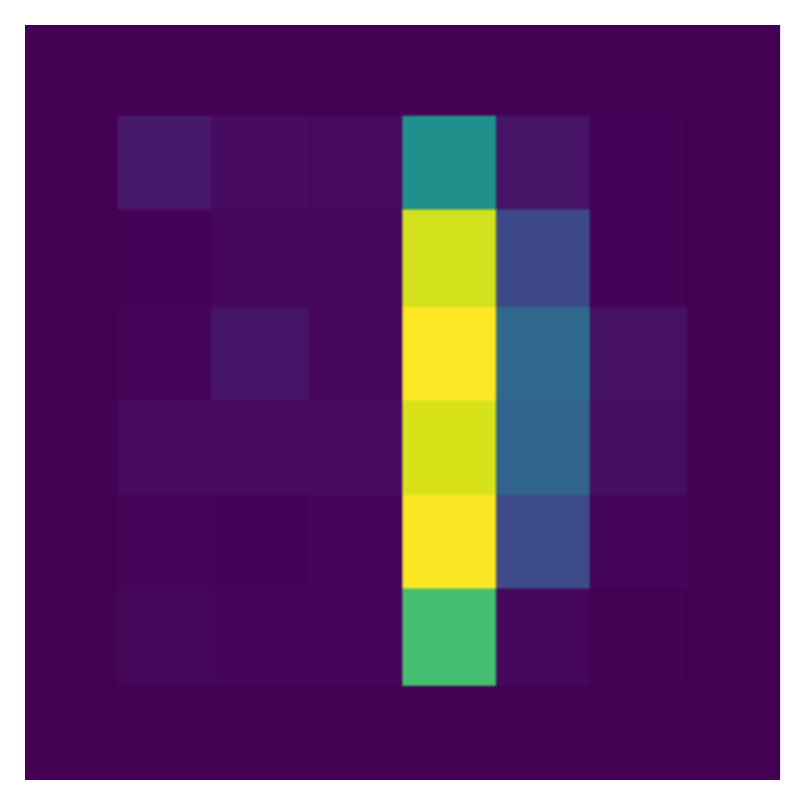}
    \\
    \includegraphics[width=0.25\textwidth]{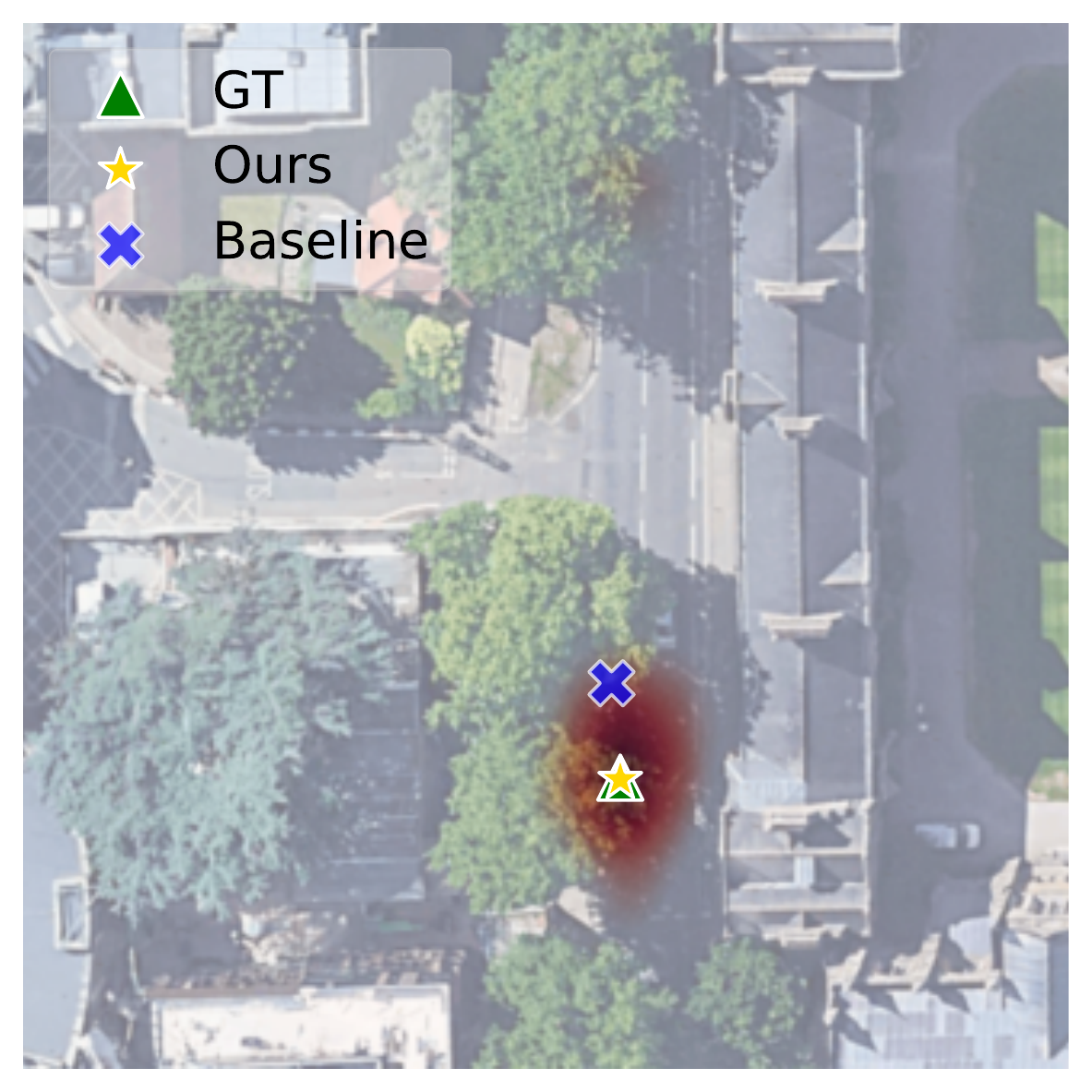}\hspace*{-0.3em}
    \includegraphics[width=0.25\textwidth]{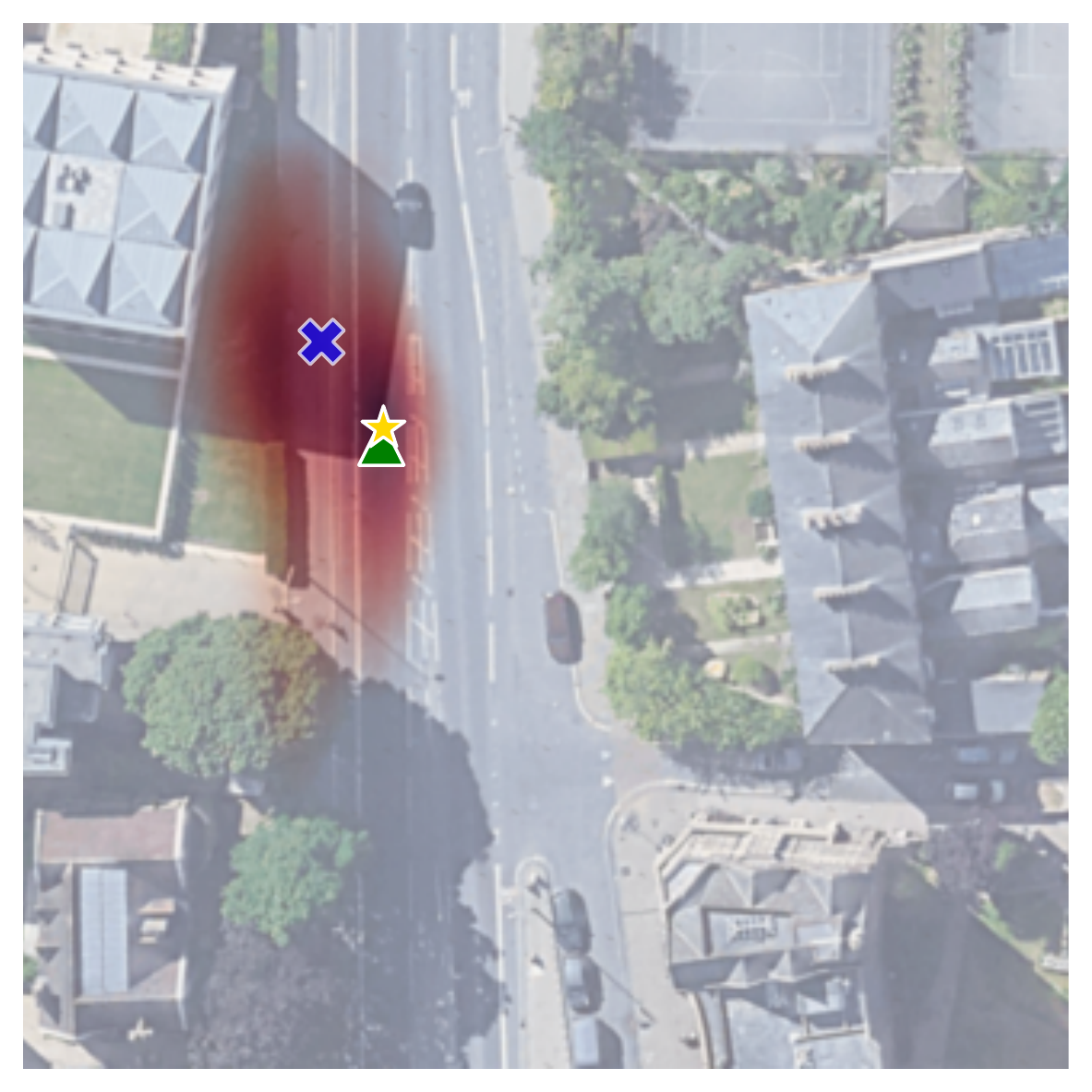}\hspace*{-0.3em}
    \includegraphics[width=0.25\textwidth]{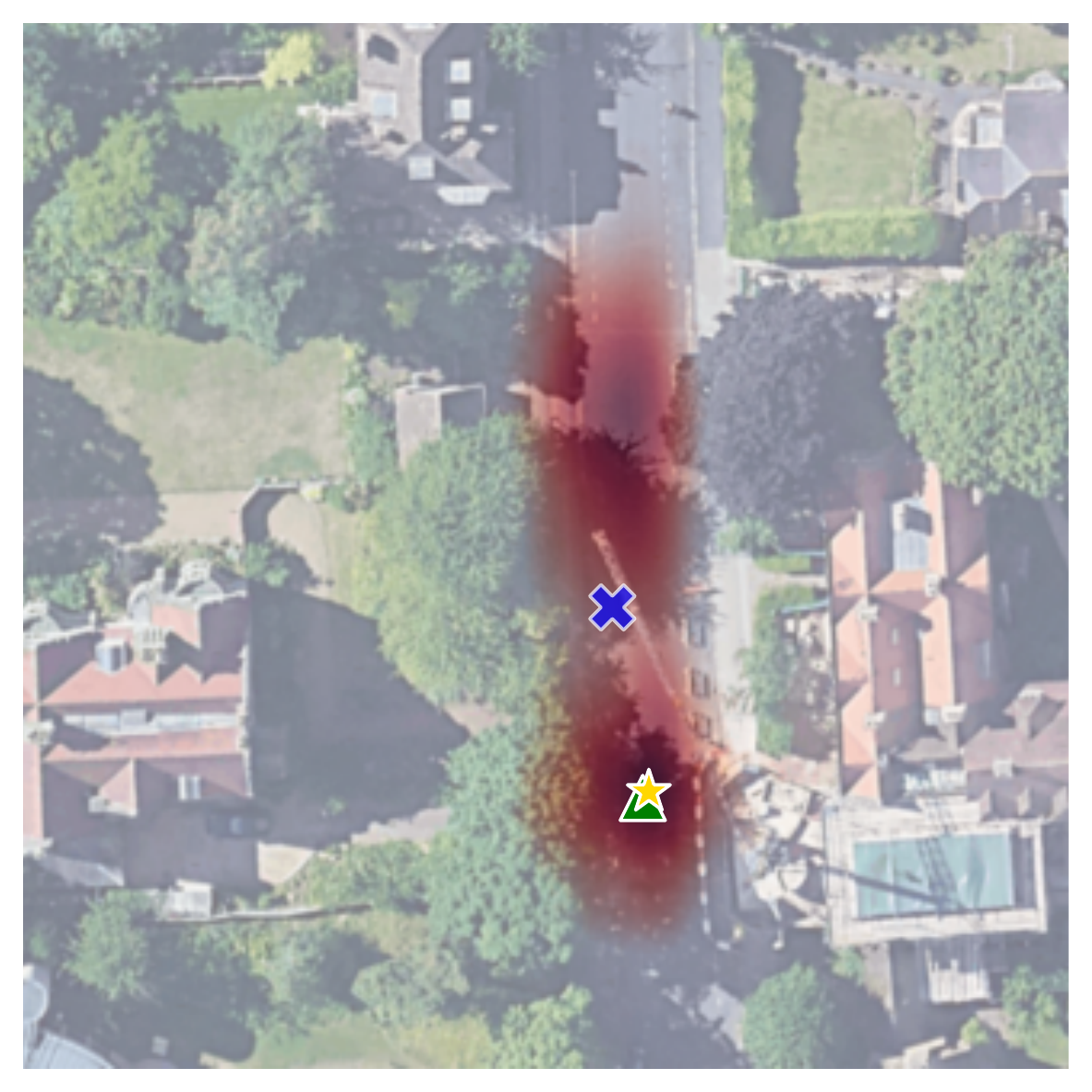}\hspace*{-0.3em}
    \includegraphics[width=0.25\textwidth]{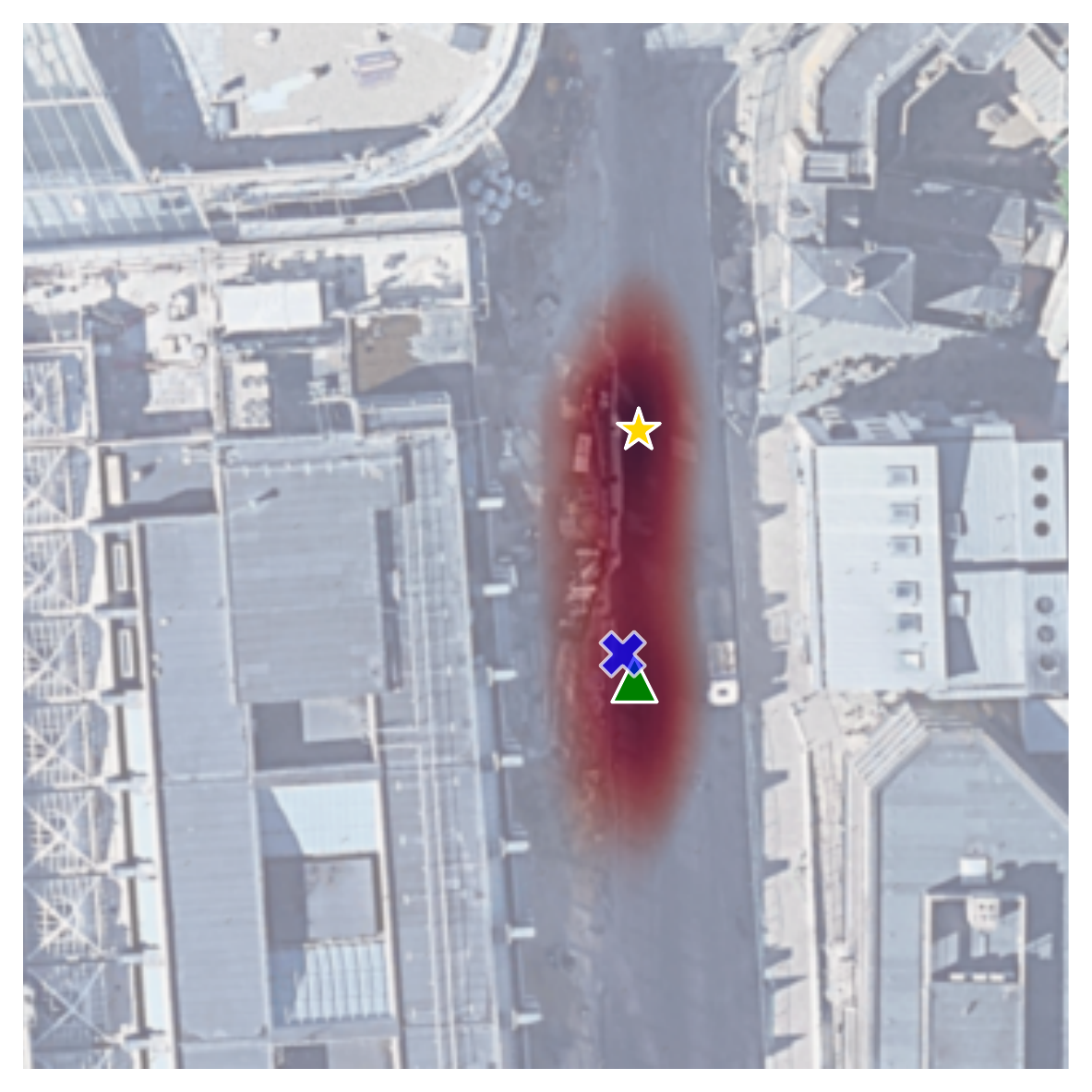}
    \\
    \caption{Qualitative results on Oxford RobotCar dataset. 
    Each column has an input ground image, a matching score map at our model bottleneck, and an input satellite image overlayed with outputs from CVR and our method.
    Column 1/2/3 (from left to right): good predictions in test traversal 1/2/3.
    Column 4: a failure case example.
    % Best viewed in color.
    }
    \label{fig:Oxford}
\end{figure}

\begin{figure}[t]
    \centering
    \includegraphics[width=0.25\textwidth]{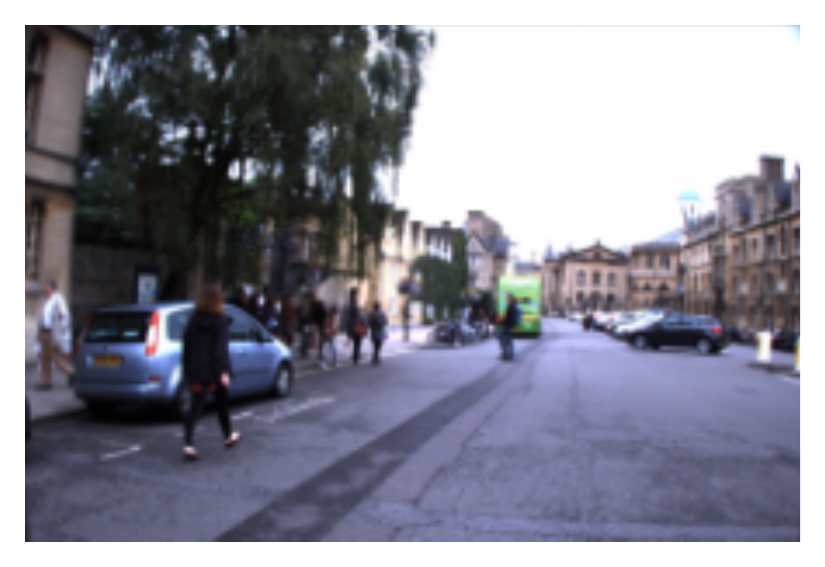}\hspace*{-0.4em}
    \includegraphics[width=0.25\textwidth]{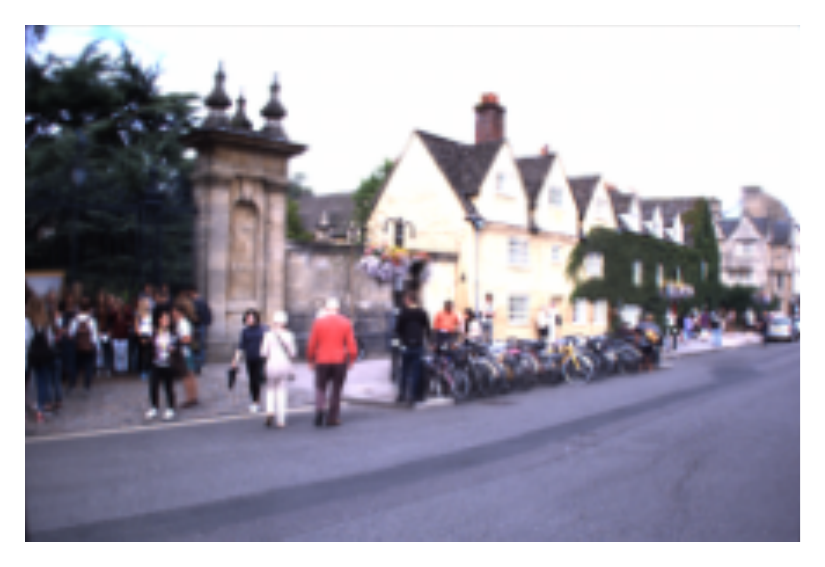}\hspace*{-0.4em}
    \includegraphics[width=0.25\textwidth]{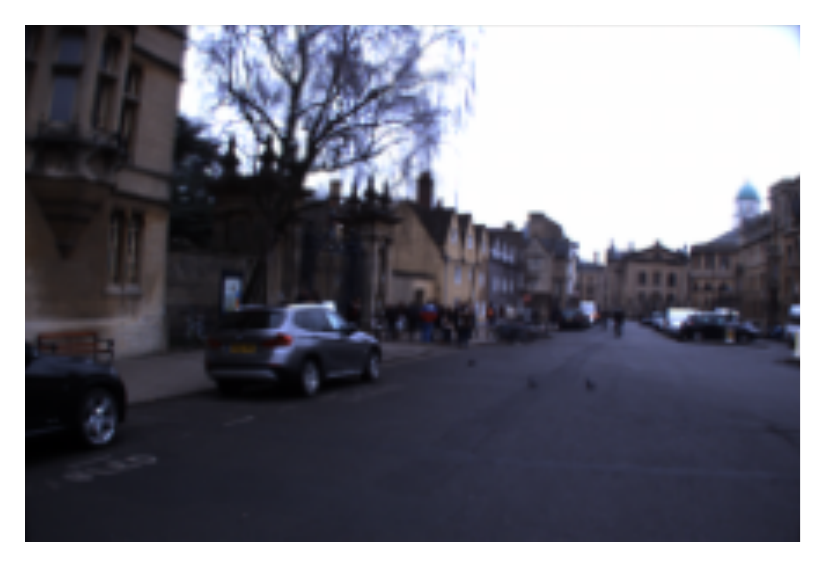}\hspace*{-0.4em}
    \\
    \includegraphics[width=0.25\textwidth]{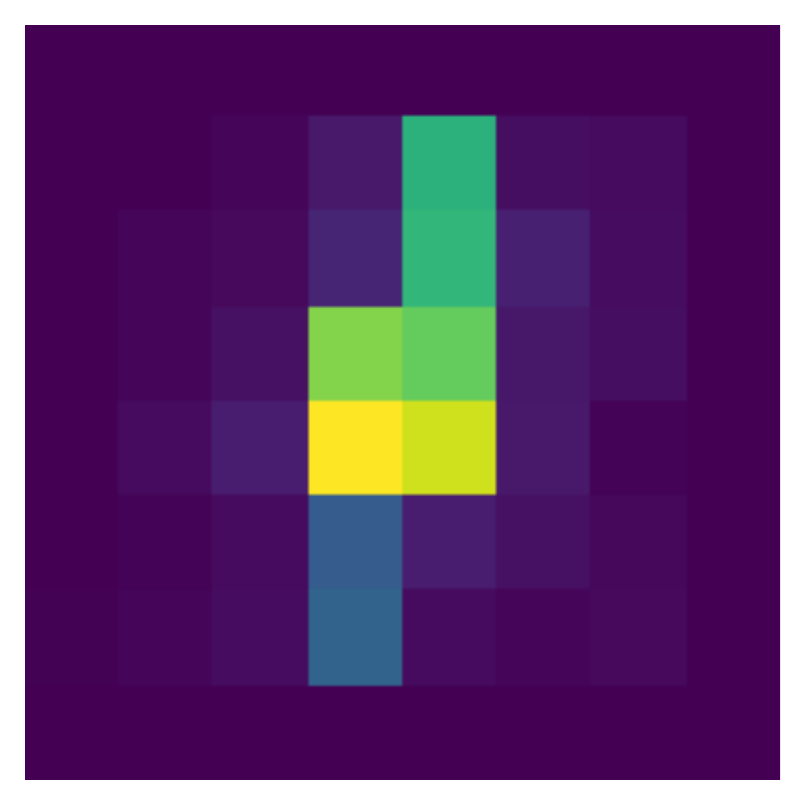}\hspace*{-0.4em}
     \includegraphics[width=0.25\textwidth]{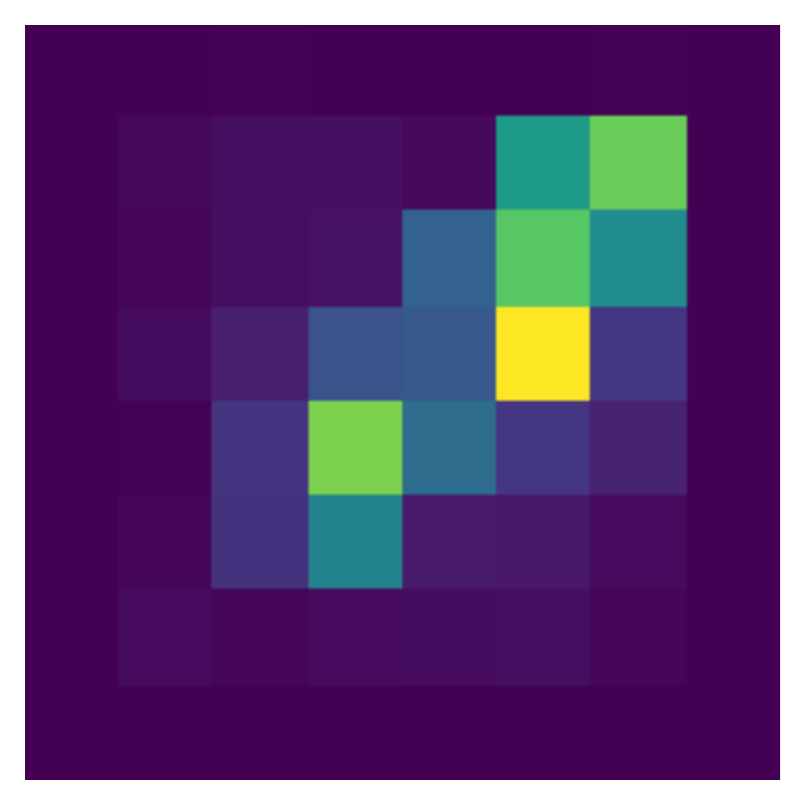}\hspace*{-0.4em}
     \includegraphics[width=0.25\textwidth]{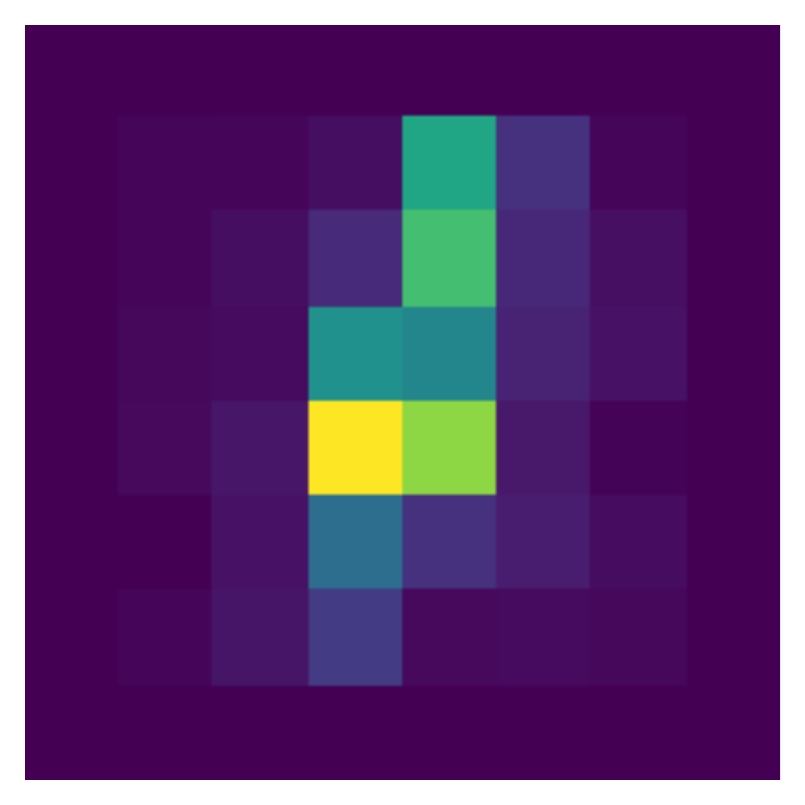}\hspace*{-0.4em}
    \\
    \includegraphics[width=0.25\textwidth]{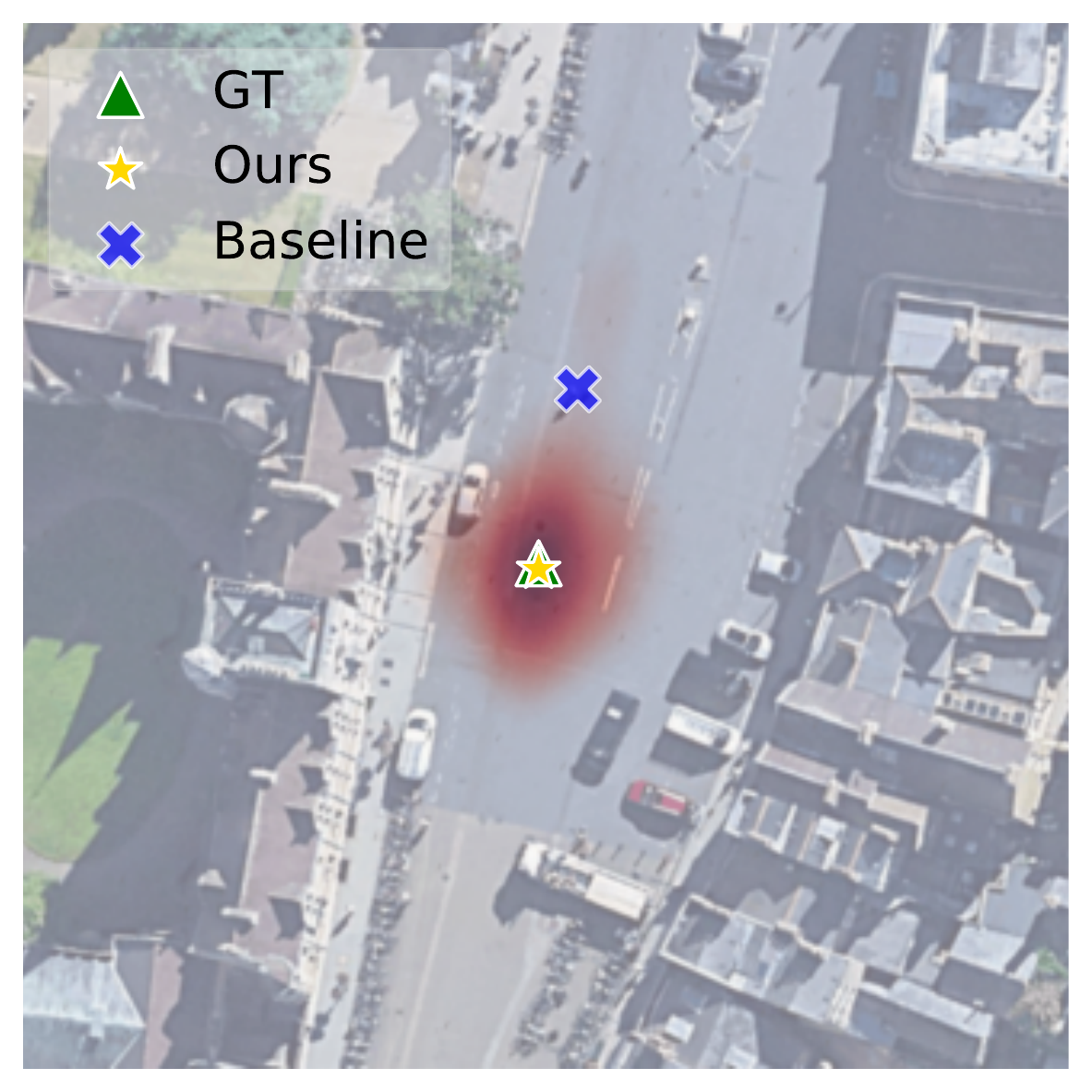}\hspace*{-0.3em}
    \includegraphics[width=0.25\textwidth]{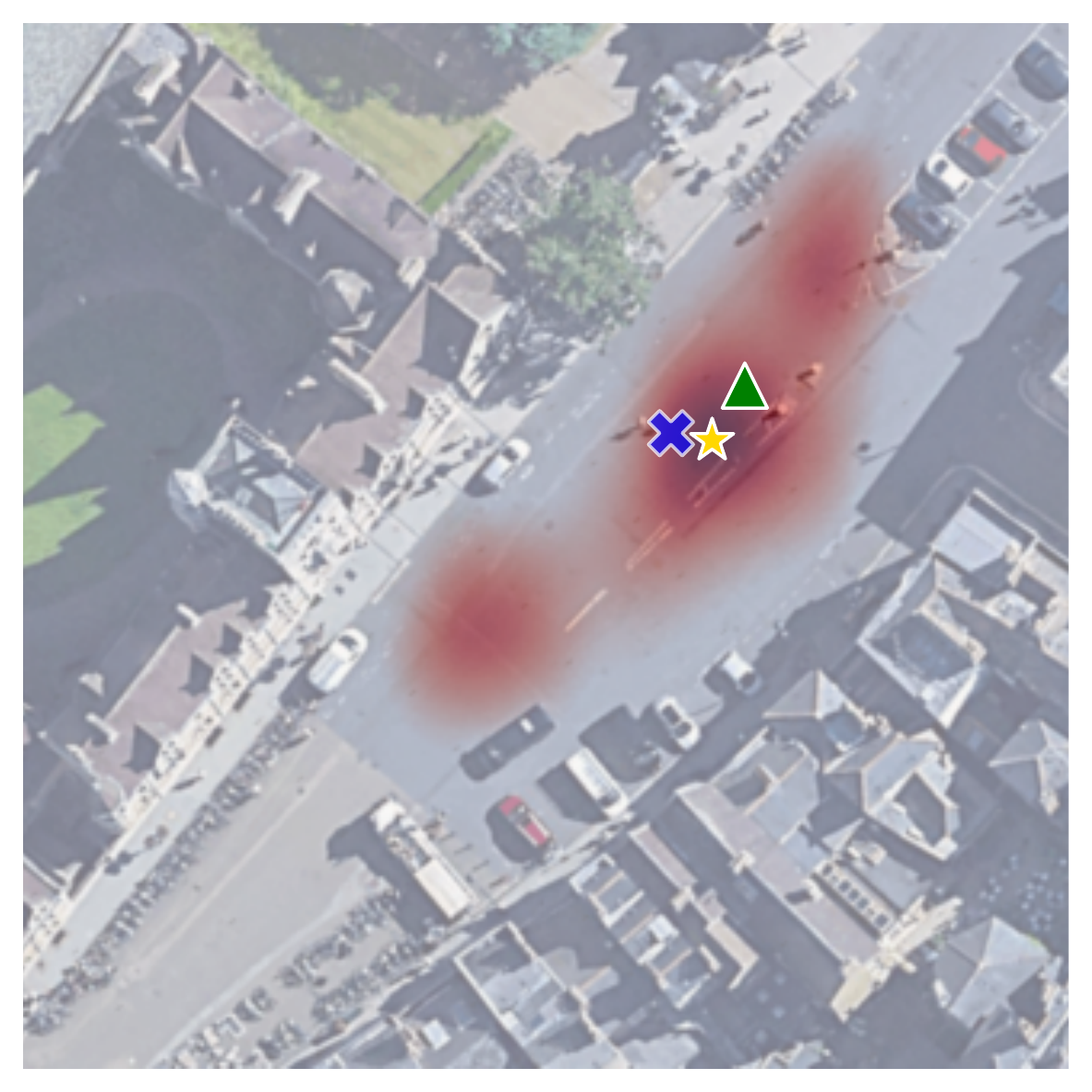}\hspace*{-0.3em}
    \includegraphics[width=0.25\textwidth]{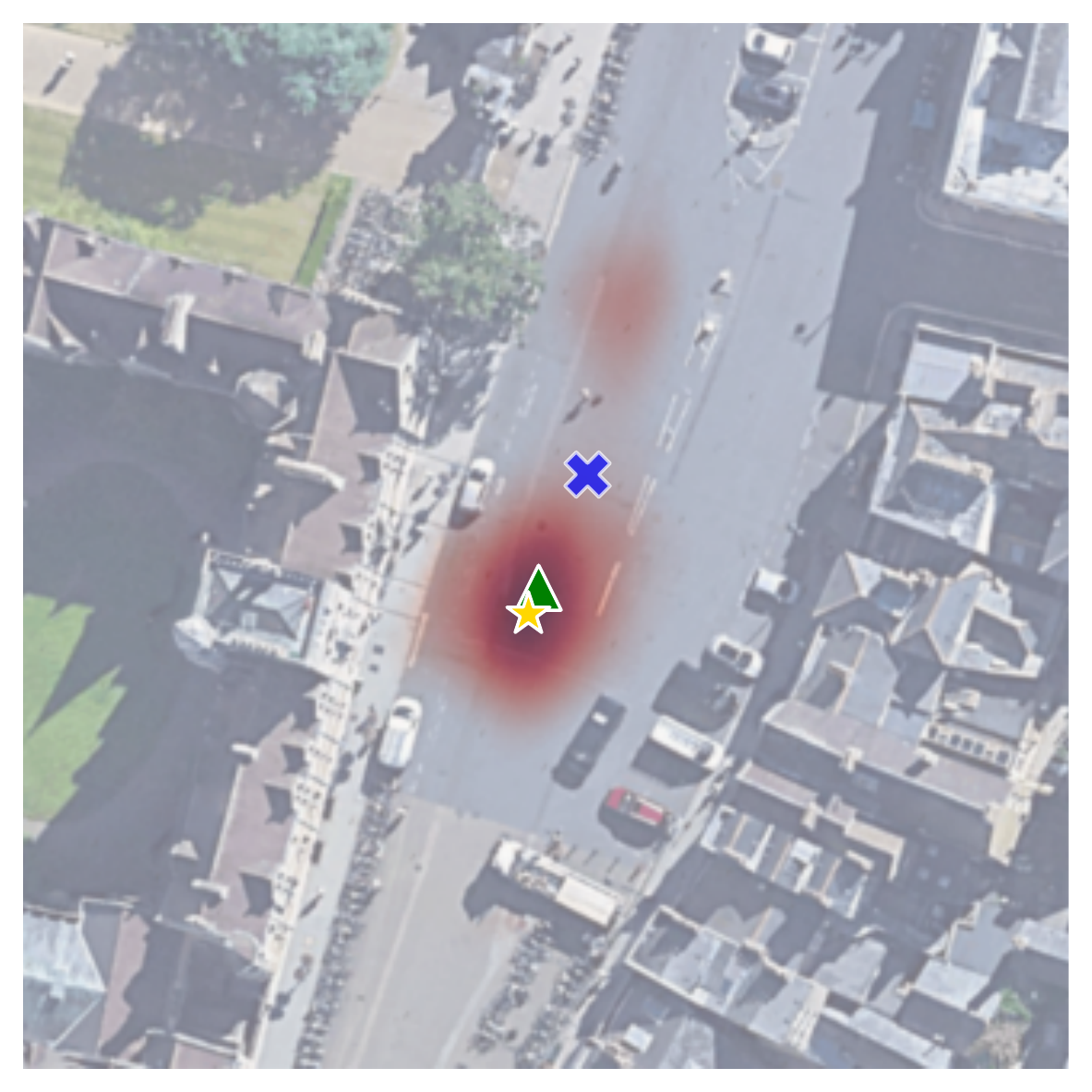}\hspace*{-0.3em}
    \\
    \caption{Examples at a rough same location in 3 test traversals on Oxford RobotCar dataset. 
    Ground images in different test traversals are collected at different times and weather.
    Each column has an input ground image, a matching score map at our model bottleneck, and an input satellite image overlayed with outputs from CVR and our method.
    Column 1/2/3 (from left to right): test traversal 1/2/3.
    % Best viewed in color.
    }
    \label{fig:Oxford_same_location}
\end{figure}

We provide more qualitative results on metric localization for both CVR and our method on three test traversals in Figure~\ref{fig:Oxford}.
Different traversals vary in time, weather, and lighting conditions.
% On this dataset, the training images are distributed in the central area in the satellite patch, and both CVR and our method explore this information.
% As we denoted in the main paper, when there are multiple visually similar locations, our model expresses the underlying multi-modal localization uncertainty.
% While the baseline could regress to the ``average'' among them.
% In the failure case, our peak probability is in the wrong mode.
In Figure~\ref{fig:Oxford_same_location}, we show the predictions from 3 test traversals at roughly the same location.
Note that the headings of the ground camera are slightly different.

\section{Network Diagram}
\begin{figure}[t]
    \centering
    \includegraphics[width=0.8\textwidth]{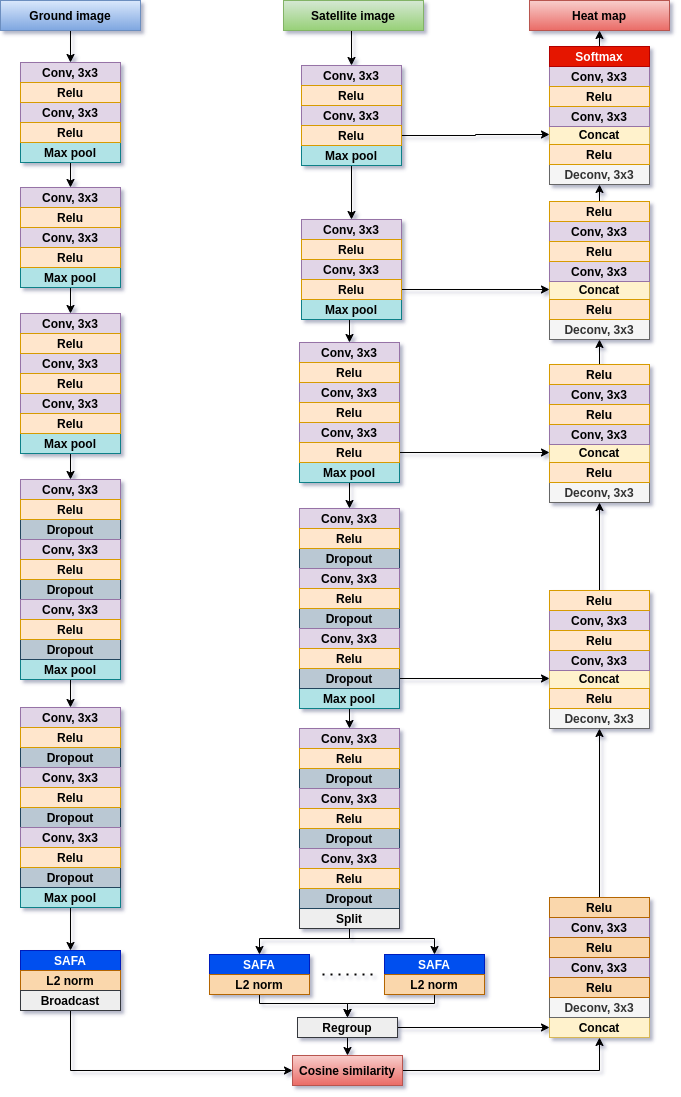}
    \caption{Network diagram
    }
    \label{fig:network_diagram}
\end{figure}

Finally, we provide our network diagram in Figure~\ref{fig:network_diagram}.
We adopt the off-the-shelf SAFA module from \cite{SAFA} for image descriptor construction.

\end{document}